\RequirePackage{fix-cm}
\documentclass[twocolumn]{svjour3}          
\smartqed  
\usepackage{graphicx}
\usepackage{amsmath,amssymb} 
\usepackage[utf8x]{inputenc}
\usepackage{pgfplots}
\pgfplotsset{compat=newest}
\usepackage{tikz}
\usetikzlibrary{backgrounds}
\usepackage{color}
\usepackage{xcolor}
\usepackage{booktabs,siunitx} 
\usepackage{makecell}
\usepackage{enumitem}
\usepackage[caption=false]{subfig}
\usepackage[noadjust]{cite}
\usepackage[colorlinks=true,allcolors=green]{hyperref}
\usepackage{rotating}

\usepackage{times}
\usepackage{pifont}
\usepackage{arydshln}
\usepackage{adjustbox}
\usepackage{array,multirow}
\newcommand{\cmark}{\ding{51}}%
\newcommand{\ie}{\mbox{\emph{i.e.\ }}}
\newcommand{\eg}{\mbox{\emph{e.g.\ }}}

\newcommand{\vecnorm}[1]{\left\|#1\right\|}

\newcommand{\best}[1]{\textbf{#1}}

\newcommand{\fnn}[1]{$\scriptstyle^{#1}$}

\definecolor{dcn}                 {RGB}{227,111, 38}
\definecolor{dcn-ft-0005}         {RGB}{128, 64,128}
\definecolor{dcn-ft-001}          {RGB}{ 62,135,207}
\definecolor{dcn-ft-002}          {RGB}{190,153,153}
\definecolor{dcn-ft-003}          {RGB}{107,142, 35}
\definecolor{dcn-ft-006}          {RGB}{200,  0,  0}

\definecolor{road}                {RGB}{128, 64,128}
\definecolor{sidewalk}            {RGB}{244, 35,232}
\definecolor{building}            {RGB}{ 70, 70, 70}
\definecolor{wall}                {RGB}{102,102,156}
\definecolor{fence}               {RGB}{190,153,153}
\definecolor{pole}                {RGB}{153,153,153}
\definecolor{traffic light}       {RGB}{250,170, 30}
\definecolor{traffic sign}        {RGB}{220,220,  0}
\definecolor{vegetation}          {RGB}{107,142, 35}
\definecolor{terrain}             {RGB}{152,251,152}
\definecolor{sky}                 {RGB}{ 70,130,180}
\definecolor{person}              {RGB}{220, 20, 60}
\definecolor{rider}               {RGB}{255,  0,  0}
\definecolor{car}                 {RGB}{  0,  0,142}
\definecolor{truck}               {RGB}{  0,  0, 70}
\definecolor{bus}                 {RGB}{  0, 60,100}
\definecolor{train}               {RGB}{  0, 80,100}
\definecolor{motorcycle}          {RGB}{  0,  0,230}
\definecolor{bicycle}             {RGB}{119, 11, 32}
\definecolor{void}                {RGB}{  0,  0,  0}

\begin{document}

\title{Curriculum Model Adaptation with Synthetic and Real Data for Semantic Foggy Scene Understanding
}

\titlerunning{Curriculum Model Adaptation with Synthetic and Real Data for Semantic Foggy Scene Understanding}

\author{Dengxin Dai         \and
        Christos Sakaridis  \and 
        Simon Hecker \and
        Luc Van Gool 
}

\institute{D. Dai  \and C. Sakaridis \and S. Hecker \and L. Van Gool \at
              ETH Z\"urich, Zurich, Switzerland
           \and
           L. Van Gool \at
              KU Leuven, Leuven, Belgium
}
\date{Received: date / Accepted: date}

\maketitle

\begin{abstract}
\sloppy{This work addresses the problem of semantic scene understanding under fog. Although marked progress has been made in semantic scene understanding, it is mainly concentrated on clear-weather scenes. Extending semantic segmentation methods to adverse weather conditions such as fog is crucial for outdoor applications. In this paper, we propose a novel method, named Curriculum Model Adaptation (CMAda), which \emph{gradually} adapts a semantic segmentation model from light synthetic fog to dense real fog in multiple steps, using both labeled synthetic foggy data and unlabeled real foggy data. The method is based on the fact that the results of semantic segmentation in moderately adverse conditions (light fog) can be bootstrapped to solve the same problem in highly adverse conditions (dense fog). CMAda is extensible to other adverse conditions and provides a new paradigm for learning with synthetic data and unlabeled real data. In addition, we present four other main stand-alone contributions: 1) a novel method to add synthetic fog to real, clear-weather scenes using semantic input; 2) a new fog density estimator; 3) a novel fog densification method for real foggy scenes without known depth; and 4) the \emph{Foggy Zurich} dataset comprising $3808$ real foggy images, with pixel-level semantic annotations for $40$ images with dense fog. Our experiments show that 1) our fog simulation and fog density estimator outperform their state-of-the-art counterparts with respect to the task of semantic foggy scene understanding (SFSU); 2) CMAda improves the performance of state-of-the-art models for SFSU significantly, benefiting both from our synthetic and real foggy data. The foggy datasets and code are publicly available.}

\end{abstract}

\section{Introduction}
\label{intro}

\sloppy{Adverse weather or illumination conditions create visibility problems for both people and the sensors that power automated systems~\cite{vision:atmosphere,vision:rain:07,SFSU_synthetic,daytime:2:nighttime}. While sensors and the downstream vision algorithms are constantly getting better, their performance is mainly benchmarked on clear-weather images~\cite{Cityscapes,drive:surroundview:route:planner}. Many outdoor applications, however, cannot escape from ``bad'' weather~\cite{vision:atmosphere}. One typical example of adverse weather conditions is fog, which degrades the visibility of a scene significantly~\cite{contrast:weather:degraded,tan2008visibility}. The denser the fog is, the more severe this problem becomes.}

During the past years, the community has made a tremendous progress in image dehazing (defogging) to increase the visibility in foggy images~\cite{bayesian:defogging,dark:channel,dehazing:mscale:depth}. The last few years have also witnessed a leap in object recognition. A great deal of effort is made specifically in semantic road scene understanding~\cite{road:scene:eccv12,Cityscapes,Tealtime3DTrafficConeDetection}. However, the extension of these techniques to other weather/illumination conditions has not received due attention, despite its importance in outdoor applications. For example, an automated car still needs to detect other traffic agents and traffic control devices in the presence of fog or rain. This work investigates the problem of semantic foggy scene understanding (SFSU).

\begin{figure*}[t]
  \centering
  \includegraphics[width=0.95\textwidth]{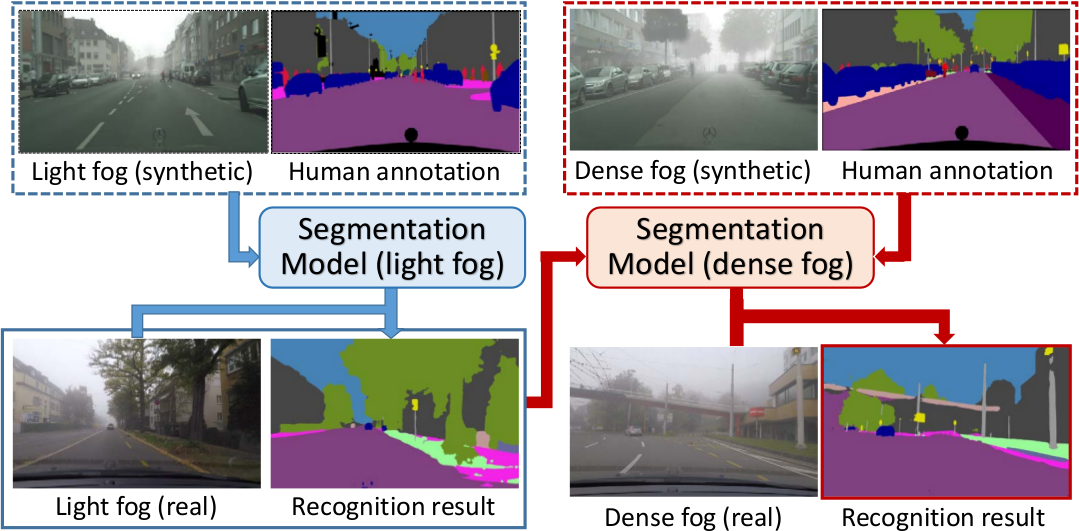}
  \caption{The illustrative pipeline of a two-stage instantation of CMAda for semantic scene understanding under dense fog}
  \label{fig:pipeline}
\end{figure*}

The current ``standard'' policy for addressing semantic scene understanding is to train a neural network with numerous annotated real images~\cite{pascal:2011,imagenet:2015,Cityscapes}. While this trend of creating and using more human annotations may still continue, extending the same protocol to all conditions seems to be problematic, as the manual annotation part is hard to scale.
The problem is more pronounced for adverse weather conditions, as the difficulty of data collection and annotation increases significantly. To overcome this problem, a few streams of research have gained extensive attention: learning with limited, weak supervision~\cite{dai:EnPro:iccv13,Misra_2015_CVPR}, transfer learning~\cite{LSDA:nips:14,DomainAdaptiveFasterRCNN}, and learning with synthetic data~\cite{Synthia:dataset,SFSU_synthetic}.  

Our method falls into the middle ground, and aims to combine the strength of these two kinds of methods. In particular, our method is developed to learn from 1) a dataset with high-quality synthetic fog and the corresponding human annotations, and 2) a dataset with a large number of unlabeled images with real fog. The goal of our method is to improve the performance of SFSU without requiring extra human annotations for foggy images.

To this end, this work proposes a novel fog simulator to add high-quality synthetic fog to real images of clear-weather outdoor scenes, and then leverage these partially synthetic foggy images for SFSU. Our fog simulator builds on the recent work of Sakaridis et al.~\cite{SFSU_synthetic}, by introducing a semantic-aware filter to exploit the structures of object instances. We show that learning with our synthetic foggy data improves the performance for SFSU. Furthermore, we learn a fog density estimator from synthetic images of varying fog density, and order unlabeled real images by increasing fog density. This ordering forms the foundation of our novel learning method Curriculum Model Adaptation (CMAda) to \emph{gradually} adapt a semantic segmentation model from \emph{clear weather} to \emph{dense fog}, through \emph{light fog}. CMAda is based on the fact that recognition in moderately adverse conditions (light fog) is easier and its results can be re-used via \emph{knowledge distillation} to solve a harder problem, \ie{}recognition in highly adverse conditions (dense fog).

CMAda is iterative by nature and can be implemented for different numbers of steps. The pipeline of a two-step implementation of CMAda is shown in Figure~\ref{fig:pipeline}. CMAda has the potential to be used for other adverse weather conditions, and opens a new avenue for learning with synthetic data and unlabeled real data in general.  
Experiments show that CMAda yields the best results on two datasets with dense real fog as well as a dataset with real fog of varying density.

A shorter version of this work has been published to European Conference on Computer Vision~\cite{dense:SFSU:eccv18}. Compared to the conference version, this paper makes the following six additional contributions: 
\begin{enumerate}
    \item An extension of the formulation of CMAda to accommodate multiple adaptation steps instead of only two steps, leading to improved performance over the conference paper as well.
    \item A novel fog densification method for real foggy scenes. The fog densification method can close the domain gap between light real fog and dense real fog; using it in CMAda significantly increases the performance for SFSU.
    \item A method named Model Selection for the task of semantic scene understanding \emph{in multiple weather conditions} where test images are a mixture of clear-weather images and foggy images. This extension is important for real world applications, as weather conditions change constantly. Semantic scene understanding methods need to be robust to such changes.
    \item An enlarged annotated dense foggy set for our \emph{Foggy Zurich} dataset, increasing its size from $16$ to $40$ images.\footnote{Creating fine pixel-level annotations for dense foggy scenes is very difficult.}
    \item More extensive experiments to diagnose the contribution of each component of the CMAda pipeline, to compare with more competing methods, and to comprehensively study the usefulness of image dehazing for SFSU. 
    \item Other sections are also enhanced, including related work as well as dataset collection and annotation. 
\end{enumerate}

The paper is structured as follows. Section~\ref{sec:related} presents the related work. Section~\ref{sec:simulation} is devoted to our method for simulating synthetic fog, which is followed by Section~\ref{sec:semantic:learning} for our learning approach. Section~\ref{sec:dataset} summarizes our data collection and annotation. Finally, Section~\ref{sec:experiments} presents our experimental results and Section~\ref{sec:conclusion} concludes this paper. Our foggy datasets and fog simulation code are publicly available at \url{https://www.vision.ee.ethz.ch/~csakarid/Model_adaptation_SFSU_dense/}.

\section{Related Work}
\label{sec:related}

Our work is relevant to image defogging, joint image filtering, foggy scene understanding, and domain adaptation.

\subsection{Image Defogging/Dehazing}
Fog fades the color of observed objects and reduces their contrast. Extensive research has been conducted on image defogging (dehazing) to increase the visibility of foggy scenes~\cite{contrast:weather:degraded,tan2008visibility,bayesian:defogging,fattal2008single,nonlocal:image:dehazing,fattal2014dehazing,dark:channel}. Certain works focus particularly on enhancing foggy road scenes~\cite{THC+12,exponential:contrast:restoration}. Recent approaches also rely on trainable architectures~\cite{TYW14}, which have evolved to end-to-end models~\cite{joint:transmission:estimation:dehazing,deep:transmission:network}. For a comprehensive overview of defogging/dehazing algorithms, we point the reader to~\cite{review:defogging:restoration,dehazing:survey:benchmarking}. Our work is complementary and mainly focuses on SFSU, while it also investigates the usefulness of image dehazing in the context of SFSU. 

\subsection{Joint Image Filtering}
Using additional images as input for filtering a target image has been originally studied in settings where the target image has low photometric quality~\cite{flash:enhance:crossbilateral:Eisemann,flash:enhance:crossbilateral:Petschnigg} or low resolution~\cite{joint:bilateral:upsampling}. Compared to the bilateral filtering formulation of these approaches, subsequent works propose alternative formulations, such as the guided filter~\cite{guided:filtering} and mutual structure filtering~\cite{mutual:structure:filtering}, for better incorporating the reference image into the filtering process. In comparison, we extend the classical cross-bilateral filter to a dual-reference cross-bilateral filter by accepting \emph{two} reference images, one of which is a discrete label image that helps our filter adhere to the semantics of the scene.

\subsection{Foggy Scene Understanding}
Typical examples in this line include road and lane detection~\cite{recent:progress:lane}, traffic light detection~\cite{traffic:light:survey:16}, car and pedestrian detection~\cite{kitti}, and a dense, pixel-level segmentation of road scenes into most of the relevant semantic classes~\cite{recognition:sfm:eccv08,Cityscapes}. While deep recognition networks have been developed~\cite{dilated:convolution,refinenet,pspnet,fast:rcnn,faster:rcnn} and large-scale datasets have been presented~\cite{kitti,Cityscapes}, that research mainly focused on clear weather. There is also a large body of work on fog detection~\cite{fog:detection:cv:09,fog:detection:vehicles:12,night:fog:detection,fast:fog:detection}. Classification of scenes into foggy and fog-free has been tackled as well~\cite{fog:nonfog:classification:13}. In addition, visibility estimation has been extensively studied for both daytime~\cite{visibility:road:fog:10,visibility:detection:fog:15,fog:detection:visibility:distance} and nighttime~\cite{night:visibility:analysis:15}, in the context of assisted and autonomous driving. The closest of these works to ours is~\cite{visibility:road:fog:10}, in which synthetic fog is generated and foggy images are segmented to \emph{free-space area} and \emph{vertical objects}. Our work differs in that our semantic scene understanding task is more complex and we tackle the problem from a different route by learning jointly from synthetic fog and real fog.

\subsection{Domain Adaptation}
Our work bears resemblance to transfer learning and model adaptation. Model adaptation across weather conditions to semantically segment simple road scenes is studied in~\cite{road:scene:2013}. More recently, domain adversarial based approaches were proposed to adapt semantic segmentation models both at pixel level and feature level from simulated to real environments~\cite{adversarial:training:simulated:17,synthetic:semantic:segmentation,CyCADA,incremental:adversarial:DA:18}. 
Most of these works are based on adversarial domain adaptation. Our work is complementary to methods in this vein; we adapt the model parameters with carefully generated data, leading to an algorithm whose behavior is easy to understand and whose performance is more predictable. Combining our method and adversarial domain adaptation is a promising direction.  Our work also shares similarity to~\cite{curriculum:domain:adaptation:17} in applying the general idea of curriculum learning to domain adaptation. 

The concurrent work in~\cite{daytime:2:nighttime} on adaptation of semantic segmentation models from daytime to nighttime using solely real data, which was preceded by the conference version of this paper, shows that real images captured at twilight are helpful for supervision transfer from daytime to nighttime. CMAda constitutes a more complex framework, since it leverages both synthetic foggy data and real foggy data \emph{jointly} for adapting semantic segmentation models to fog, whereas the method in~\cite{daytime:2:nighttime} uses solely real data for the adaptation. Moreover, the assignment of real foggy images to the correct target foggy domain through fog density estimation is another crucial and nontrivial component of CMAda and it is a prerequisite for using these real images as training data in the method. By contrast, the partition of the real dataset in~\cite{daytime:2:nighttime} into subsets that correspond to different times of day from daytime to nighttime is trivially performed by using the time of capture of the images.

\section{Fog Simulation on Real Scenes Using Semantics}
\label{sec:simulation}

\begin{figure*}[tb]
  \centering
  \includegraphics[width=\textwidth]{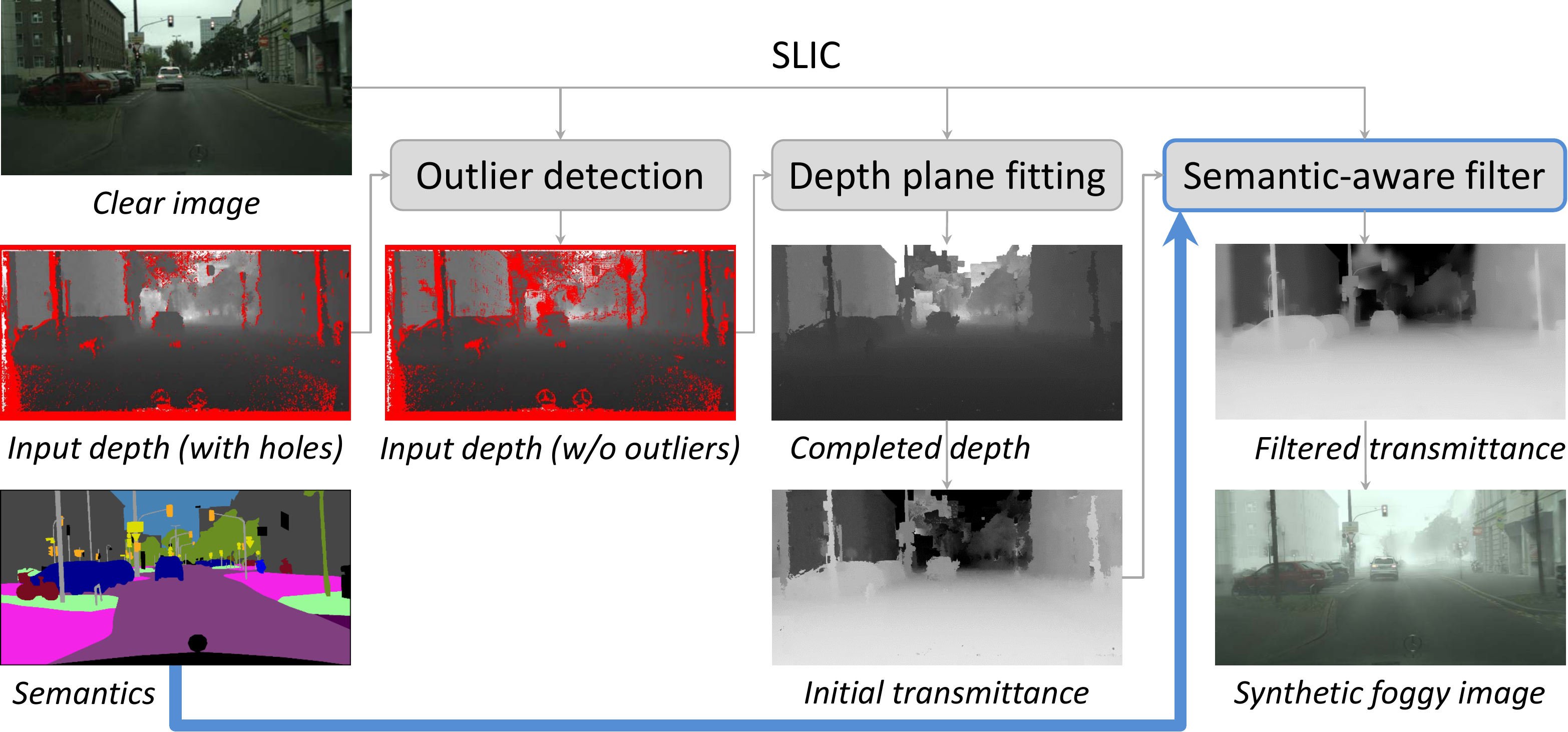}
  \caption{The pipeline of our fog simulation using semantics}
  \label{fig:simulation:pipeline}
\end{figure*}

\subsection{Motivation}

We drive our motivation for fog simulation on real scenes using semantic input from the pipeline that was used in~\cite{SFSU_synthetic} to generate the Foggy Cityscapes dataset, which primarily focuses on depth denoising and completion. This pipeline is denoted in Figure~\ref{fig:simulation:pipeline} with thin gray arrows and consists of three main steps: depth outlier detection, robust depth plane fitting at the level of SLIC superpixels~\cite{slic:superpixels} using RANSAC, and postprocessing of the completed depth map with guided image filtering~\cite{guided:filtering}. Our approach adopts the general configuration of this pipeline, but aims to improve its postprocessing step by leveraging the semantic annotation of the scene as additional reference for filtering, which is indicated in Figure~\ref{fig:simulation:pipeline} with the thick blue arrow.

\sloppy{The guided filtering step in~\cite{SFSU_synthetic} uses the clear-weather color image as guidance to filter the depth map. However, as previous works on image filtering~\cite{mutual:structure:filtering} have shown, guided filtering and similar joint filtering methods such as cross-bilateral filtering~\cite{flash:enhance:crossbilateral:Eisemann,flash:enhance:crossbilateral:Petschnigg} transfer every structure that is present in the guidance/reference image to the output target image. Thus, any structure that is specific to the reference image but irrelevant for the target image is transferred to the latter erroneously.}

Whereas previous approaches such as mutual-structure filtering~\cite{mutual:structure:filtering} attempt to estimate the common structure between reference and target images, we identify this common structure with the structure that is present in the ground-truth \emph{semantic labeling} of the image. In other words, we assume that edges which are shared by the color image and the depth map generally coincide with \emph{semantic edges}, \ie{}locations in the image where the semantic classes of adjacent pixels are different. Under this assumption, the semantic labeling can be used directly as the reference image in a classical cross-bilateral filtering setting, since it contains exactly the mutual structure between the color image and the depth map. In practice, however, the boundaries drawn by humans when creating semantic annotations are not pixel-accurate, and using the color image as additional reference helps to capture the precise location and orientation of edges better. As a result, we formulate the postprocessing step of the completed depth map in our fog simulation as a \emph{dual-reference} cross-bilateral filter, with color and semantic reference.

\sloppy{Before delving into the formulation of our filter, we briefly argue against alternative usage cases of semantic annotations in our fog simulation pipeline which might seem attractive at first sight. First, replacing SLIC superpixels with superpixels induced by the semantic labeling for the depth plane fitting step is not viable, because it induces very large superpixels, for which the planarity assumption breaks completely. Second, we have experimented with omitting the robust depth plane fitting step altogether and applying our dual-reference cross-bilateral filter directly on the incomplete depth map which is output from the outlier detection step. This approach, however, is highly sensitive to outliers that have not been detected and invalidated in the preceding step. By contrast, these remaining outliers are handled successfully by robust RANSAC-based depth plane fitting.}

\subsection{Dual-reference Cross-bilateral Filter Using Color and Semantics}
\label{sec:simulation:dual_bilateral}

Let us denote the RGB image of the clear-weather scene by $\mathbf{R}$ and its CIELAB counterpart by $\mathbf{J}$. We consider CIELAB, as it has been designed to increase perceptual uniformity and gives better results for bilateral filtering of color images~\cite{bilateral:grid}. The input image to be filtered in the postprocessing step of our pipeline constitutes a scalar-valued transmittance map $\hat{t}$. We provide more details on this transmittance map in Section~\ref{sec:simulation:rest}. Last, we are given a labeling function
\begin{equation} \label{eq:labeling}
h: \mathcal{P} \to \{1,\,\dots,\,C\}
\end{equation}
which maps pixels to semantic labels, where $\mathcal{P}$ is the discrete domain of pixel positions and $C$ is the total number of semantic classes in the scene. We define our dual-reference cross-bilateral filter with color and semantic reference as

\begin{align}
{t}(\mathbf{p}) = &\left\{\displaystyle\sum_{\mathbf{q} \in \mathcal{N}(\mathbf{p})} G_{\sigma_s}(\vecnorm{\mathbf{q}-\mathbf{p}}) \left[\delta(h(\mathbf{q})-h(\mathbf{p}))\vphantom{\mu G_{\sigma_c}(\vecnorm{\mathbf{J}(\mathbf{q})-\mathbf{J}(\mathbf{p})})}\right.\vphantom{\hat{t}(\mathbf{q})}\right. \nonumber\\
&{+}\:\left.\vphantom{\displaystyle\sum_{q \in \mathcal{N}(\mathbf{p})} G_{\sigma_s}(\vecnorm{\mathbf{q}-\mathbf{p}})}\left.\vphantom{\delta(h(\mathbf{q})-h(\mathbf{p}))}\mu G_{\sigma_c}(\vecnorm{\mathbf{J}(\mathbf{q})-\mathbf{J}(\mathbf{p})})\right] \hat{t}(\mathbf{q})\right\} \nonumber\\
&\left/\:\middle\{\displaystyle\sum_{\mathbf{q} \in \mathcal{N}(\mathbf{p})} G_{\sigma_s}(\vecnorm{\mathbf{q}-\mathbf{p}}) \left[\delta(h(\mathbf{q})-h(\mathbf{p}))\vphantom{\mu G_{\sigma_c}(\vecnorm{\mathbf{J}(\mathbf{q})-\mathbf{J}(\mathbf{p})})}\right.\right. \nonumber\\
&{+}\:\left.\vphantom{\displaystyle\sum_{q \in \mathcal{N}(\mathbf{p})} G_{\sigma_s}(\vecnorm{\mathbf{q}-\mathbf{p}})}\left.\vphantom{\delta(h(\mathbf{q})-h(\mathbf{p}))}\mu G_{\sigma_c}(\vecnorm{\mathbf{J}(\mathbf{q})-\mathbf{J}(\mathbf{p})})\right]\right\},  \label{eq:dual_bilateral}
\end{align}
where $\mathbf{p}$ and $\mathbf{q}$ denote pixel positions, $\mathcal{N}(\mathbf{p})$ is the neighborhood of $\mathbf{p}$, $\delta$ denotes the Kronecker delta, $G_{\sigma_s}$ is the spatial Gaussian kernel, $G_{\sigma_c}$ is the color-domain Gaussian kernel and $\mu$ is a positive constant. The novel dual reference is demonstrated in the second factor of the filter weights, which constitutes a sum of the terms $\delta(h(\mathbf{q})-h(\mathbf{p}))$ for semantic reference and $G_{\sigma_c}(\vecnorm{\mathbf{J}(\mathbf{q})-\mathbf{J}(\mathbf{p})})$ for color reference, weighted by $\mu$. The formulation of the semantic term implies that only pixels $\mathbf{q}$ with the same semantic label as the examined pixel $\mathbf{p}$ contribute to the output at $\mathbf{p}$ through this term, which prevents blurring of semantic edges. At the same time, the color term helps to better preserve true depth edges that do not coincide with any semantic boundary but are present in $\mathbf{J}$, \eg{}due to self-occlusion of an object.

The formulation of~\eqref{eq:dual_bilateral} enables an efficient implementation of our filter based on the bilateral grid~\cite{bilateral:grid}. More specifically, we construct two separate bilateral grids that correspond to the semantic and color domains respectively and operate separately on each grid to perform filtering, combining the results in the end. In this way, we handle a 3D bilateral grid for the semantic domain and a 5D grid for the color domain instead of a single joint 6D grid that would dramatically increase computation time~\cite{bilateral:grid}.

In our experiments, we set $\mu = 5$, $\sigma_s = 20$, and $\sigma_c = 10$.

\subsection{Remaining Steps}
\label{sec:simulation:rest}

\begin{figure*}[tb]
    \centering
    \subfloat{\includegraphics[width=0.31\textwidth]{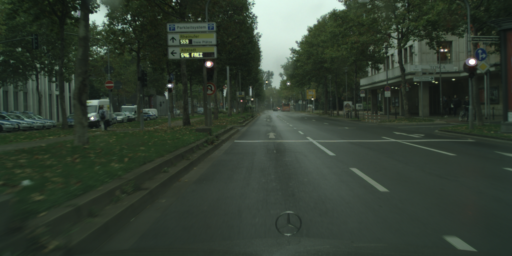}}
    \hfil
    \subfloat{\includegraphics[width=0.31\textwidth]{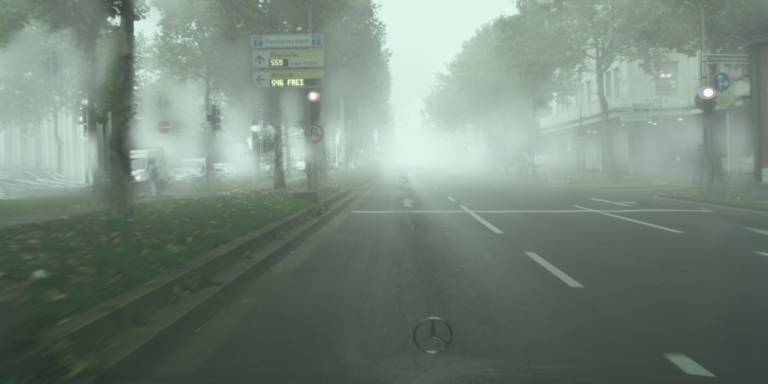}}
    \hfil
    \subfloat{\includegraphics[width=0.31\textwidth]{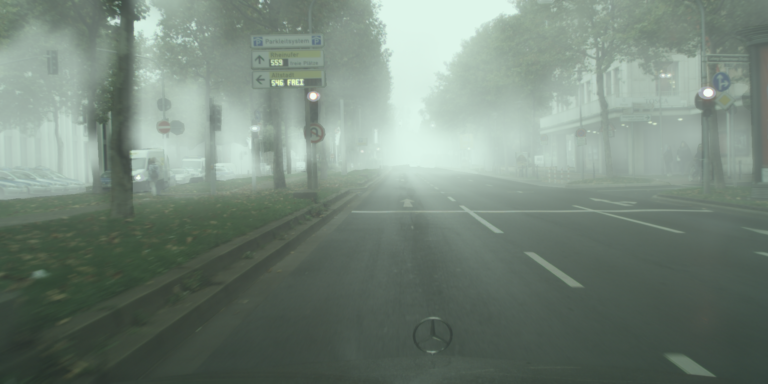}}
    \\
    \subfloat{\includegraphics[width=0.31\textwidth]{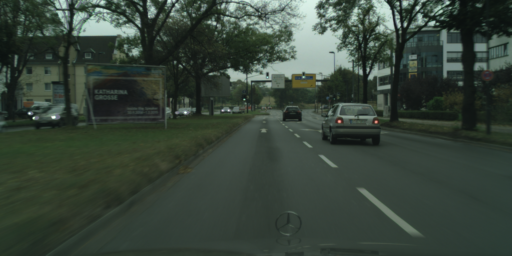}}
    \hfil
    \subfloat{\includegraphics[width=0.31\textwidth]{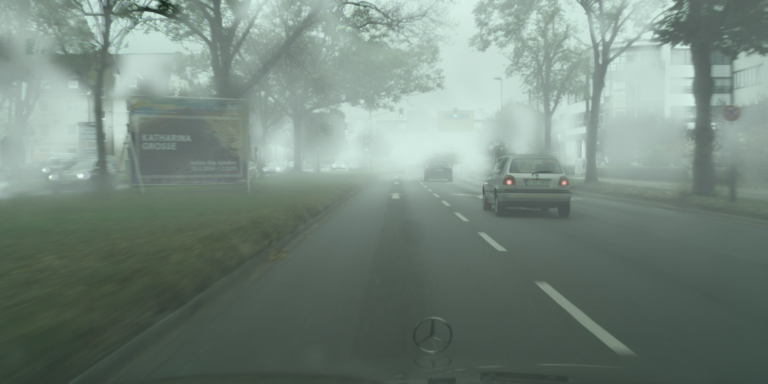}}
    \hfil
    \subfloat{\includegraphics[width=0.31\textwidth]{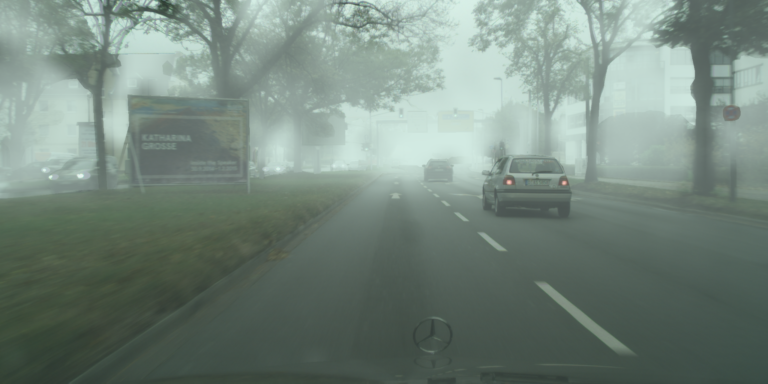}}
    \\
    \addtocounter{subfigure}{-6}
    \subfloat[Cityscapes]{\includegraphics[width=0.31\textwidth]{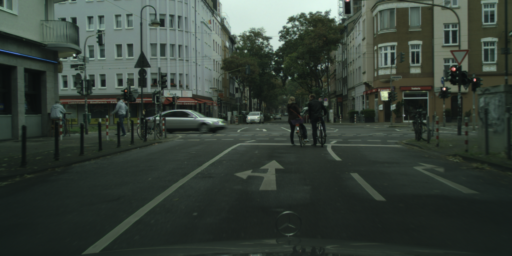}\label{fig:fog:simulation:input}}
    \hfil
    \subfloat[Foggy Cityscapes]{\includegraphics[width=0.31\textwidth]{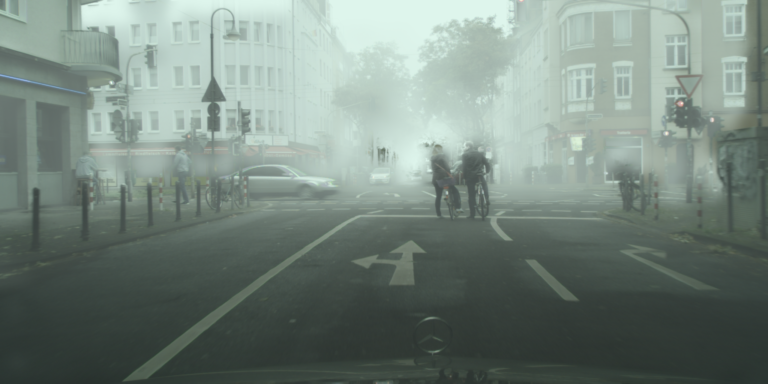}\label{fig:fog:simulation:sfsu_synthetic}}
    \hfil
    \subfloat[Our foggy image - \emph{Foggy Cityscapes-DBF}]{\includegraphics[width=0.31\textwidth]{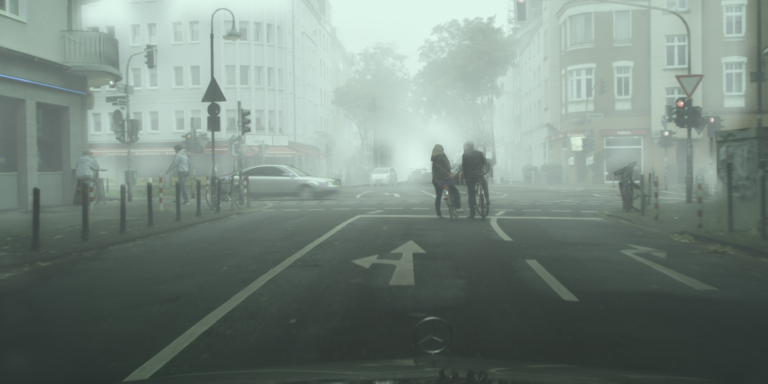}\label{fig:fog:simulation:ours}}
    \caption{Comparison of our synthetic foggy images against Foggy Cityscapes~\cite{SFSU_synthetic}. This figure is better seen on a screen and zoomed in}
    \label{fig:fog:simulation}
\end{figure*}

Here we outline the rest parts of our fog simulation pipeline of Figure~\ref{fig:simulation:pipeline}. For more details, we refer the reader to~\cite{SFSU_synthetic}, with which most parts of the pipeline are common. The standard optical model for fog that forms the basis of our fog simulation was introduced in~\cite{Koschmieder:optical:model} and is expressed as
\begin{equation} \label{eq:fog:model}
\mathbf{I}(\mathbf{x}) = \mathbf{R}(\mathbf{x})t(\mathbf{x}) + \mathbf{L}(1 - t(\mathbf{x})),
\end{equation}
where $\mathbf{I}(\mathbf{x})$ is the observed foggy image at pixel $\mathbf{x}$, $\mathbf{R}(\mathbf{x})$ is the clear scene radiance and $\mathbf{L}$ is the atmospheric light, which is assumed to be globally constant. The transmittance $t(\mathbf{x})$ determines the amount of scene radiance that reaches the camera. For homogeneous fog, transmittance depends on the distance $\ell(\mathbf{x})$ of the scene from the camera through
\begin{equation} \label{eq:transmittance}
t(\mathbf{x}) = \exp\left(-\beta\ell(\mathbf{x})\right).
\end{equation}
The attenuation coefficient $\beta$ controls the density of the fog: larger values of $\beta$ mean denser fog. Fog decreases the meteorological optical range (MOR), also known as visibility, to less than 1 km by definition~\cite{Federal:meteorological:handbook}. For homogeneous fog $\text{MOR}=2.996/\beta$, which implies
\begin{equation} \label{eq:beta:bound:fog}
\beta \geq 2.996\times{}10^{-3}{\text{ m}}^{-1},
\end{equation}
where the lower bound corresponds to the lightest fog configuration. In our fog simulation, the value that is used for $\beta$ always obeys \eqref{eq:beta:bound:fog}.

The required inputs for fog simulation with \eqref{eq:fog:model} are the image $\mathbf{R}$ of the original clear scene, atmospheric light $\mathbf{L}$ and a complete transmittance map $t$. We use the same approach for atmospheric light estimation as that in~\cite{SFSU_synthetic}. Moreover, we adopt the stereoscopic inpainting method of~\cite{SFSU_synthetic} for depth denoising and completion to obtain an initial complete transmittance map $\hat{t}$ from a noisy and incomplete input disparity map $D$, using the recommended parameters. We filter $\hat{t}$ with our dual-reference cross-bilateral filter~\eqref{eq:dual_bilateral} to compute the final transmittance map $t$, which is used in~\eqref{eq:fog:model} to synthesize the foggy image $\mathbf{I}$.

Results of the presented pipeline for fog simulation on example images from Cityscapes~\cite{Cityscapes} are provided in Figure~\ref{fig:fog:simulation} for $\beta = 0.02$, which corresponds to visibility of ca.\ $150\text{m}$. We specifically leverage the instance-level semantic annotations that are provided in Cityscapes and set the labeling $h$ of~\eqref{eq:labeling} to a different value for each distinct instance of the same semantic class in order to distinguish adjacent instances. We compare our synthetic foggy images against the respective images of Foggy Cityscapes that were generated with the approach of~\cite{SFSU_synthetic}. Our synthetic foggy images generally preserve the edges between adjacent objects with large discrepancy in depth better than the images in Foggy Cityscapes, because our approach utilizes semantic boundaries, which usually encompass these edges. The incorrect structure transfer of color textures to the transmittance map, which deteriorates the quality of Foggy Cityscapes, is also reduced with our method. 

We have applied our fog simulation using semantics to the entire Cityscapes dataset. The resulting foggy dataset is named \emph{Foggy Cityscapes-DBF} (\textbf{D}ual-reference cross-\textbf{B}ilateral \textbf{F}ilter). \emph{Foggy Cityscapes-DBF} is publicly available at the Cityscapes website \url{https://www.cityscapes-dataset.com}.

\section{Semantic Foggy Scene Understanding}
\label{sec:semantic:learning} 

In this section, we first present a standard supervised learning approach for semantic segmentation under dense fog using our synthetic foggy data with the novel fog simulation of Section~\ref{sec:simulation}, and then elaborate on our novel CMAda approach which uses both synthetic and real foggy data. 

\subsection{Learning with Synthetic Fog}
\label{sec:learning:syn}
Generating synthetic fog from real clear-weather scenes grants the potential of inheriting the existing human annotations of these scenes, such as those from the Cityscapes dataset~\cite{Cityscapes}. This is a significant asset that enables training of standard segmentation models. Therefore, an effective way of evaluating the merit of a fog simulator is to adapt a segmentation model originally trained on clear weather to the synthesized foggy images and then evaluate the adapted model against the original one on real foggy images. The primary goal is to verify that the standard learning methods for semantic segmentation can benefit from our simulated fog in the challenging scenario of real fog. This evaluation policy has been proposed in~\cite{SFSU_synthetic}. We adopt this policy and fine-tune the RefineNet model~\cite{refinenet} on synthetic foggy images from our \emph{Foggy Cityscapes-DBF} dataset. The performance of our adapted models on real fog is compared to that of the original clear-weather model as well as the models that are adapted on Foggy Cityscapes~\cite{SFSU_synthetic}, providing an objective comparison of our simulation method against~\cite{SFSU_synthetic}.

The learned model can be used as a standalone approach for semantic foggy scene understanding as shown in~\cite{SFSU_synthetic}, or it can be used as an initialization step for our CMAda method, which is described next and learns both from synthetic and real data. 

\subsection{Curriculum Model Adaptation (CMAda)}

In the previous section, the proposed method learns to adapt semantic segmentation models from the domain of clear weather to the domain of foggy weather in a single step. While considerable improvement can be achieved (as shown in Section~\ref{sec:experiments:synthetic}), the method falls short when it is presented with dense fog. This is because domain discrepancies become more accentuated for denser fog: 1) the domain discrepancy between synthetic foggy images and real foggy images increases with fog density; and 2) the domain discrepancy between real clear-weather images and real foggy images increases with fog density. This section presents a method to gradually adapt the semantic segmentation model which was originally trained with clear-weather images to images with dense fog by using both labeled synthetic foggy images and unlabeled real foggy images. The method, which we term Curriculum Model Adaptation (CMAda), uses synthetic fog with a range of varying fog density---from light fog to dense fog---and a large dataset of unlabeled real foggy scenes with variable, unknown fog density. The goal is to improve the performance of state-of-the-art semantic segmentation models on dense foggy scenes without using any human annotations of foggy scenes. Below, we first present our fog density estimator and our method for densification of fog in real foggy images without depth information, and then proceed to the complete learning approach. 

\subsubsection{Fog Density Estimation}
\label{sec:fog:density:estimation}
Fog density is usually determined by the visibility of the foggy scene. An accurate estimate of fog density can benefit many applications, such as image defogging~\cite{fog:density:15}. Since annotating images in a fine-grained manner regarding fog density is very challenging, previous methods are trained on a few hundreds of images divided into only two classes: foggy and fog-free~\cite{fog:density:15}. The performance of the system, however, is affected by the small amount of training data and the coarse class granularity.

In this paper, we leverage our fog simulation applied to Cityscapes~\cite{Cityscapes} for fog density estimation. Since simulated fog density is directly controlled through $\beta$, we generate several versions of \emph{Foggy Cityscapes-DBF} with varying $\beta \in \{0,\,0.005,\,0.01,\,0.02\}$ and train AlexNet~\cite{alexnet} to regress the value of $\beta$ for each image, lifting the need to handcraft features relevant to fog and to collect human annotations as \cite{fog:density:15} did. The predicted fog density with our method on real images correlates well with human judgments of fog density, based on a user study conducted on our large real \emph{Foggy Zurich} dataset via Amazon Mechanical Turk (cf.\ Section~\ref{sec:exp:fog:density} for results). The fog density estimator is used to order images in \emph{Foggy Zurich} according to fog density, paving the way for our curriculum adaptation which learns from images with progressively denser fog. We denote the estimator by $f: \mathbf{x} \rightarrow \mathbb{R}^+$, where $\mathbf{x}$ is an image.

\subsubsection{CMAda with Synthetic and Real Fog}
\label{sec:CMAda}

The CMAda algorithm has a \emph{source domain} denoted by $S$, an \emph{ultimate target domain} denoted by $T$, and an ordered sequence of \emph{intermediate target domains} indicated by $(\dot{T}_1, ..., \dot{T}_K)$ with $K$ being the number of intermediate domains. In this work, $S$ is clear weather, $T$ is dense fog, and $\dot{T_k}$'s correspond to fog density that increases with $k$, ranging between the density of $S$ (zero) and $T$. Our method adapts semantic segmentation models through the sequence of domains $(S,\,\dot{T}_1,\,\dot{T}_2,\,\dots,\,\dot{T}_K,\,T)$.
The \emph{intermediate target domains} $\dot{T}_k$'s are optional; when $K=0$, the method reduces to a single-stage adaptation as presented in Section~\ref{sec:learning:syn}. Similarly, $K=1$ leads to a two-stage adaptation approach as presented in the conference version of this paper~\cite{dense:SFSU:eccv18}, $K=2$ to a three-stage adaptation approach, and so on. We abbreviate these instantiations of CMAda as CMAda1 ($K=0$), CMAda2 ($K=1$), CMAda3 ($K=2$), and so on.  

Let us denote by $z \in \{1, ..., Z\}$ the domain index in the above ordered sequence $(S,\,\dot{T}_1,\,\dot{T}_2,\,\dots,\,\dot{T}_K,\,T)$, with $Z=K+2$. In this work, the sequence of domains is sorted in ascending order with respect to fog density. For instance, it could be $($\emph{clear weather}, \emph{light fog}, \emph{dense fog}$)$, with \emph{clear weather} being the source domain, \emph{dense fog} the ultimate target domain and \emph{light fog} the intermediate target domain. The approach proceeds progressively and adapts the semantic segmentation model from the current domain (fog density) to the subsequent one by learning from the corresponding synthetic foggy dataset and the corresponding real foggy dataset. Once the model for the subsequent domain has been trained, its knowledge is distilled on unlabeled real foggy images from that domain, and then used along with a denser version of synthetic foggy data to adapt this model to the next domain (\ie{}the immediately higher fog density).

Since the method proceeds in an iterative manner, we only present the algorithmic details for model adaptation from $z-1$ to $z$. Let us use $\beta_z$ to indicate the fog density for domain $z$, represented as the attenuation coefficient. In order to adapt the semantic segmentation model $\phi^{z-1}$ from the previous domain $z-1$ to the current domain $z$, we generate synthetic fog of the exact fog density $\beta_z$ and inherit the human annotations of the original clear-weather images. Thus, the synthetic foggy dataset for adapting to $z$ is
\begin{equation}
\label{eq:dataset:syn}
 \mathcal{D}_{\text{syn}}^z=\{(\vec{\bar{x}}_m^{\beta_z},\vec{y}_m^1)\}_{m=1}^M,   
\end{equation}
where $M$ is the total number of synthetic foggy images, $\vec{y}_m^1(i,j) \in\{1, ..., C\}$ is the label of pixel $(i,j)$ of the clear-weather image $\vec{x}_m^{\beta_1}$ ($\beta_1 = 0$), and $C$ is the total number of classes.

For real foggy images, since no human annotations are available, we rely on a strategy of self-learning or curriculum learning. Objects in lighter fog are easier to recognize than in denser fog, hence models trained for lighter fog are more generalizable to real data. The model $\phi^{z-1}$ for the previous domain $z-1$ can be applied to all real foggy images with fog density less than $\beta_{z-1}$ in order to generate supervisory labels for training model $\phi^{z}$ for domain $z$. Specifically, the real foggy dataset for adapting to $z$ is
\begin{equation}
\label{eq:dataset:real}
    \mathcal{D}_{\text{real}}^z=\{(\vec{x}_n, \hat{\vec{y}}_n^{z-1}) \: | \: f(\vec{x}_n)  \leq \beta_{z-1}\}_{n=1}^N,
\end{equation} where $\hat{\vec{y}}_n^{z-1}=\phi^{z-1}(\vec{x}_n)$ denotes the predicted labels of image $\vec{x}_n$ using the model $\phi^{z-1}$. 

Once the two training sets are formed, the aim is to learn $\phi^z$ from $\mathcal{D}_{\text{syn}}^z$ and $\mathcal{D}_{\text{real}}^z$. The proposed scheme balances the contributions of both the synthetic foggy dataset $\mathcal{D}_{\text{syn}}^z$ from domain $z$ with human annotations and the real foggy dataset $\mathcal{D}_{\text{real}}^z$ from domain $z-1$ with labels inferred using model $\phi^{z-1}$:
\begin{equation}
\label{eq:ssl}
\min_{\phi^z}  \bigg(\sum_{\substack{(\vec{x}^\prime, \vec{y}^\prime)  \\ \in \mathcal{D}_{\text{syn}}^z}} L(\phi^z(\vec{x}^\prime),\vec{y}^\prime)  +  \lambda  \sum_{\substack{(\vec{x}^{\prime \prime}, \vec{y}^{\prime \prime}) \\ \in \mathcal{D}_{\text{real}}^z}} L(\phi^z(\vec{x}^{\prime \prime}),\vec{y}^{\prime \prime}) \bigg),
\end{equation}
where $L(.,.)$ is the cross entropy loss function and $\lambda=w\frac{R}{M}$ is a hyper-parameter balancing the weights of the two datasets, with $w$ serving as the relative weight of each real noisily labeled image compared to each synthetic labeled one and $R$ being the number of images in $\mathcal{D}_{\text{real}}^z$. We empirically set $w=1$ in our experiments, but an optimal value can be obtained via cross-validation if needed. The optimization of \eqref{eq:ssl} is implemented by generating a hybrid data stream and feeding it to a CNN for standard supervised training. More specifically, during training, training images are fetched from the randomly shuffled $\mathcal{D}_{\text{syn}}^z$ and $\mathcal{D}_{\text{real}}^z$ with a ratio  of $1:w$.   

We now describe the initialization stage of our method, which is also a variant of our method when no \emph{intermediate target domains} are used.
When $z=1$, we are in the clear-weather domain and the model $\phi^1$ is directly trained on a labeled real dataset, so no adaptation is required. For the case $z=2$, there are no real foggy images falling into the domain $z-1=1$ which is the clear-weather domain. In this case, the model $\phi^2$ is trained with the synthetic dataset $\mathcal{D}_{\text{syn}}^2$ only, as specified in Section~\ref{sec:learning:syn}. For the remaining steps from $z=3$ on, we iteratively apply the adaptation approach introduced above to adapt to domain $Z$, which constitutes the \emph{ultimate target domain} $T$.
In this work, we have experimented with three instantiations of our method for $Z=\{2,\,3,\,4\}$, which we name CMAda1, CMAda2 and CMAda3 respectively. The sequences of attenuation coefficients (fog densities) for the three versions are $(0,\,0.01)$, $(0,\,0.005,\,0.01)$ and $(0,\,0.0025,\,0.005,\,0.01)$ respectively.

Figure~\ref{fig:pipeline} provides an overview of CMAda2. Below, we summarize the complete operations of CMAda2 to further help understand the method.
With the chosen sequence of attenuation coefficients $(0,\,0.005,\,0.01)$, the whole pipeline of CMAda2 is as follows: 
\begin{enumerate}
    \item generate a synthetic foggy dataset with multiple versions of varying fog density; \label{itemone}
    \item train a model for fog density estimation on the dataset of step~\ref{itemone}; \label{itemtwo}
    \item rank the images in the real foggy dataset with the model of step~\ref{itemtwo} according to fog density; \label{itemthree}
    \item generate a dataset with light synthetic fog ($\beta=0.005$), and train a segmentation model on it; \label{itemfour}
    \item apply the segmentation model from step~\ref{itemfour} to the light-fog images of the real dataset (ranked lower in step~\ref{itemtwo}) to obtain noisy semantic labels; \label{itemfive}
    \item generate a dataset with dense synthetic fog ($\beta=0.01$); \label{itemsix}
    \item adapt the segmentation model from step~\ref{itemfour} to the union of the dense synthetic foggy dataset from step~\ref{itemsix} and the light real foggy one from step~\ref{itemfive} according to \eqref{eq:ssl}. \label{itemseven}
\end{enumerate}

\subsubsection{Discussion}

CMAda adapts segmentation models from clear weather to dense fog and is inspired by curriculum learning~\cite{curriculum:learning}, in the sense that we first solve easier tasks with our synthetic data, \ie{}fog density estimation and semantic scene understanding under light fog, and then acquire new knowledge from the already ``solved'' tasks in order to better tackle the harder task, \ie{}semantic scene understanding under dense real fog. CMAda also exploits the direct control of fog density for synthetic foggy images.  

This learning approach also bears resemblance to model distillation~\cite{hinton2015distilling,supervision:transfer} or imitation~\cite{model:compression,dai:metric:imitation}. The underpinnings of our proposed approach are the following: 1) in light fog objects are easier to recognize than in dense fog, hence models trained on synthetic data are more generalizable to real data in case both data sources contain light rather than dense fog; and 2) models trained on the source domain can be successfully applied to the target domain when the domain gap is small, hence incremental (curriculum) domain adaptation can better propagate semantic knowledge from the source domain to the ultimate target domain than single-step domain adaptation approaches. 

The goal of CMAda is to train a semantic segmentation model for the ultimate target domain $z$. The standard recipe is to record foggy images $\vec{x}^{\beta_z}$'s and then to manually create semantic labels $\vec{y}^{\beta_z}$'s for those foggy images so that the standard supervised learning can be applied. As discussed in Section~\ref{intro}, there is difficulty to apply this recipe to all adverse weather conditions because manual creation of $\vec{y}^{\beta_z}$'s is very time-consuming and expensive. To address this problem, this work develops methods to automatically create two proxy datasets for $(\vec{x}^{\beta_z}, \vec{y}^{\beta_z})$. The two proxies are defined in \eqref{eq:dataset:syn} and in \eqref{eq:dataset:real}. These two proxies reflect different and complementary characteristics of $(\vec{x}^{\beta_z}, \vec{y}^{\beta_z})$.  On the one hand, dense synthetic fog features a similar overall visibility obstruction to dense real fog, but includes artifacts. On the other hand, light real fog captures the true nonuniform and spatially varying structure of fog, but at a different density than dense fog. Learning jointly from both proxy datasets in CMAda reduces the influence of their individual drawbacks.

The CMAda pipeline presented in Section~\ref{sec:CMAda} is an extension of the original method proposed in the conference version~\cite{dense:SFSU:eccv18} of this paper from a two-stage approach to a general multiple-stage approach. CMAda is a stand-alone approach and already outperforms competing methods for SFSU, as discussed in Section~\ref{sec:experiments}. In the next section, we present an extension of CMAda, CMAda+, that further boosts performance.
       
\subsection{CMAda+ with Synthetic and Densified Real Fog}
\label{sec:CMAda+}

As defined in \eqref{eq:dataset:syn}, images in the synthetic training set $\mathcal{D}_{\text{syn}}^z$ have exactly the same fog density $\beta_z$ as images in the target domain $z$. Images in the real dataset $\mathcal{D}_{\text{real}}^z$, however, have lower fog density than the target fog density $\beta_z$, as defined in \eqref{eq:dataset:real}. While the lower fog density of the real training images facilitates the self-learning stream of CMAda with real foggy images, the remaining domain gap due to the disparity in fog density hampers finding a better solution. In Section~\ref{sec:fog:densification}, we present a method to densify fog in real foggy images so that it matches the desired fog density. The fog densification method is general and can be applied beyond CMAda. In Section~\ref{sec:fog:densification:dataset}, we use our fog densification method to upgrade the dataset defined in \eqref{eq:dataset:real} to a densified foggy dataset, which is used in CMAda+ along with the synthetic dataset to train the model $\phi^z$. 

\subsubsection{Fog Densification of a Real Foggy Scene}
\label{sec:fog:densification}

\begin{figure*}[tb]
  \centering
  \subfloat{\includegraphics[width=0.45\textwidth]{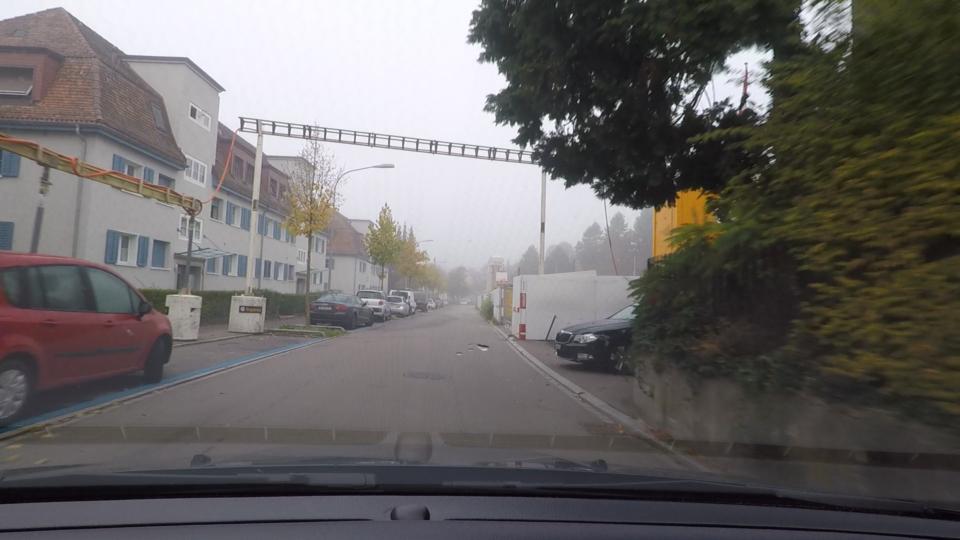}}
  \hfil
  \subfloat{\includegraphics[width=0.45\textwidth]{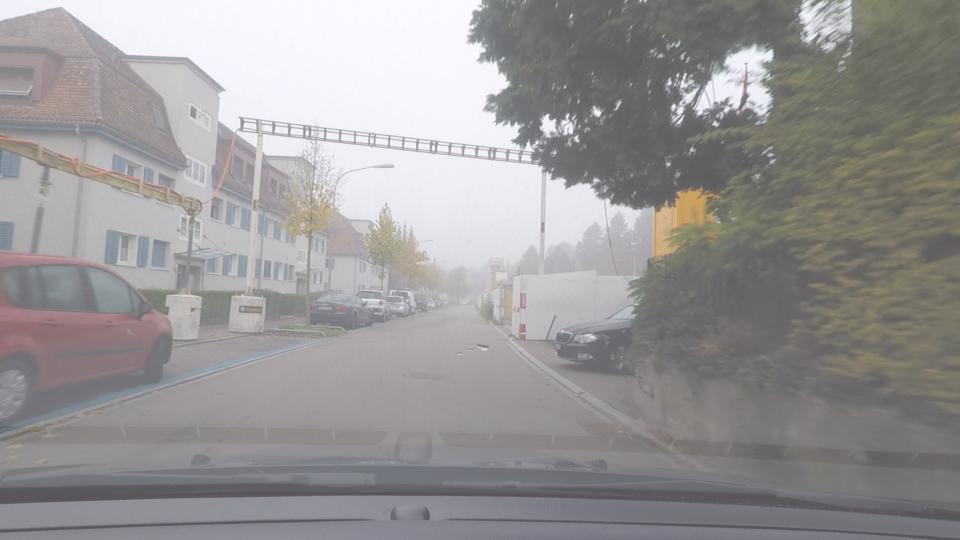}}\\
  \subfloat{\includegraphics[clip,width=0.14\textwidth,trim=25mm 65mm 25mm 72mm]{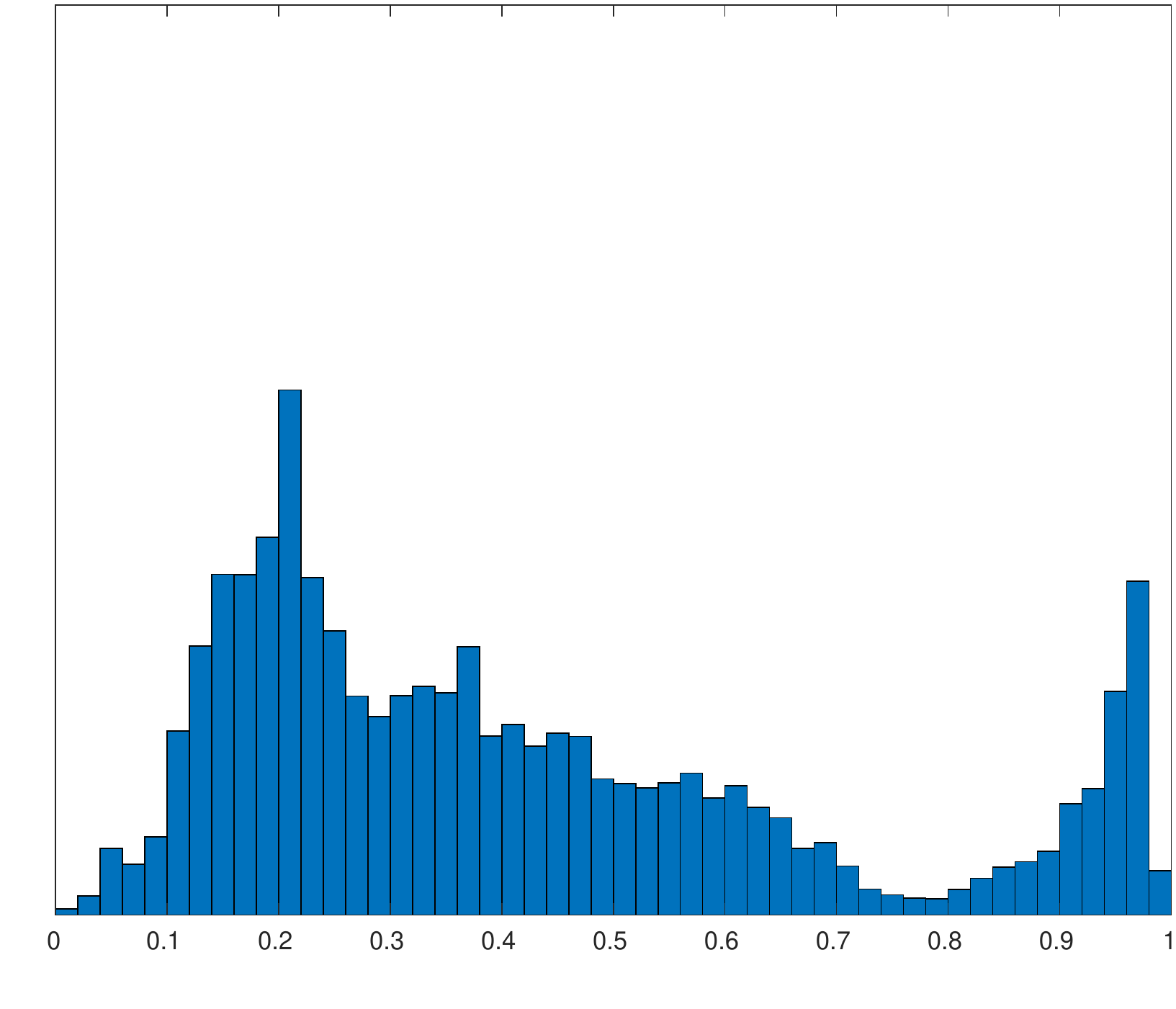}}
  \hfil
  \subfloat{\includegraphics[clip,width=0.15\textwidth,trim=25mm 70mm 25mm 70mm]{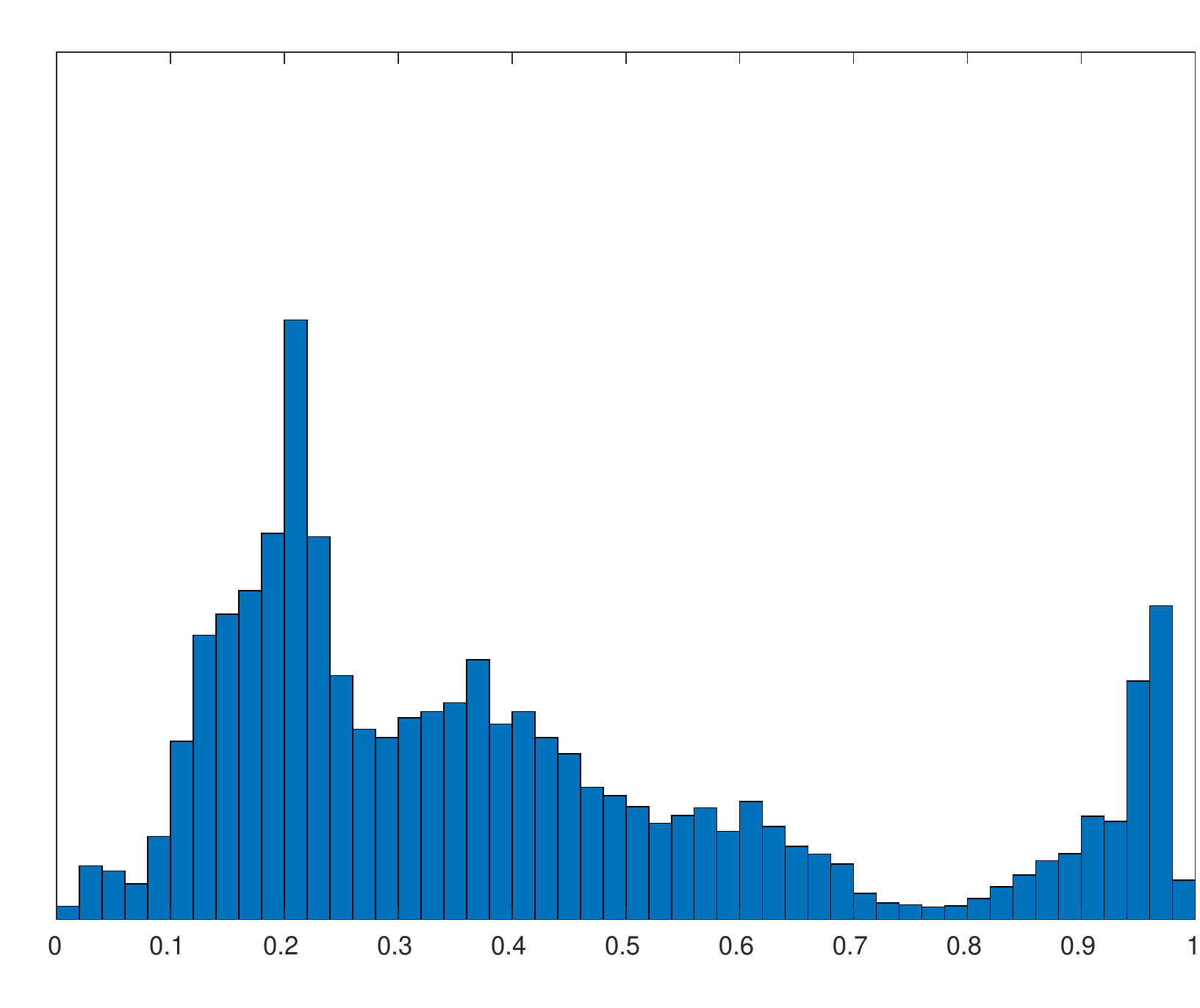}}
  \hfil
  \subfloat{\includegraphics[clip,width=0.15\textwidth,trim=25mm 70mm 25mm 70mm]{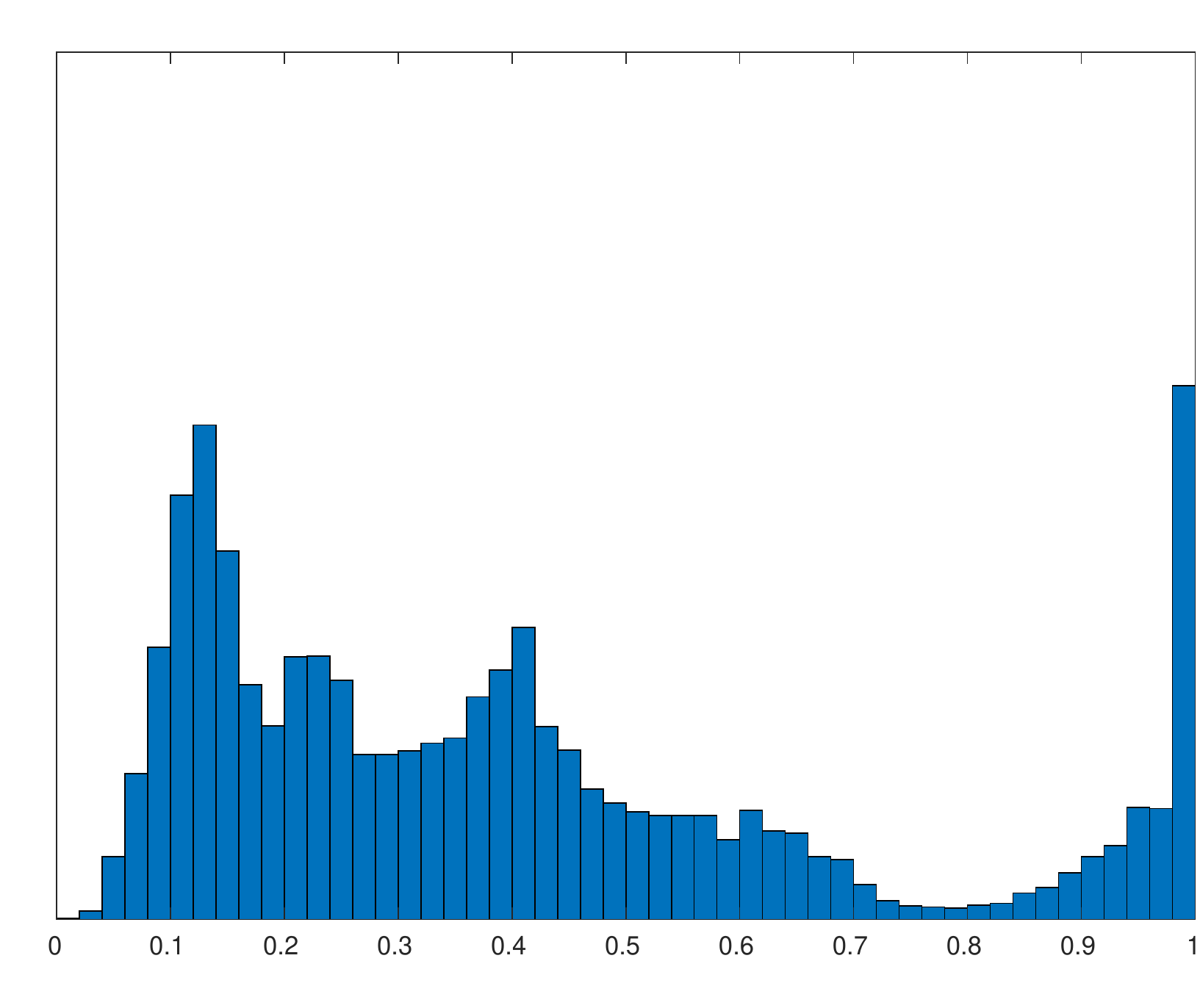}}
  \hfil
  \subfloat{\includegraphics[clip,width=0.15\textwidth,trim=25mm 70mm 25mm 70mm]{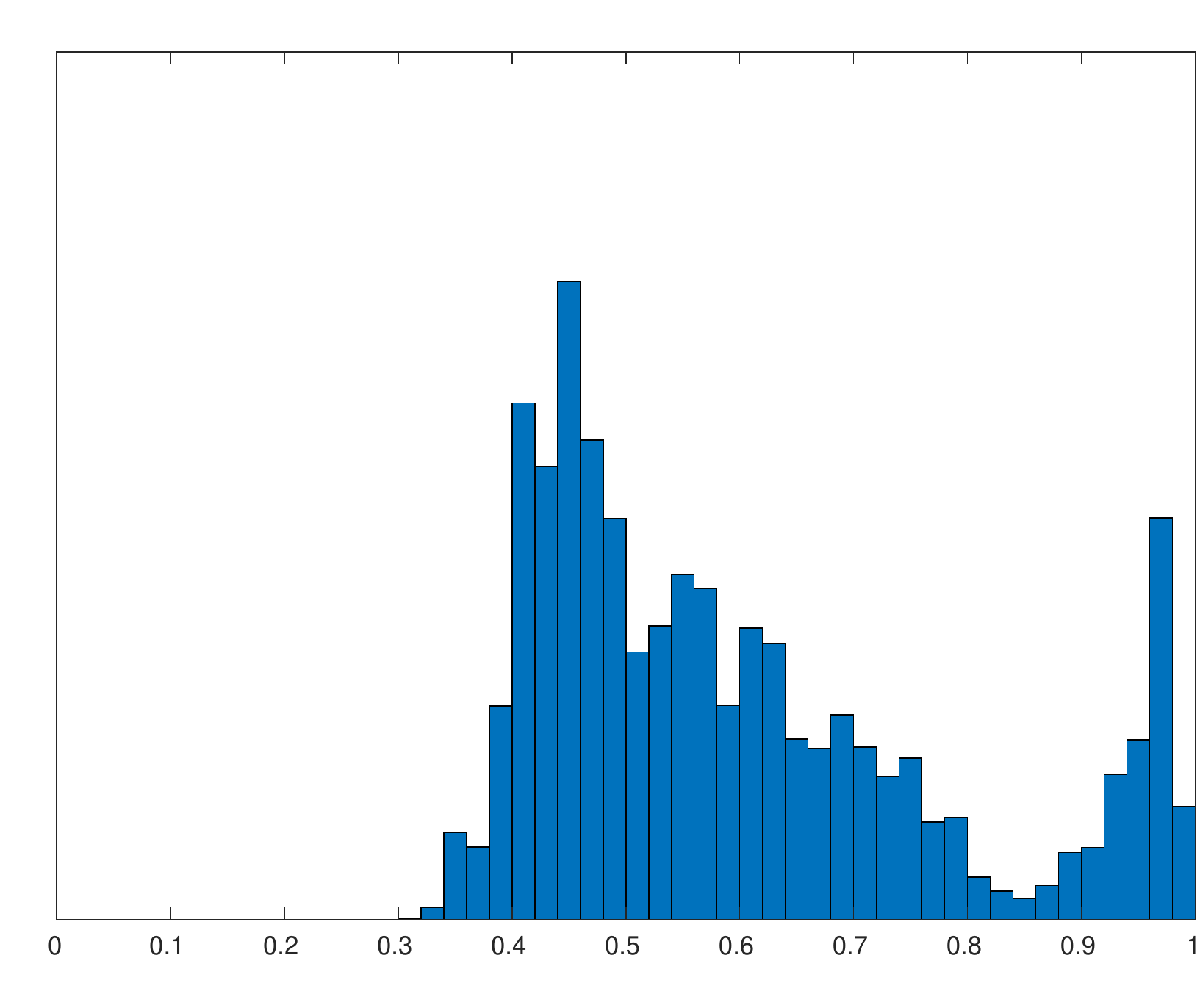}}
  \hfil
  \subfloat{\includegraphics[clip,width=0.15\textwidth,trim=25mm 109mm 67mm 57mm]{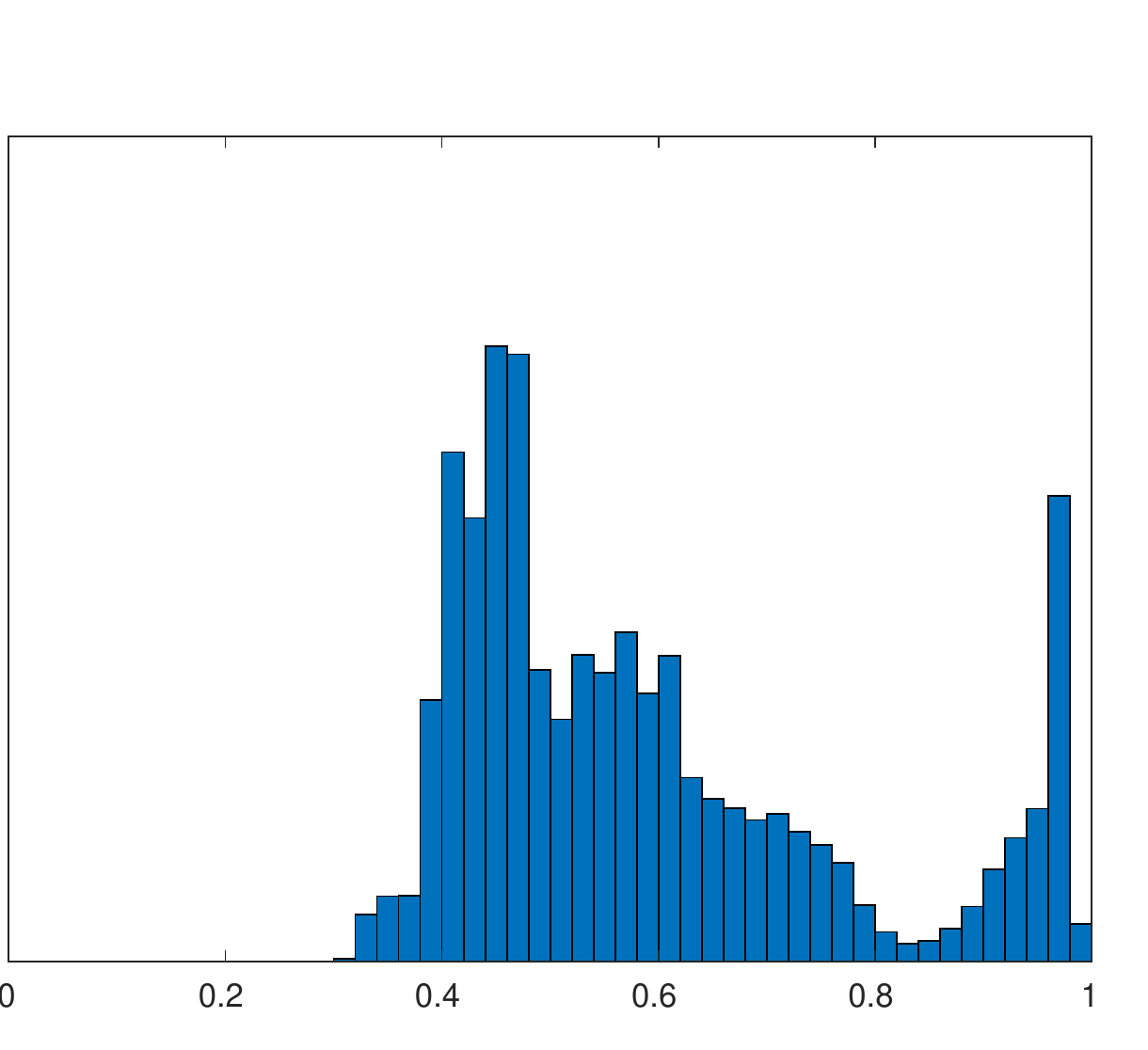}}
  \hfil
  \subfloat{\includegraphics[clip,width=0.15\textwidth,trim=25mm 109mm 67mm 57mm]{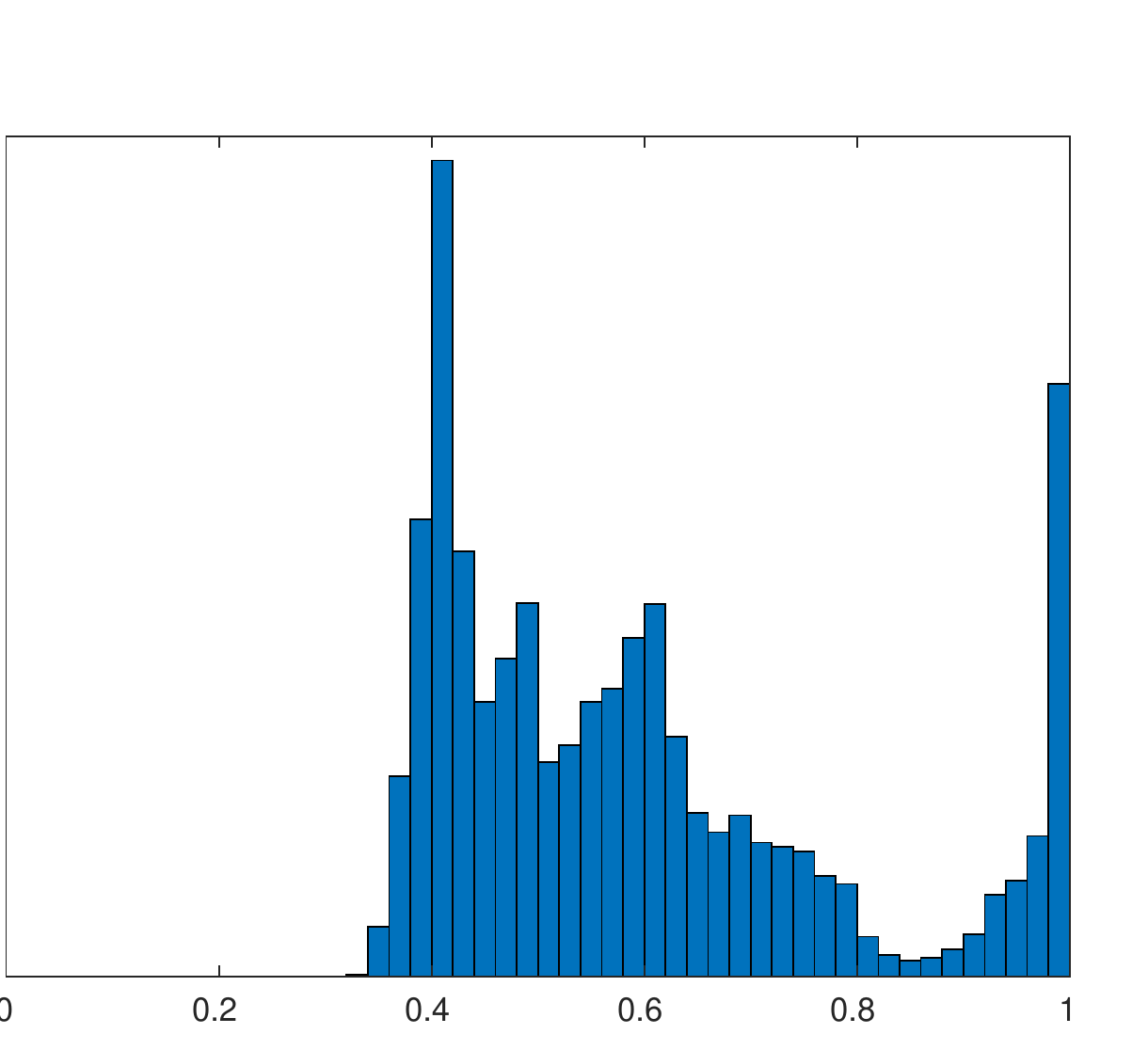}}
  \caption{Top row, left to right: example input image from \emph{Foggy Zurich} and synthesized output image with our fog densification. Bottom row, left to right: R, G, and B histogram of the input image, R, G, and B histogram of the output image}
  \label{fig:fog:densification:example}
\end{figure*}

We aim at synthesizing images with increased fog density compared to \emph{already foggy} real input images for which no depth information is available. In this way, we can generate multiple synthetic versions of each split of our real \emph{Foggy Zurich} dataset, where each synthetic version is characterized by a different, controlled range of fog densities, so that these densified foggy images can be leveraged in our curriculum adaptation. To this end, we utilize our fog density estimator and propose a simple yet effective approach for increasing fog density when no depth information is available for the input foggy image, by using the assumption of constant transmittance in the scene.

More formally, we denote the input real foggy image with $\mathbf{I}_l$ and assume that it can be expressed through the optical model \eqref{eq:fog:model}. Contrary to our fog simulation on clear-weather scenes in Section~\ref{sec:simulation}, the clear scene radiance $\mathbf{R}$ is unknown and the input foggy image $\mathbf{I}_l$ cannot be directly used as its substitute for synthesizing a foggy image $\mathbf{I}_d$ with increased fog density, as $\mathbf{I}_l$ does not correspond to clear weather. Since the scene distance $\ell$ which determines the transmittance through \eqref{eq:transmittance} is also unknown, we make the simplifying assumption that the transmittance map for $\mathbf{I}_l$ is globally constant, i.e.
\begin{equation} \label{eq:const:transmittance}
t(\mathbf{x}) = t_l,
\end{equation}
and use the statistics for scene distance $\ell$ computed on Cityscapes, which features depth maps, to estimate $t_l$. By using the distance statistics from Cityscapes, we implicitly assume that the distribution of distances of Cityscapes roughly matches that of our \emph{Foggy Zurich} dataset, which is supported by the fact that both datasets contain similar, road scenes. In particular, we apply our fog density estimator on $\mathbf{I}_l$ to get an estimate $\beta_l$ of the input attenuation coefficient. The values for scene distance $\ell$ of all pixels in Cityscapes are collected into a histogram $\mathcal{H} = \{(\ell_i,p_i):\,i=1,\,\dots,\,N\}$ with $N$ distance bins, where $\ell_i$ are the bin centers and $p_i$ are the relative frequencies of the bins. We use each bin center as representative of all samples in the bin and compute $t_l$ as a weighted average of the transmittance values that correspond to the different bins through \eqref{eq:transmittance}:
\begin{equation} \label{eq:transmittance:weighted}
t_l = \sum_{i=1}^N p_i \exp\left(-\beta_l \ell_i\right).
\end{equation}

The calculation of $t_l$ via \eqref{eq:transmittance:weighted} enables the estimation of the clear scene radiance $\mathbf{R}$ by re-expressing \eqref{eq:fog:model} for $\mathbf{I}_l$ when \eqref{eq:const:transmittance} holds as
\begin{equation} \label{eq:clear:radiance}
\mathbf{R}(\mathbf{x}) = \frac{\mathbf{I}_l(\mathbf{x}) - \mathbf{L}}{t_l} + \mathbf{L}.
\end{equation}
The globally constant atmospheric light $\mathbf{L}$ which is involved in \eqref{eq:clear:radiance} is estimated in the same way as in Section~\ref{sec:simulation:rest}.

For the output densified foggy image $\mathbf{I}_d$, we select a target attenuation coefficient $\beta_d > \beta_l$ and again estimate the corresponding global transmittance value $t_d$ similarly to \eqref{eq:transmittance:weighted}, this time plugging $\beta_d$ into the formula. The output image $\mathbf{I}_d$ is finally computed via \eqref{eq:fog:model} as
\begin{equation} \label{eq:densified:definition}
\mathbf{I}_d(\mathbf{x}) = \mathbf{R}(\mathbf{x})t_d + \mathbf{L}\left(1 - t_d\right).
\end{equation}
If we substitute $\mathbf{R}$ in \eqref{eq:densified:definition} using \eqref{eq:clear:radiance}, the output image is expressed only through $t_l$, $t_d$, the input image $\mathbf{I}_l$ and atmospheric light $\mathbf{L}$ as
\begin{align}
\mathbf{I}_d(\mathbf{x}) &{=}\;\mathbf{I}_l(\mathbf{x}) + \frac{t_d - t_l}{t_l}\left(\mathbf{I}_l(\mathbf{x}) - \mathbf{L}\right) \nonumber\\
&{=}\;\frac{t_d}{t_l}\mathbf{I}_l(\mathbf{x}) + \left(1 - \frac{t_d}{t_l}\right)\mathbf{L}. \label{eq:densified:analysis}
\end{align}
Equation \eqref{eq:densified:analysis} implies that our fog densification method can bypass the explicit calculation of the clear scene radiance $\mathbf{R}$ in \eqref{eq:clear:radiance}, as the output image does not depend on $\mathbf{R}$. In this way, we completely avoid dehazing our input foggy image as an intermediate step, which would pose challenges as it constitutes an inverse problem, and reduce the inference problem just to the estimation of the attenuation coefficient by assuming a globally constant transmittance. Moreover, \eqref{eq:densified:analysis} implies that the change in the value of a pixel $\mathbf{I}_d(\mathbf{x})$ with respect to $\mathbf{I}_l(\mathbf{x})$ is linear in the difference $\mathbf{I}_l(\mathbf{x}) - \mathbf{L}$. This means that distant parts of the scene, where $\mathbf{I}_l(\mathbf{x}) \approx \mathbf{L}$, are not modified significantly in the output, \ie{}$\mathbf{I}_d(\mathbf{x}) \approx \mathbf{I}_l(\mathbf{x})$. On the contrary, our fog densification modifies the appearance of those parts of the scene which are closer to the camera and shifts their color closer to that of the estimated atmospheric light irrespective of their exact distance from the camera. This can be observed in the example of Figure~\ref{fig:fog:densification:example}, where the closer parts of the input scene such as the red car on the left and the vegetation on the right have brighter colors in the synthesized output. The overall shift to brighter colors is verified by the accompanying RGB histograms of the input and output images in Figure~\ref{fig:fog:densification:example}.

\subsubsection{Fog Densification of a Real Foggy Dataset}
\label{sec:fog:densification:dataset}
When applying our fog densification to an entire dataset in the context of CMAda+, a simple choice is to specify the same target fog density $\beta_z$ for all images in the dataset. This may completely close the domain gap due to different fog density, but it ignores the variability of the true fog density across different images in the dataset and introduces other domain discrepancies, as our fog densification makes simplifying assumptions. Thus, we propose to define the target fog density independently for each input image.

Given the dataset $\mathcal{D}_{\text{real}}^z$ defined in \eqref{eq:dataset:real}, instead of mapping all $\beta_l \in [0, \beta_{z-1}]$ to $\beta_d = \beta_z$, we choose to perform a linear mapping from $[0, \beta_{z-1}]$  to $[\beta_{z-1}, \beta_{z}]$. In particular, given a real foggy image with its estimated attenuation coefficient $\beta_l \in [0, \beta_{z-1}]$, the target attenuation coefficient is determined as 
\begin{equation}
    \beta_d=\beta_{z-1}+\frac{\beta_l (\beta_z-\beta_{z-1})}{\beta_{z-1}}.
\end{equation}

Using $\psi_{\beta_l \rightarrow \beta_d}(\vec{x}_n)$ to indicate the densified image for $\vec{x}_n$, the densified real foggy dataset for CMAda+ at step $z$ is 
\begin{equation}
    \label{eq:dataset:real+}
    \mathcal{D}_{\text{real}}^z=\{(\psi_{\beta_l \rightarrow \beta_d}(\vec{x}_n), \hat{\vec{y}}_n^{z-1}) \: | \: f(\vec{x}_n)  \leq \beta_{z-1}\}_{n=1}^N.
\end{equation}
This densified dataset is then used in CMAda+ for training, along with the synthetic dataset defined in \eqref{eq:dataset:syn}, based on the same formulation \eqref{eq:ssl} as CMAda.

\subsection{Semantic Scene Understanding in Multiple Weather Conditions} 
\label{sec:model:selection:fusion}
In Section~\ref{sec:CMAda} and Section~\ref{sec:CMAda+}, specialized approaches have been developed for semantic scene understanding under fog. However, in real world applications weather conditions change constantly, \eg{}the weather can change from foggy to sunny or vice versa at any time. We argue that semantic scene understanding methods need to be robust and adaptive to these changes. With this aim, we propose Model Selection, a method for selecting the appropriate model depending on the encountered weather condition.

\subsubsection{Model Selection} 
\label{sec:model:selection} 
Our method uses two expert models, one specialized for clear weather and the other for fog. In particular, a two-class classifier is trained to distinguish \emph{clear weather} from \emph{fog}, with images from the Cityscapes dataset used as samples of the former class and images from three versions of our \emph{Foggy Cityscapes-DBF} dataset with attenuation coefficients $0.005$, $0.01$, and $0.02$ as samples of the latter class. We select AlexNet~\cite{alexnet} as the architecture of this classifier.

Denoting the semantic segmentation model specialized for fog by $\phi^Z$, the respective model for clear weather by $\phi^1$, and the aforementioned classifier by $g$, the semantic labels of a test image $\vec{x}$ are obtained through
\begin{equation}
    \hat{\vec{y}} = \begin{cases}
     \phi^1(\vec{x}), & \text{if $g(\vec{x})=1$},\\
     \phi^Z(\vec{x}) & \text{otherwise,}
  \end{cases}
\end{equation}
where label $1$ indicates the \emph{clear weather} class and label $0$ indicates \emph{fog}.  

The method is not limited to these two conditions and can be directly generalized to handle multiple adverse conditions, such as \emph{rain} or \emph{snow}.

\section{The Foggy Zurich Dataset}
\label{sec:dataset}

We present the \emph{Foggy Zurich} dataset, which comprises $3808$ images depicting foggy road scenes in the city of Zurich and its suburbs. We provide annotations for semantic segmentation for $40$ of these scenes that contain dense fog.

\subsection{Data Collection}

\emph{Foggy Zurich} was collected during multiple rides with a car inside the city of Zurich and its suburbs using a GoPro~Hero~5 camera. We recorded four large video sequences, and extracted video frames corresponding to those parts of the sequences where fog is (almost) ubiquitous in the scene at a rate of one frame per second. The extracted images are manually cleaned by removing the duplicates (if any), resulting in $3808$ foggy images in total. The resolution of the frames is $1920 \times 1080$ pixels. We mounted the camera inside the front windshield, since we found that mounting it outside the vehicle resulted in significant deterioration in image quality due to blurring artifacts caused by dew.

\begin{figure}
  \centering
  \subfloat[Camera inside windshield]{\includegraphics[width=0.45\textwidth]{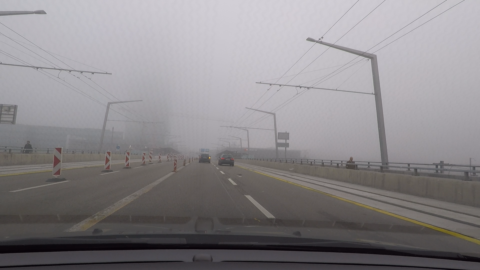}\label{fig:windshield:inside}}
  \hfil
  \subfloat[Camera outside windshield]{\includegraphics[width=0.45\textwidth]{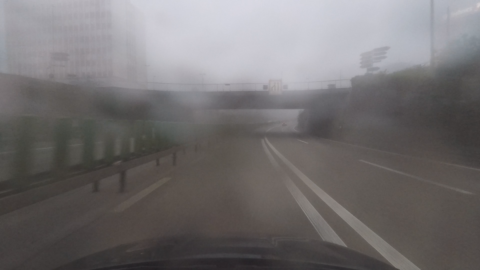}\label{fig:windshield:outside}}
  \caption{Comparison of images taken in fog with the camera mounted \protect\subref{fig:windshield:inside} inside and \protect\subref{fig:windshield:outside} outside the front windshield of the vehicle. We opt for the former configuration for collecting \emph{Foggy Zurich}}
  \label{fig:windshield}
\end{figure}

In particular, the small water droplets that compose fog condense and form dew on the surface of the lens very shortly after the vehicle starts moving, which causes severe blurring artifacts and contrast degradation in the image, as shown in Figure~\ref{fig:windshield}\subref{fig:windshield:outside}. On the contrary, mounting the camera inside the windshield, as we did when collecting \emph{Foggy Zurich}, prevents these blurring artifacts and affords much sharper images, to which the windshield surface incurs minimal artifacts, as shown in Figure~\ref{fig:windshield}\subref{fig:windshield:inside}.

\subsection{Annotation of Images with Dense Fog}
\label{sec:dataset:annotations}

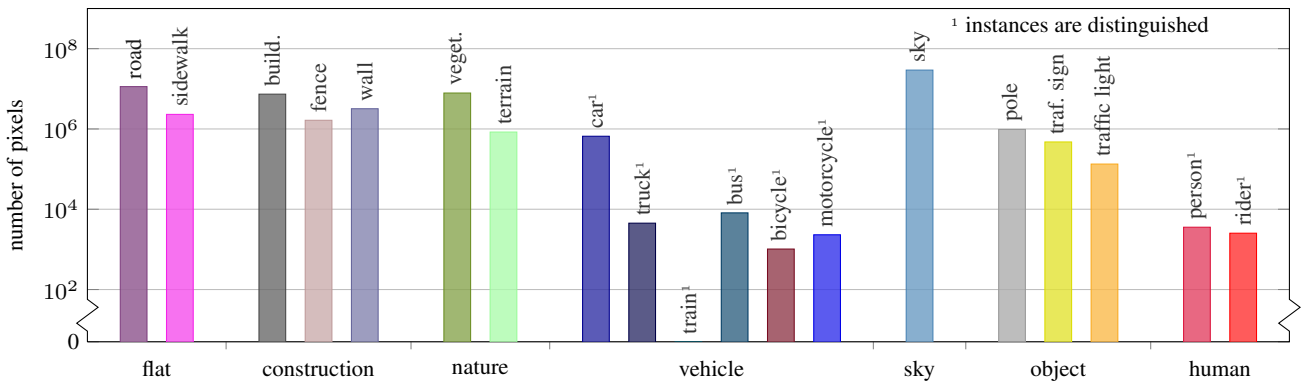
\begin{figure*}[tb]
    \centering
    \begin{tikzpicture}
    \tikzstyle{every node}=[font=\small]
    \begin{axis}[
            ybar,
            ymode=log,
            width=\textwidth,
            height=6cm,
            xmin=0,
            xmax=26,
            ymin=5,
            ymax=1e9,
            axis y discontinuity=crunch,
            ymajorgrids=true,
            ylabel={number of pixels},
      		ytick={5,1e2,1e4,1e6,1e8},
      		yticklabels={0,$10^2$,$10^4$,$10^6$,$10^8$,},
            xtick={1.5,5,8.5,13.5,18,21,24.5},
            minor xtick={3,7,10,17,19,23},
            xticklabels = {
                flat,
                construction,
                nature,
                vehicle,
                sky,
                object,
                human,
            },
            major x tick style = {opacity=0},
            minor x tick num = 1,
            xtick pos=left,
            every node near coord/.append style={
                    anchor=west,
                    rotate=90,
                    font=\footnotesize,
            }
            ]

    \addplot[bar shift=0pt,draw=road,          fill opacity=0.9,fill=road!80!white           , nodes near coords=road                 ] plot coordinates{ ( 1,     11348429  ) };
    \addplot[bar shift=0pt,draw=sidewalk,      fill opacity=0.8,fill=sidewalk!80!white       , nodes near coords=sidewalk             ] plot coordinates{ ( 2,     2305675   ) };

    \addplot[bar shift=0pt,draw=building,      fill opacity=0.8,fill=building!80!white       , nodes near coords=build.               ] plot coordinates{ ( 4,     7359559   ) };
    \addplot[bar shift=0pt,draw=fence,         fill opacity=0.8,fill=fence!80!white          , nodes near coords=fence                ] plot coordinates{ ( 5,     1644065   ) };
    \addplot[bar shift=0pt,draw=wall,          fill opacity=0.8,fill=wall!80!white           , nodes near coords=wall                 ] plot coordinates{ ( 6,     3192243   ) };

    \addplot[bar shift=0pt,draw=vegetation,    fill opacity=0.8,fill=vegetation!80!white     , nodes near coords=veget.               ] plot coordinates{ ( 8,    7852510  ) };
    \addplot[bar shift=0pt,draw=terrain,       fill opacity=0.8,fill=terrain!80!white        , nodes near coords=terrain              ] plot coordinates{ ( 9,    837299   ) };

    \addplot[bar shift=0pt,draw=car,           fill opacity=0.8,fill=car!80!white            , nodes near coords=car\fnn{1}           ] plot coordinates{ ( 11,    658258   ) };
    \addplot[bar shift=0pt,draw=truck,         fill opacity=0.8,fill=truck!80!white          , nodes near coords=truck\fnn{1}         ] plot coordinates{ ( 12,    4500     ) };
    \addplot[bar shift=0pt,draw=train,         fill opacity=0.8,fill=train!80!white          , nodes near coords=train\fnn{1}         ] plot coordinates{ ( 13,    5        ) };
    \addplot[bar shift=0pt,draw=bus,           fill opacity=0.8,fill=bus!80!white            , nodes near coords=bus\fnn{1}           ] plot coordinates{ ( 14,    8122     ) };
    \addplot[bar shift=0pt,draw=bicycle,       fill opacity=0.8,fill=bicycle!80!white        , nodes near coords=bicycle\fnn{1}       ] plot coordinates{ ( 15,    1020     ) };
    \addplot[bar shift=0pt,draw=motorcycle,    fill opacity=0.8,fill=motorcycle!80!white     , nodes near coords=motorcycle\fnn{1}    ] plot coordinates{ ( 16,    2311     ) };

    \addplot[bar shift=0pt,draw=sky,           fill opacity=0.8,fill=sky!80!white            , nodes near coords=sky                  ] plot coordinates{ ( 18,    29247575   ) };

    \addplot[bar shift=0pt,draw=pole,          fill opacity=0.8,fill=pole!80!white           , nodes near coords=pole                 ] plot coordinates{ ( 20,    971680    ) };
    \addplot[bar shift=0pt,draw=traffic sign,  fill opacity=0.8,fill=traffic sign!80!white   , nodes near coords=traf. sign         ] plot coordinates{ ( 21,    474260    ) };
    \addplot[bar shift=0pt,draw=traffic light, fill opacity=0.8,fill=traffic light!80!white  , nodes near coords=traffic light        ] plot coordinates{ ( 22,    133471    ) };

    \addplot[bar shift=0pt,draw=person,        fill opacity=0.8,fill=person!80!white         , nodes near coords=person\fnn{1}        ] plot coordinates{ ( 24,    3568    ) };
    \addplot[bar shift=0pt,draw=rider,         fill opacity=0.8,fill=rider!80!white          , nodes near coords=rider\fnn{1}         ] plot coordinates{ ( 25,    2530    ) };

    \node at (axis cs:18.5,1e9) [draw=none,anchor=north west,font=\footnotesize] {\fnn{1} instances are distinguished};

    \end{axis}
    \end{tikzpicture}
    \caption{Number of annotated pixels per class for \emph{Foggy Zurich-test}}
    \label{fig:dataset:stats:segmentation}
\end{figure*}

We use our fog density estimator presented in Section~\ref{sec:fog:density:estimation} to order all images in \emph{Foggy Zurich} according to fog density. Based on this ordering, we manually select 40 images with \emph{dense} fog and diverse visual scenes, and construct the test set of \emph{Foggy Zurich} therefrom, which we term \emph{Foggy Zurich-test}. The aforementioned selection is performed manually in order to guarantee that the test set has high diversity, which compensates for its relatively small size in terms of statistical significance of evaluation results. We annotate these images with fine pixel-level semantic annotations using the 19 evaluation classes of the Cityscapes dataset~\cite{Cityscapes}: \emph{road}, \emph{sidewalk}, \emph{building}, \emph{wall}, \emph{fence}, \emph{pole}, \emph{traffic light}, \emph{traffic sign}, \emph{vegetation}, \emph{terrain}, \emph{sky}, \emph{person}, \emph{rider}, \emph{car}, \emph{truck}, \emph{bus}, \emph{train}, \emph{motorcycle} and \emph{bicycle}. In addition, we assign the \emph{void} label to pixels which do not belong to any of the above 19 classes, or the class of which is uncertain due to the presence of fog. Every such pixel is ignored for semantic segmentation evaluation. Comprehensive statistics for the semantic annotations of \emph{Foggy Zurich-test} are presented in Figure~\ref{fig:dataset:stats:segmentation}. Furthermore, we note that individual instances of \emph{person}, \emph{rider}, \emph{car}, \emph{truck}, \emph{bus}, \emph{train}, \emph{motorcycle} and \emph{bicycle} are annotated separately, which additionally induces bounding box annotations for object detection for these 8 classes, although we focus solely on semantic segmentation in this paper. 

We also distinguish the semantic classes that occur frequently in \emph{Foggy Zurich-test}. These ``frequent'' classes are: \emph{road}, \emph{sidewalk}, \emph{building}, \emph{wall}, \emph{fence}, \emph{pole}, \emph{traffic light}, \emph{traffic sign}, \emph{vegetation}, \emph{sky}, and \emph{car}. When performing evaluation on \emph{Foggy Zurich-test}, we occasionally report the average score over this set of frequent classes, which feature plenty of examples, as a second metric to support the corresponding results.

Despite the fact that there exists a number of prominent large-scale datasets for semantic road scene understanding, such as KITTI~\cite{kitti}, Cityscapes~\cite{Cityscapes} and Mapillary Vistas~\cite{Mapillary}, most of these datasets contain few or even no foggy scenes, which can be attributed partly to the rarity of the condition of fog and the difficulty of annotating foggy images. Through manual inspection, we found that even Mapillary Vistas, which was specifically designed to also include scenes with adverse conditions such as snow, rain or nighttime, in fact contains very few images with fog, \ie{}in the order of 10 images out of 25000, with relatively more images depicting \emph{misty} scenes, which have $\text{MOR} \geq 1 \text{km}$, \ie{}significantly better visibility than foggy scenes~\cite{Federal:meteorological:handbook}.

To the best of our knowledge, the only previous dataset for semantic foggy scene understanding whose scale exceeds that of \emph{Foggy Zurich-test} is \emph{Foggy Driving}~\cite{SFSU_synthetic}, with 101 annotated images. However, most images in \emph{Foggy Driving} contain relatively light fog and most images with dense fog are annotated \emph{coarsely}. Compared to \emph{Foggy Driving}, \emph{Foggy Zurich} comprises a much greater number of high-resolution foggy images. Its larger, unlabeled part is highly relevant for unsupervised or semi-supervised approaches such as the one we have presented in Section~\ref{sec:CMAda}, while the smaller, labeled \emph{Foggy Zurich-test} set features \emph{fine} semantic annotations for the particularly challenging setting of dense fog, making a significant step towards evaluation of semantic segmentation models in this setting. In Table~\ref{table:dataset:stats:comparison}, we compare the overall annotation statistics of \emph{Foggy Zurich-test} to some of the aforementioned existing datasets; we note that the comparison involves a test set (\emph{Foggy Zurich-test}) and unions of training plus validation sets (KITTI and Cityscapes), which are much larger than the respective test sets. The comparatively lower number of humans and vehicles per image in \emph{Foggy Zurich-test} is not a surprise, as the condition of dense fog that characterizes the dataset discourages road transportation and reduces traffic.

\begin{table}[!tb]
    \centering
    \caption{Absolute and average number of annotated pixels, humans and vehicles for \emph{Foggy Zurich-test}, \emph{Foggy Driving}, KITTI and Cityscapes. Only the training and validation sets of KITTI and Cityscapes are considered. ``h/im'' stands for humans per image, ``v/im'' for vehicles per image and ``Foggy Zurich'' for \emph{Foggy Zurich-test}}
    \label{table:dataset:stats:comparison}
    \def\arraystretch{1}
    \setlength\tabcolsep{4pt}
    \begin{tabular*}{\linewidth}{l @{\extracolsep{\fill}} rrrrr}
    \toprule
    & Pixels & Humans & Vehicles & h/im & v/im\\
    \midrule
    Foggy Zurich & 66.1M & 27 & 135 & 0.7 & 3.4\\
    Foggy Driving & 72.8M & 290 & 509 & 2.9 & 5.0\\
    KITTI & 0.23G & 6.1k & 30.3k & 0.8 & 4.1\\
    Cityscapes & 9.43G & 24.0k & 41.0k & 7.0 & 11.8\\
    \bottomrule
    \end{tabular*}
\end{table}

In order to ensure a sound training and evaluation, we manually filter the unlabeled part of \emph{Foggy Zurich} and exclude from the resulting training sets that are used in CMAda those images which bear resemblance to any image in \emph{Foggy Zurich-test} with respect to the depicted scene.

\section{Experiments}
\label{sec:experiments}

Our model of choice for experiments on semantic segmentation with our CMAda pipeline is the state-of-the-art RefineNet~\cite{refinenet}. We use the publicly available \emph{RefineNet-res101-Cityscapes} model, which has been trained on the clear-weather training set of Cityscapes. In all experiments of this section, we use a constant learning rate of $5\times{}10^{-5}$ and mini-batches of size 1. Moreover, we compile all versions of \emph{Foggy Cityscapes-DBF} by applying our fog simulation (which is denoted by ``SDBF'' in the following for short) on the same \emph{refined} set of Cityscapes images that was used in~\cite{SFSU_synthetic} to compile Foggy Cityscapes-refined. This set comprises 498 training and 52 validation images; we use the former for training. In our experiments, we use the values $0.005$ and $0.01$ for attenuation coefficient $\beta$ both in SDBF and the fog simulation of~\cite{SFSU_synthetic} (denoted by ``SGF'') to generate different versions of \emph{Foggy Cityscapes-DBF} and Foggy Cityscapes respectively with varying fog density. 

\subsection{Performance on Foggy Scenes}
\label{sec:experiments:foggy}

For evaluation, we use 1) \emph{Foggy Zurich-test}, 2) a subset of \emph{Foggy Driving}~\cite{SFSU_synthetic} containing 21 images with dense fog, which we term \emph{Foggy Driving-dense}, and 3) the entire \emph{Foggy Driving}~\cite{SFSU_synthetic}.

We summarize our main experimental results in Table~\ref{table:experiments:main}. Overall, our method significantly improves the performance of semantic segmentation under dense fog compared to the original RefineNet model which has been trained on clear-weather images of Cityscapes. More specifically, we improve the performance (mIoU) from $34.6\%$ to $\mathbf{46.8\%}$ on \emph{Foggy Zurich-test} and from $35.8\%$ to $\mathbf{43.0\%}$ on \emph{Foggy Driving-dense}. With the new extensions, our fully-fledged CMAda3+ method significantly outperforms CMAda2, which was originally presented in the conference version of this paper~\cite{dense:SFSU:eccv18}.

It is worthwhile to mention that these improvements are achieved without using any extra human annotations on top of the original Cityscapes. Also, images in \emph{Foggy Driving} were taken by different cameras than the GoPro Hero 5 camera used for \emph{Foggy Zurich}, showing that CMAda also generalizes well to different sensors from that corresponding to the real training set of the method.

In the rest of Section~\ref{sec:experiments:foggy}, we analyze the effect of the individual components of our approach. This analysis demonstrates the benefit for semantic segmentation of real foggy scenes of: 1) our fog simulation for generating synthetic training data, 2) our fog density estimator against a state-of-the-art competing method, 3) combining our synthetic foggy data from \emph{Foggy Cityscapes-DBF} with unlabeled real data from \emph{Foggy Zurich} through our CMAda pipeline to adapt \emph{gradually} to dense real fog in multiple steps, and 4) using our fog densification method to further close the gap between light real fog and dense real fog. Finally, we provide some qualitative results.

\subsubsection{Benefit of Adaptation with Our Synthetic Fog}
\label{sec:experiments:synthetic}

\begin{table*}[tb]
\centering
\caption{Performance comparison on \emph{Foggy Zurich-test} (\emph{FZ}),  \emph{Foggy Driving-dense} (\emph{FDD}) and \emph{Foggy Driving} (\emph{FD}) of different variants of our CMAda pipeline as well as competing approaches, using with the mean intersection-over-union (mIoU) metric over all classes}
\label{table:experiments:main}
\bgroup
\def\arraystretch{1.2}
\setlength\tabcolsep{3pt} 
\begin{tabular}{@{}lccccccccccc@{}}
\toprule 
& Clear-weather & \multicolumn{2}{c}{Synthetic fog} &  {  } Real fog {  } & \multicolumn{2}{c}{Density Estimator}  & {  } \emph{FZ} { } &  { } \emph{FDD} { } & { } \emph{FD} { } \\
 & Cityscapes \cite{Cityscapes} & SGF \cite{SFSU_synthetic} & SDBF (ours)  & GoPro & FADE \cite{fog:density:15}  & Ours &   mIoU (\%)   &   mIoU (\%)   &   mIoU (\%) \\ \midrule
\textbf{Comparison} & & & & & & & \\  
 RefineNet \cite{refinenet} & \cmark &   &  &  &  &  & 34.6 &  35.8 & 44.3 \\ 
SFSU \cite{SFSU_synthetic} & \cmark & \cmark  &  &  &  & & 35.7  & 35.9 & 46.3\\
AdSegNet \cite{AdaptSegNet} & \cmark &  & \cmark  & \cmark &   &    & 25.0  & 15.8 &29.7 \\
\underline{CMAda2} \cite{dense:SFSU:eccv18} & \cmark &  & \cmark  & \cmark &   &  \cmark  & 42.9  & 37.3 & 48.5\\
\underline{CMAda3+} & \cmark &  & \cmark  & \cmark &   & \cmark   & \textbf{46.8}  & \textbf{43.0} & \textbf{49.8}    \\
\cline{1-10}
\textbf{Ablation Study} & & & & & & \\  
Baseline~\cite{refinenet} & \cmark &   &  &  &  &  & 34.6 &  35.8 & 44.3\\ \cdashline{1-10}
\multirow{4}{*}{CMAda1}
& \cmark  & \cmark  &        &   &       & &  35.7  & 35.9 & 46.3\\
& \cmark  &         & \cmark &   &       &   & 36.3  & 36.1 & 46.3 \\
\cdashline{2-10}
& \cmark   &   &  & \cmark & \cmark &  &     37.5    & 36.4 & 45.7 \\
& \cmark  &  &   & \cmark &   & \cmark &   38.9     &  36.6 & 46.0 \\
\cdashline{1-10}
\multirow{4}{*}{CMAda2}
& \cmark  & \cmark  &   & \cmark &  \cmark &    &  39.8  & 35.7 & 47.5\\
& \cmark  & \cmark  &   & \cmark & & \cmark   &  41.5  &  37.0 & 48.5\\ \cdashline{2-10}
& \cmark  & & \cmark  & \cmark &  \cmark &  &   40.6 &  35.5 & 47.7\\ 
& \cmark  & & \cmark  & \cmark & & \cmark  &  42.9  & 37.3 & 48.5 \\ 
\cdashline{1-10}
\underline{CMAda3} & \cmark  &    &  \cmark  & \cmark &   &  \cmark  &   43.7  &  40.6 & 48.9 \\
\cdashline{1-10}
\underline{CMAda2+} & \cmark &  & \cmark  & \cmark &   & \cmark   & 43.4  & 40.1 & \textbf{49.9}\\
\underline{CMAda3+} & \cmark &  & \cmark  & \cmark &   & \cmark   & \textbf{46.8}  & \textbf{43.0} & 49.8    \\
\bottomrule  \\	
\end{tabular}
\egroup
\end{table*}

\begin{figure*}
  \centering
  \subfloat{\includegraphics[width=0.19\textwidth]{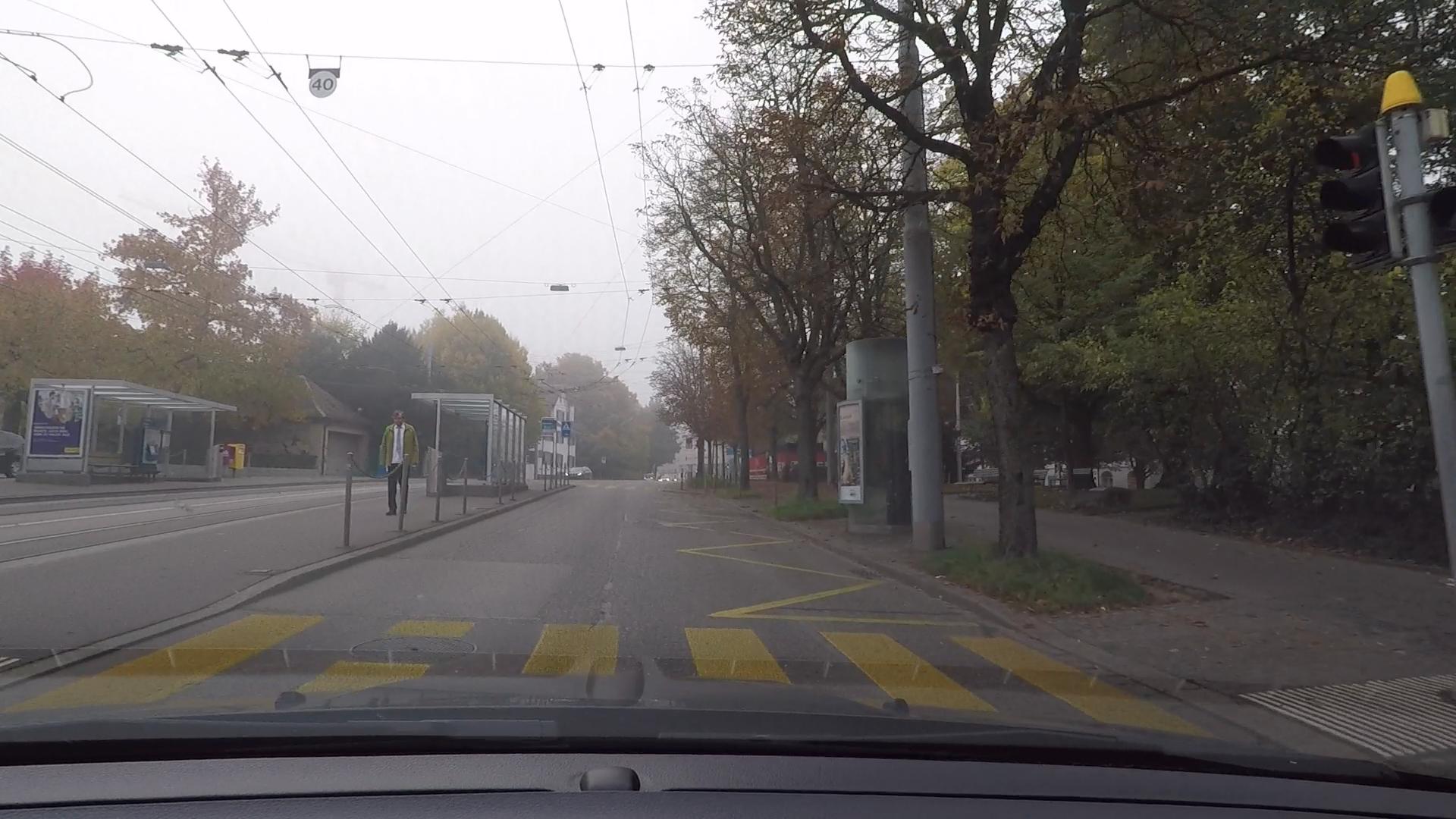}}
  \hfil
  \subfloat{\includegraphics[width=0.19\textwidth]{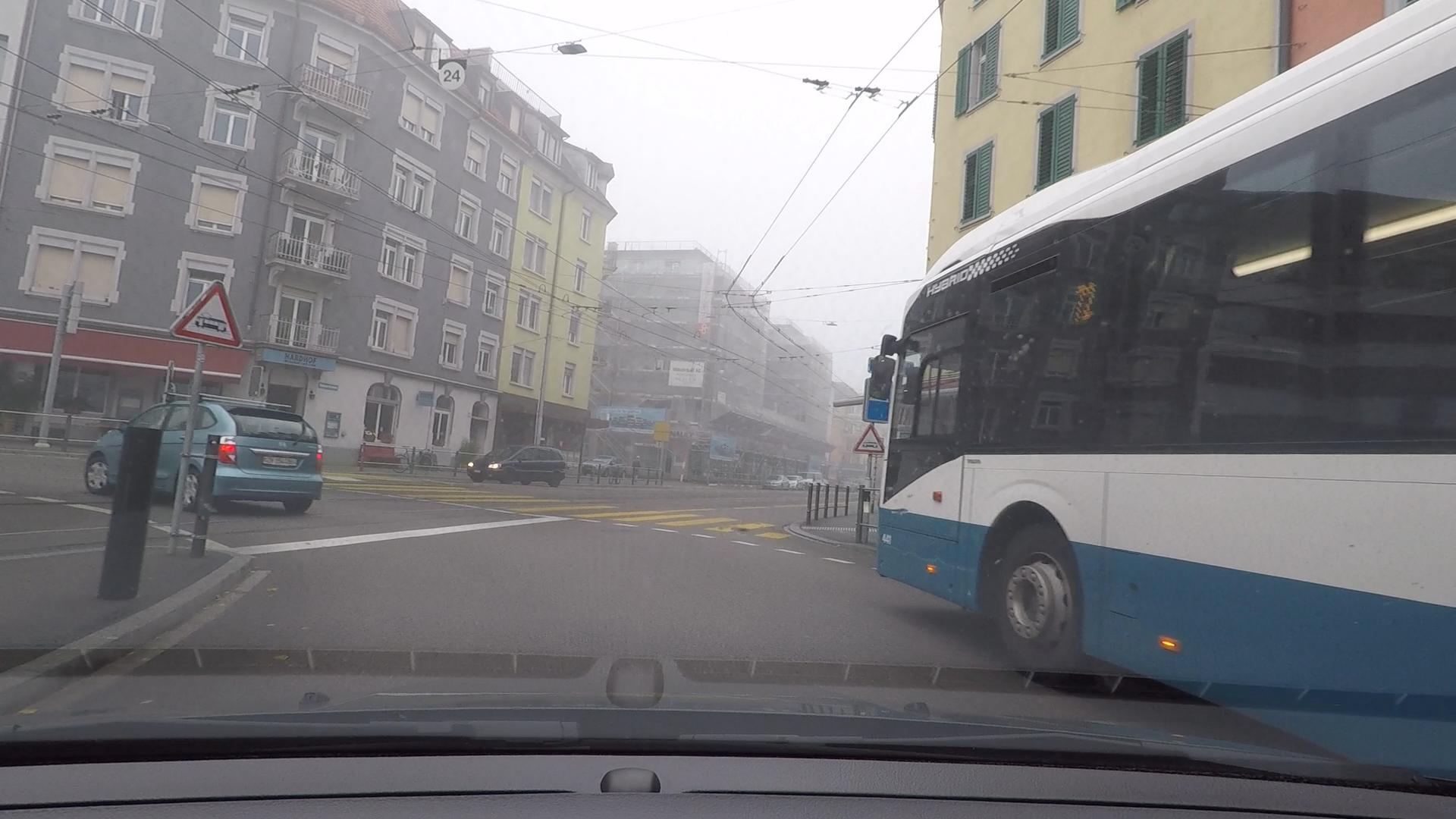}}
  \hfil
  \subfloat{\includegraphics[width=0.19\textwidth]{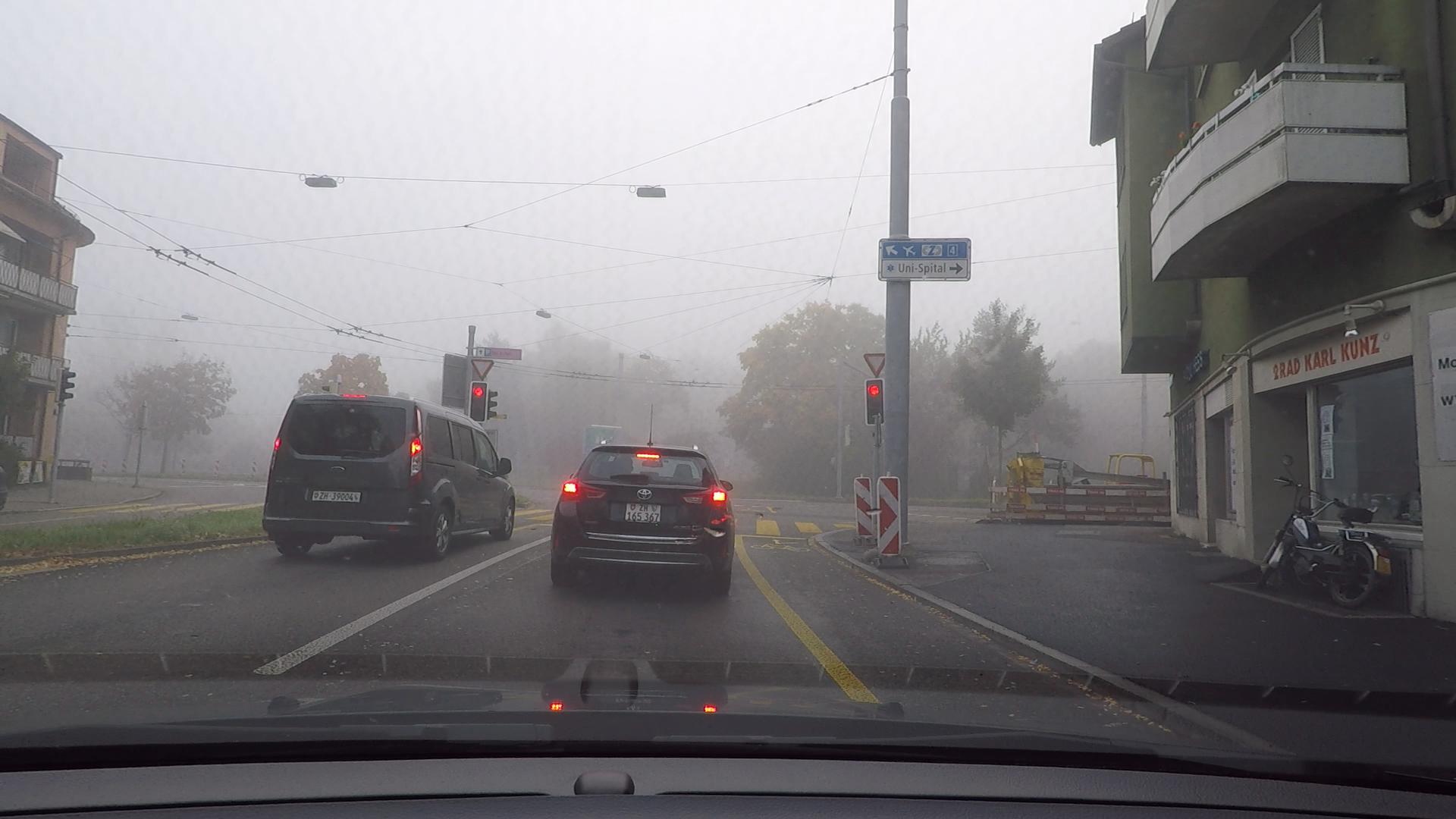}}
  \hfil
  \subfloat{\includegraphics[width=0.19\textwidth]{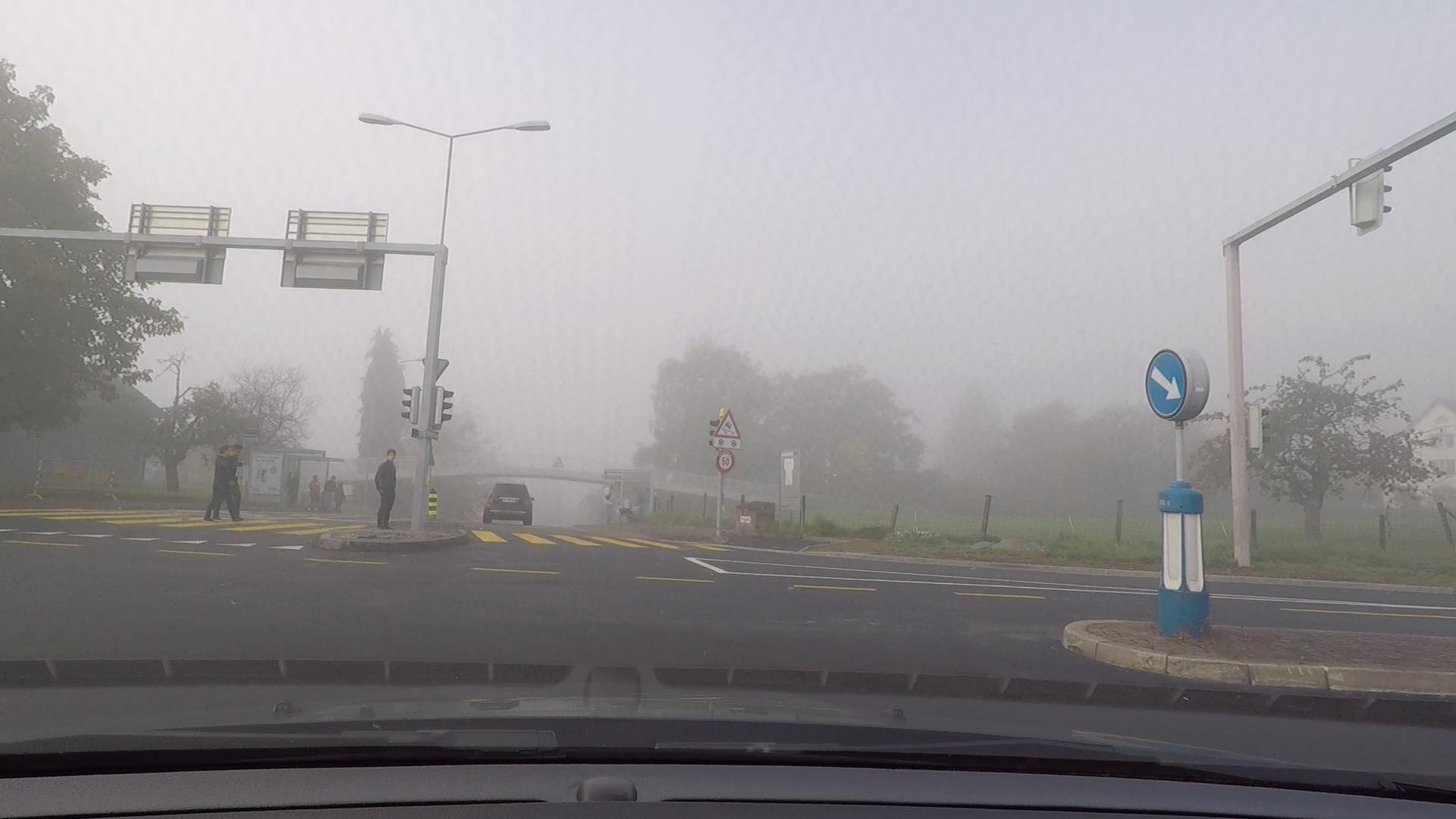}}
  \hfil
  \subfloat{\includegraphics[width=0.19\textwidth]{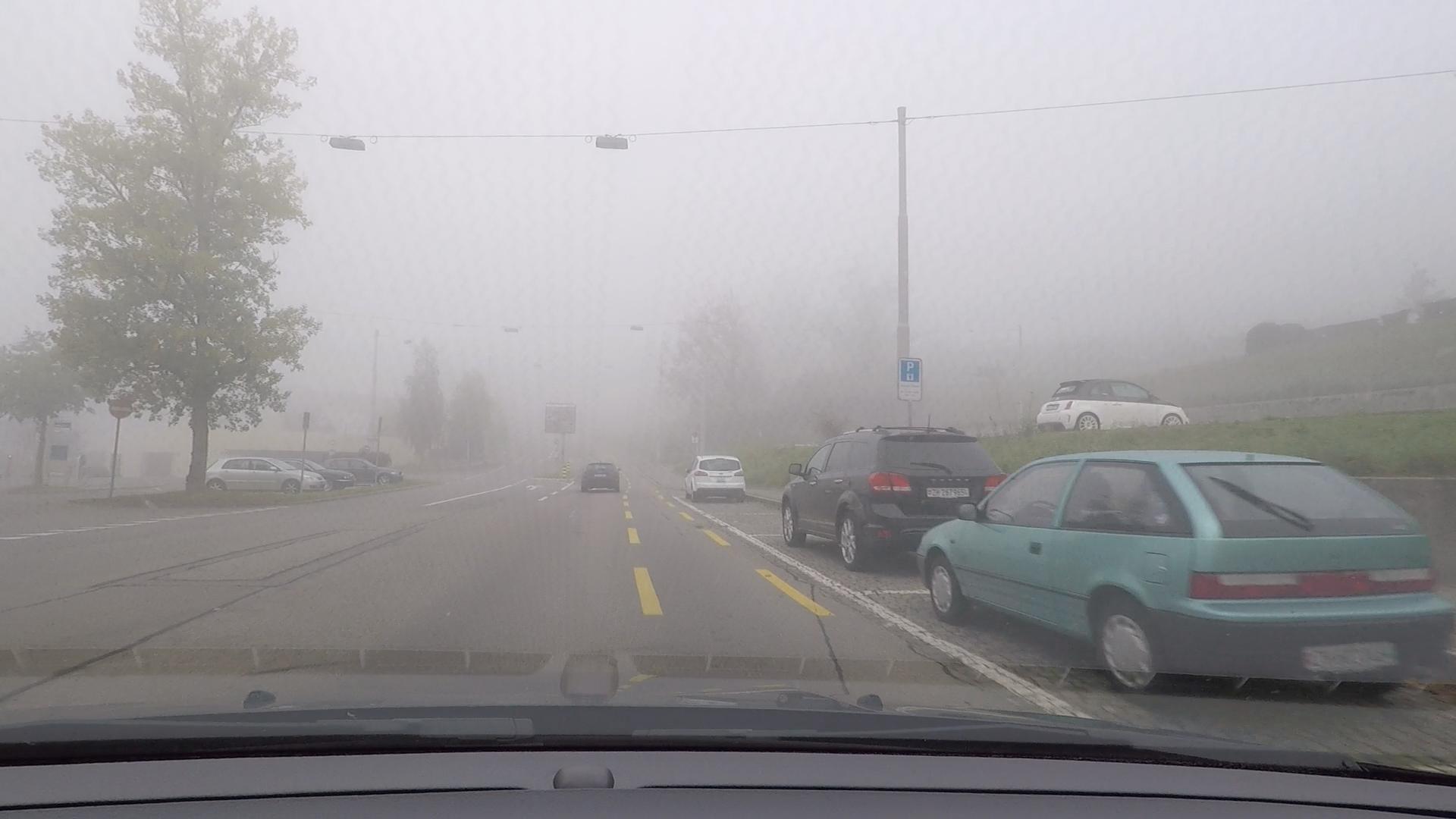}}\\
  \caption{Foggy images from \emph{Foggy Zurich}, sorted from left to right in ascending order with respect to estimated fog density using our estimator}  
  \label{fig:fog:density}
\end{figure*}

Our first segmentation experiment shows that our semantic-aware fog simulation (SDBF) performs competitively compared to the fog simulation of~\cite{SFSU_synthetic} (SGF) for generating synthetic data to adapt RefineNet to real dense fog. \emph{RefineNet-res101-Cityscapes} is fine-tuned on \emph{Foggy Cityscapes-DBF} and alternatively Foggy Cityscapes, both with attenuation coefficient $\beta=0.01$, for 8 epochs. The corresponding results in  Table~\ref{table:experiments:main} are presented in the top two rows under the group ``CMAda1''. Training on synthetic fog with either type of fog simulation helps to beat the baseline clear-weather RefineNet model on all three test sets, the improvement being more significant on \emph{Foggy Zurich-test} and \emph{Foggy Driving}. In addition, SDBF slightly outperforms SGF consistently.

\sloppy{Moreover, in all cases that both synthetic and real foggy data are used in the two-stage CMAda pipeline, corresponding to the rows of Table~\ref{table:experiments:main} grouped under ``CMAda2'', SDBF yields significantly higher segmentation performance on \emph{Foggy Zurich-test} compared to SGF, while the two methods are on a par on the other two sets.}

\subsubsection{Benefit of Our Fog Density Estimator on Real Data}
\label{sec:exp:fog:density}

\sloppy{The second component of the CMAda pipeline that we ablate is the fog density estimator. In particular, Table~\ref{table:experiments:main} includes results for the single-stage pipeline with adaptation on real images from the unlabeled part of \emph{Foggy Zurich} and the two-stage pipeline with adaptation on synthetic and real images from Foggy Cityscapes and \emph{Foggy Zurich} respectively, where the ranking of real images according to fog density is performed either with the method of~\cite{fog:density:15} or with our AlexNet-based fog density estimator described in Section~\ref{sec:fog:density:estimation}. In all experimental settings, our fog density estimator outperforms \cite{fog:density:15} significantly in terms of mIoU on all datasets. This fully lifts the need of manually designing features and labeling images for fog density estimation, as was done in ~\cite{fog:density:15}.}

For further verification of our fog density estimator, we conduct a user study on Amazon Mechanical Turk (AMT). In order to guarantee high quality, we only employ AMT Masters in our study and verify the answers via a Known Answer Review Policy. Each human intelligence task (HIT) comprises five image pairs to be compared: three pairs are the true query pairs with images from the real \emph{Foggy Zurich} dataset, and the rest two pairs contain synthetic fog of different densities and are used for validation. The participants are shown two images at a time, side by side, and are simply asked to choose the one which is more foggy. The query pairs are sampled based on the ranking results of our estimator. In order to avoid confusing cases, \ie{}two images of similar fog densities, the two images of each pair need to be ranked at least $20$ percentiles apart from each other by our estimator.

We have collected answers for $12000$ pairs in $4000$ HITs. The HITs are considered for evaluation only when both validation questions are correctly answered. $87\%$ of all HITs are valid for evaluation. On these $10400$ pairs, the agreement between our fog density estimator and human judgment is $89.3\%$. This high agreement confirms that fog density estimation is a relatively easier task which can be solved by using synthetic data, and the acquired knowledge can be further exploited for solving high-level tasks on foggy scenes. Figure~\ref{fig:fog:density} shows foggy images in ascending order of estimated fog density using our estimator.

\subsubsection{Benefit of Adaptation with Synthetic and Real Fog}
\label{sec:experiments:curriculum}

The main segmentation experiment showcases the effectiveness of our CMAda pipeline. \emph{Foggy Cityscapes-DBF} and Foggy Cityscapes~\cite{SFSU_synthetic} are the two alternatives for the synthetic foggy training sets in steps~\ref{itemfour} and \ref{itemsix} of the pipeline, corresponding to the two alternatives for fog simulation (SDBF and SGF respectively). \emph{Foggy Zurich} serves as the real foggy training set. We use the results of our fog density estimation to select 1556 images from \emph{Foggy Zurich} with light fog and name this set \emph{Foggy Zurich-light}. We implement CMAda2 by first fine-tuning RefineNet on \emph{Foggy Cityscapes-DBF} (alternatively Foggy Cityscapes) with $\beta=0.005$ for 6k iterations and then further fine-tuning it on the union of \emph{Foggy Cityscapes-DBF} (alternatively Foggy Cityscapes) with $\beta=0.01$ and \emph{Foggy Zurich-light}, where the latter set is labeled by the aforementioned initially adapted model. Two-stage curriculum adaptation to dense fog with synthetic and real data, which corresponds to the results in the rows that are grouped under ``CMAda2'' in Table~\ref{table:experiments:main}, consistently outperforms single-stage adaptation with either only synthetic or only real training data (``CMAda1''), irrespective of the selected fog simulation and fog density estimation methods. The combination of our fog simulation SDBF and our fog density estimator delivers the best result on all three test sets among all variants of CMAda2, improving upon the baseline RefineNet model on \emph{Foggy Zurich-test} by $8.3\%$. The same combination also provides a clear generalization benefit of $4.2\%$ against the baseline on \emph{Foggy Driving}, even though this dataset involves different camera sensors and scenes than \emph{Foggy Zurich}, which is the sole real-world dataset used in our training. 

We note that the significant performance benefit delivered by CMAda both on \emph{Foggy Zurich-test} and \emph{Foggy Driving} is not matched by the state-of-the-art domain-adversarial approach of~\cite{AdaptSegNet} for adaptation of semantic segmentation models, which we also trained both on our synthetic \emph{Foggy Cityscapes-DBF} set and our unlabeled real \emph{Foggy Zurich-light} set. This can be attributed to the fact that images captured under adverse conditions such as fog have large intra-domain variance as a result of poor visibility, effects of artificial lighting sources and motion blur. However, we believe that domain-adversarial approaches have the potential to be used for transferring knowledge to adverse weather domains.

\subsubsection{Benefit of Adaptation at Finer Scales}
\label{sec:benefit:finerscale}

We also experiment with the three-stage instantiation of CMAda, CMAda3, using the optimal configuration of all components of the pipeline based on the previous comparisons. Compared to CMAda2, CMAda3 adapts the semantic segmentation model at a finer scale, \ie{}1) from clear-weather to mist with synthetic misty data; 2) then to light fog with synthetic light foggy data and real misty data; and 3) finally to dense fog with synthetic dense foggy data and real light foggy data. The exact fog densities at each stage are defined in Section~\ref{sec:CMAda}. In particular, the extra stage compared to CMAda2 consists in labeling a split of \emph{Foggy Zurich} with very light estimated fog, which we term \emph{Foggy Zurich-light+}, via the clear-weather RefineNet model and using it in conjunction with \emph{Foggy Cityscapes-DBF} with $\beta=0.005$ to form the training set for the first stage of CMAda.

Including this extra stage affords higher segmentation performance on all three test sets as reported in row ``CMAda3'' of Table~\ref{table:experiments:main}, outperforming the respective best CMAda2 instance by $3.3\%$ on \emph{Foggy Driving-dense}. The improvement of CMAda3 over CMAda2 shows that our approach benefits from adaptation at finer scales, which is in line with the rationale of curriculum learning~\cite{curriculum:learning}. However, training for a large number of stages increases the computational cost significantly. Thus, selecting the ``optimal'' number of stages and the exact fog densities that correspond to the intermediate target domains needs further investigation and could be solved to some extent by cross-validation.

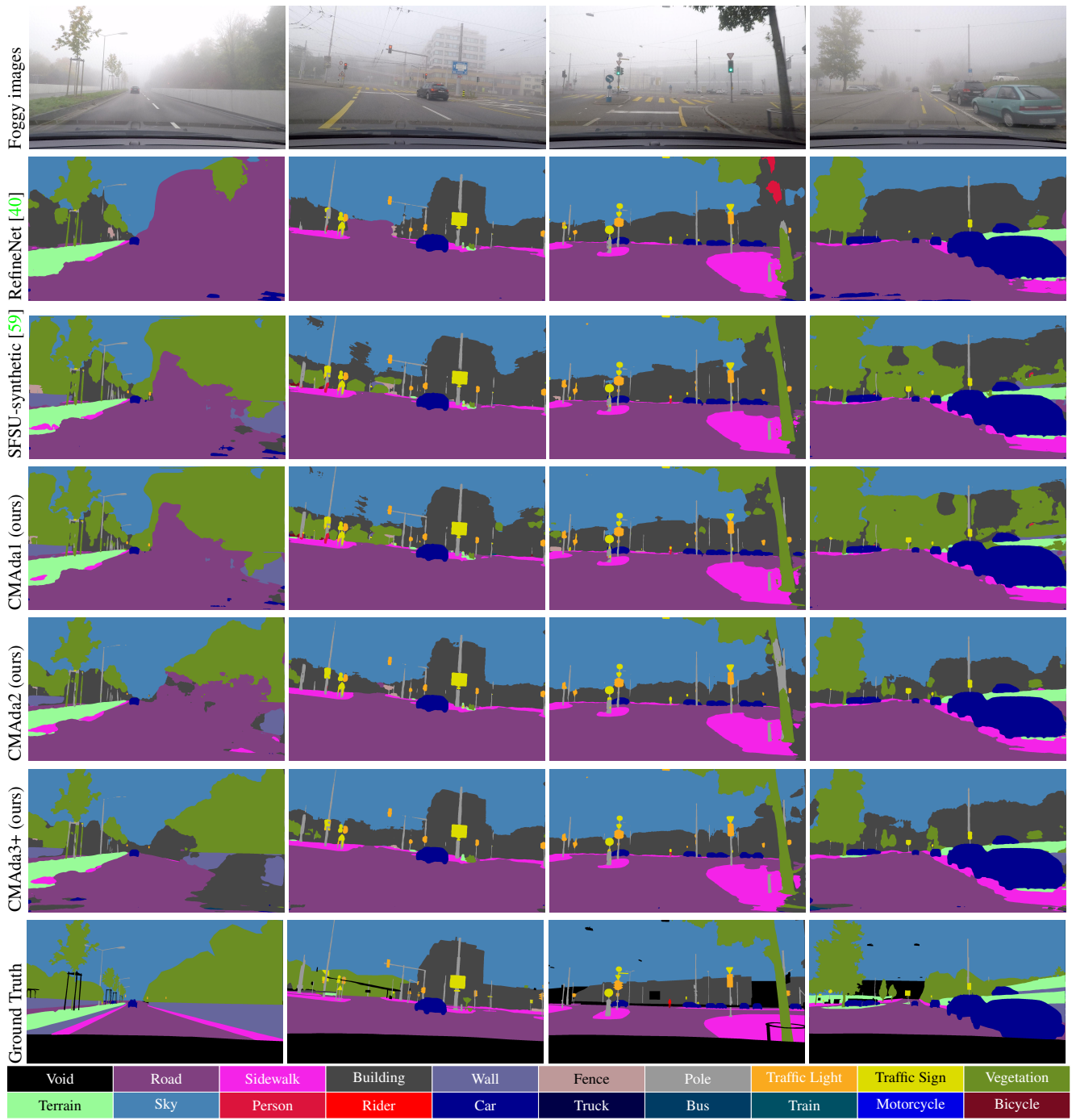
\begin{figure*}[!tb]
  \centering
  \addtocounter{subfigure}{-20}
    \subfloat{\rotatebox{90}{Foggy images}}  
  \hfil
  \subfloat{\includegraphics[width=0.24\textwidth]{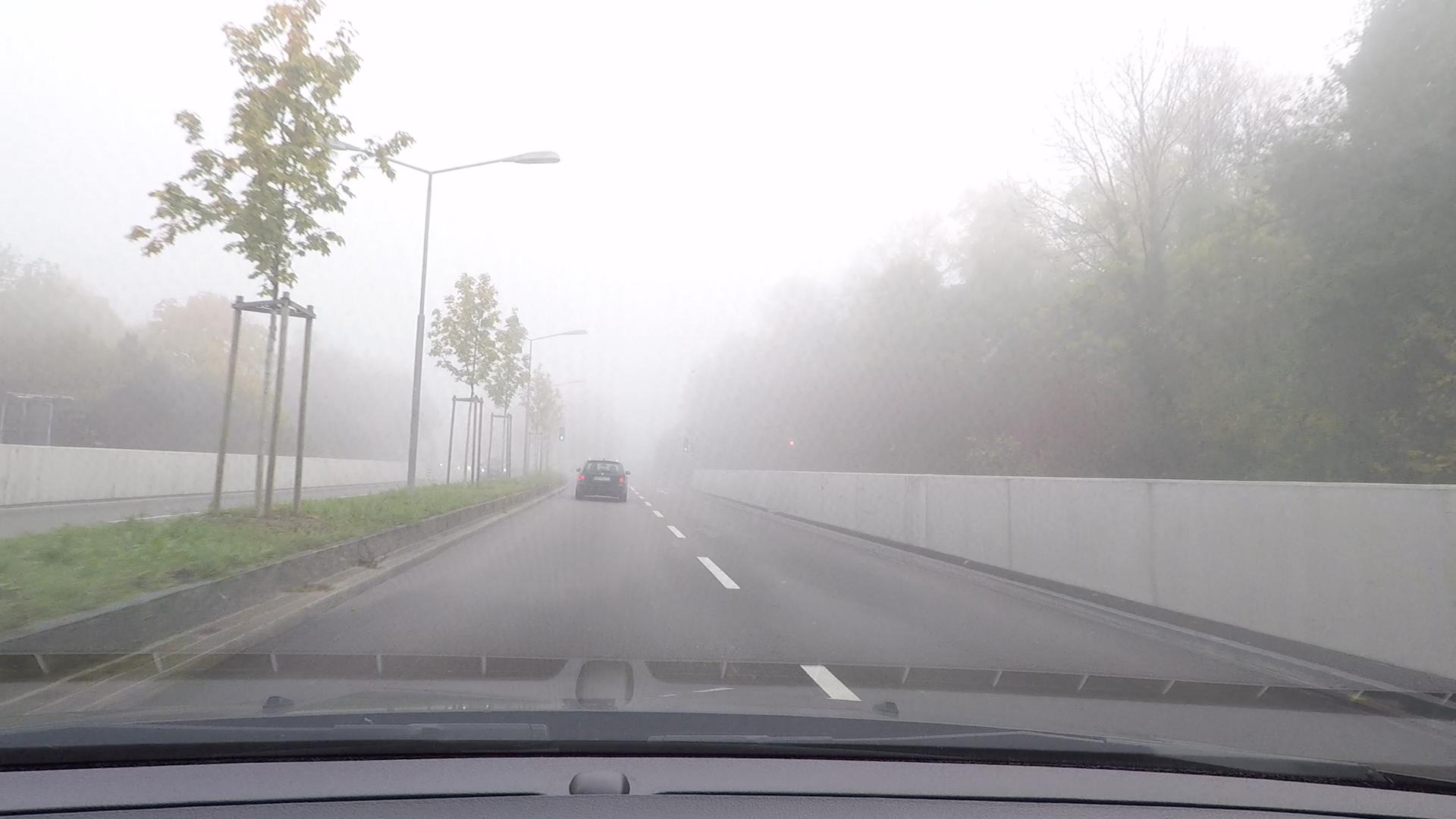}}
  \hfil
  \subfloat{\includegraphics[width=0.24\textwidth]{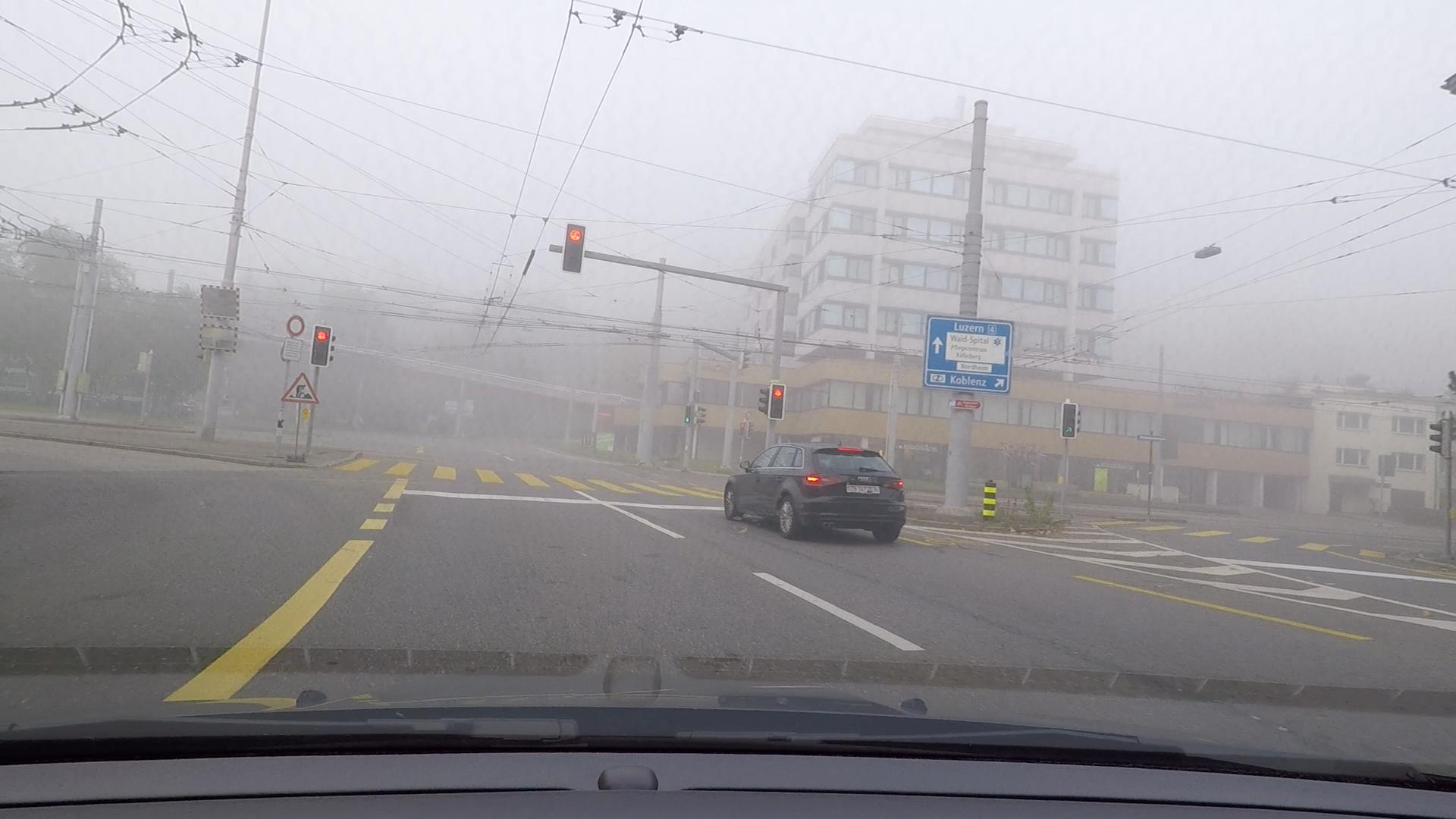}}
  \hfil
  \subfloat{\includegraphics[width=0.24\textwidth]{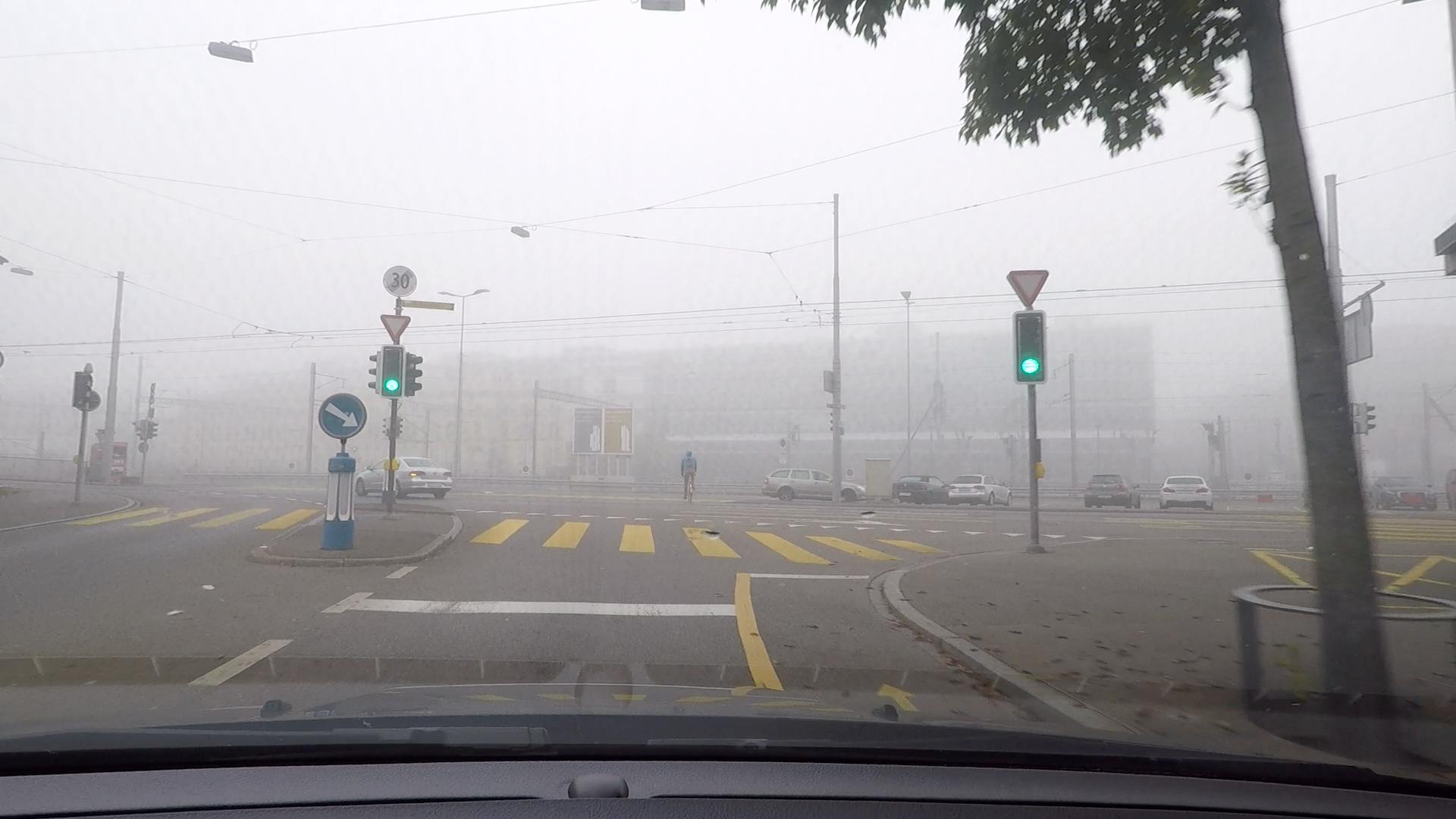}}
  \hfil
  \subfloat{\includegraphics[width=0.24\textwidth]{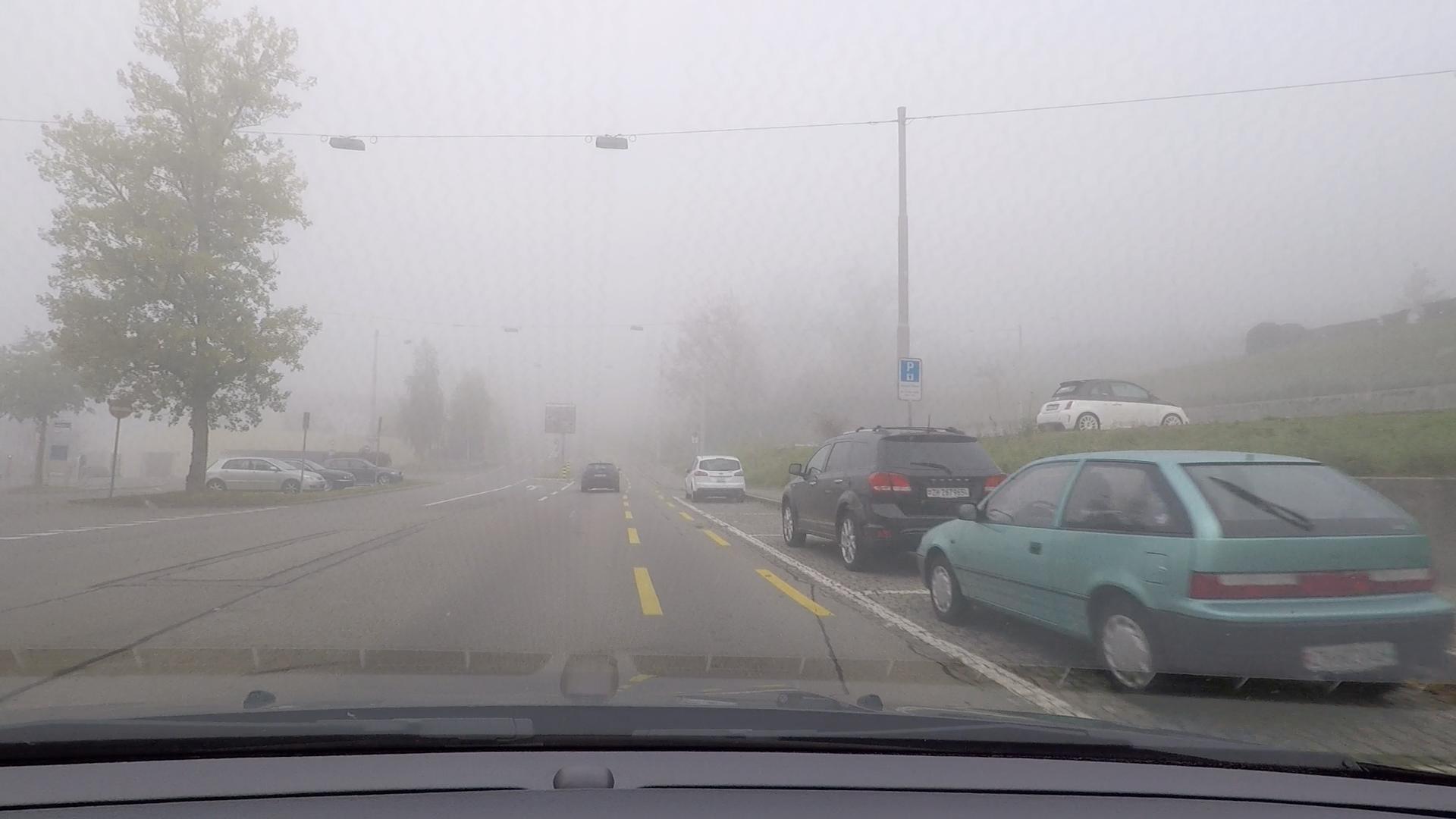}}\\  \vspace{-2.5mm}
  \subfloat{\rotatebox{90}{RefineNet~\cite{refinenet}}}
  \hfil
  \subfloat{\includegraphics[width=0.24\textwidth]{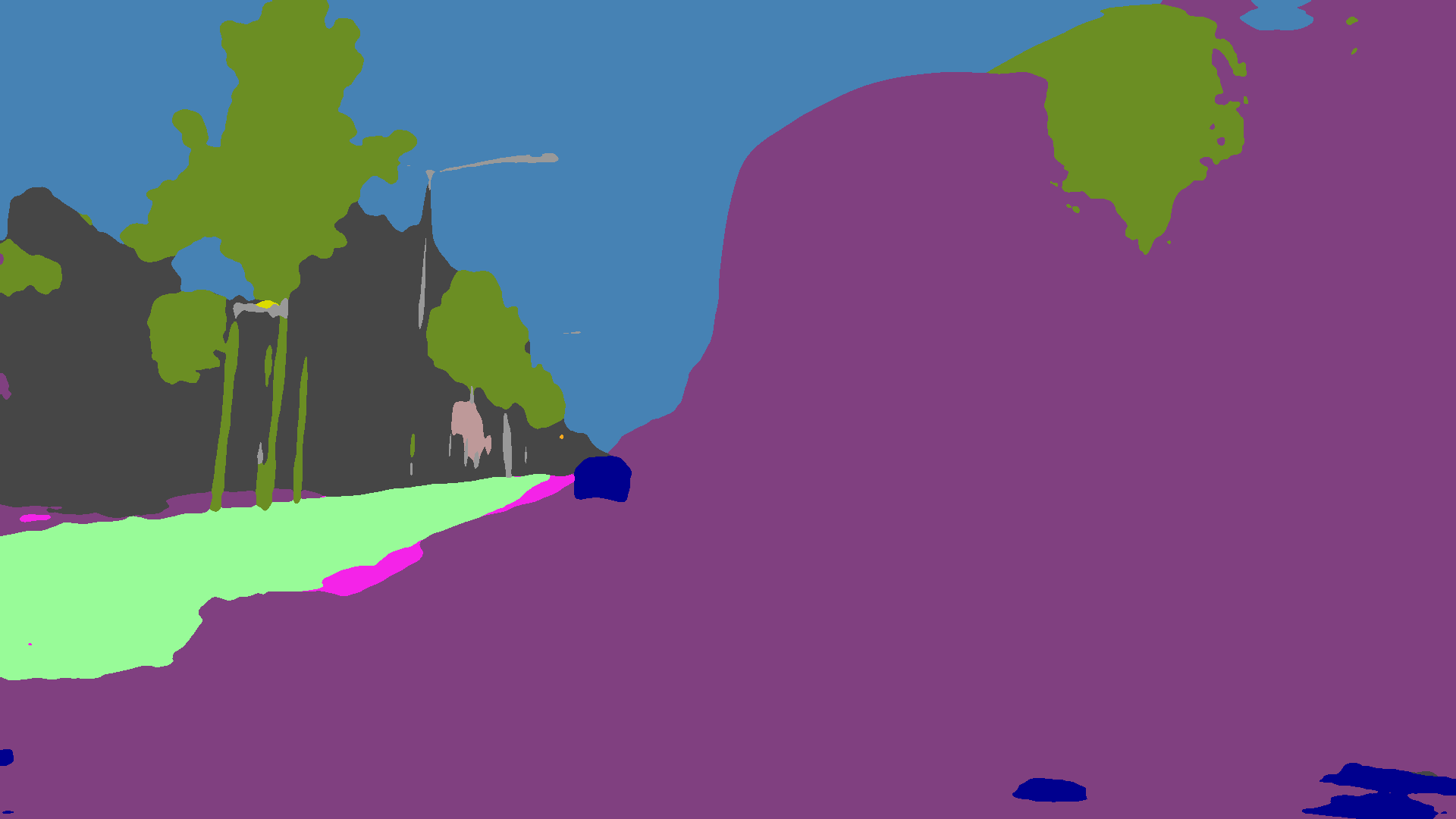}}
  \hfil
  \subfloat{\includegraphics[width=0.24\textwidth]{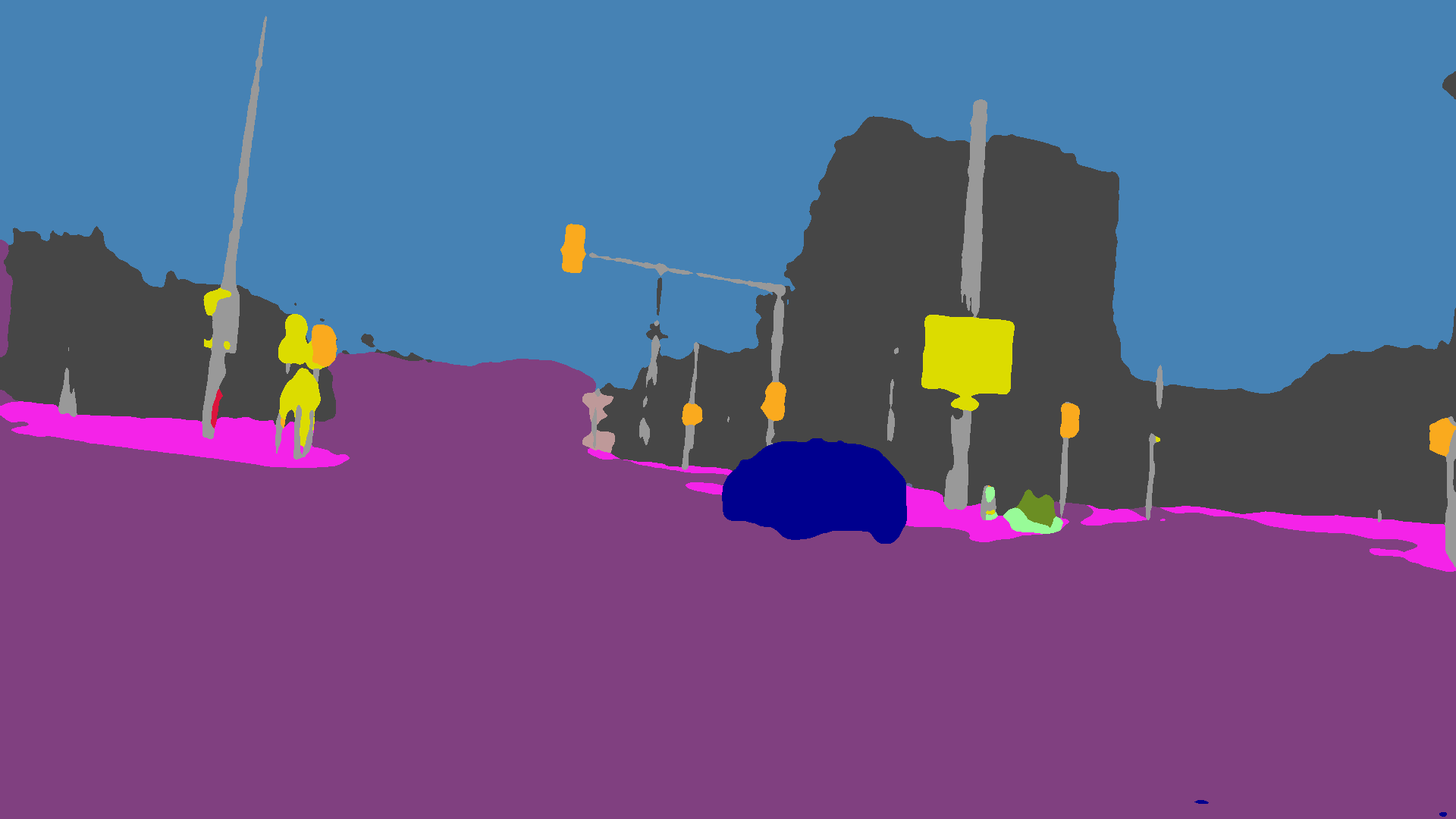}}
  \hfil
  \subfloat{\includegraphics[width=0.24\textwidth]{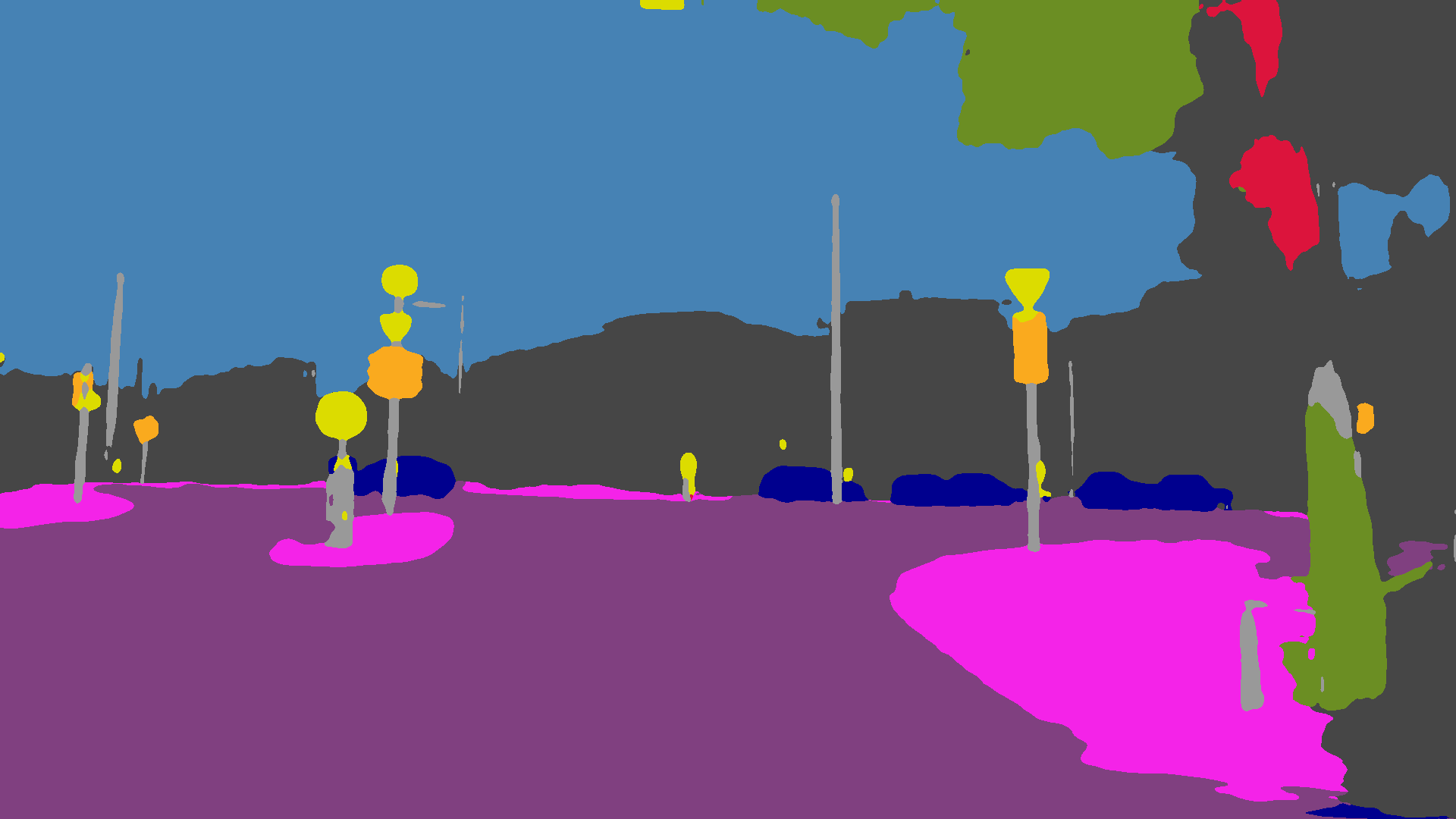}}
  \hfil
  \subfloat{\includegraphics[width=0.24\textwidth]{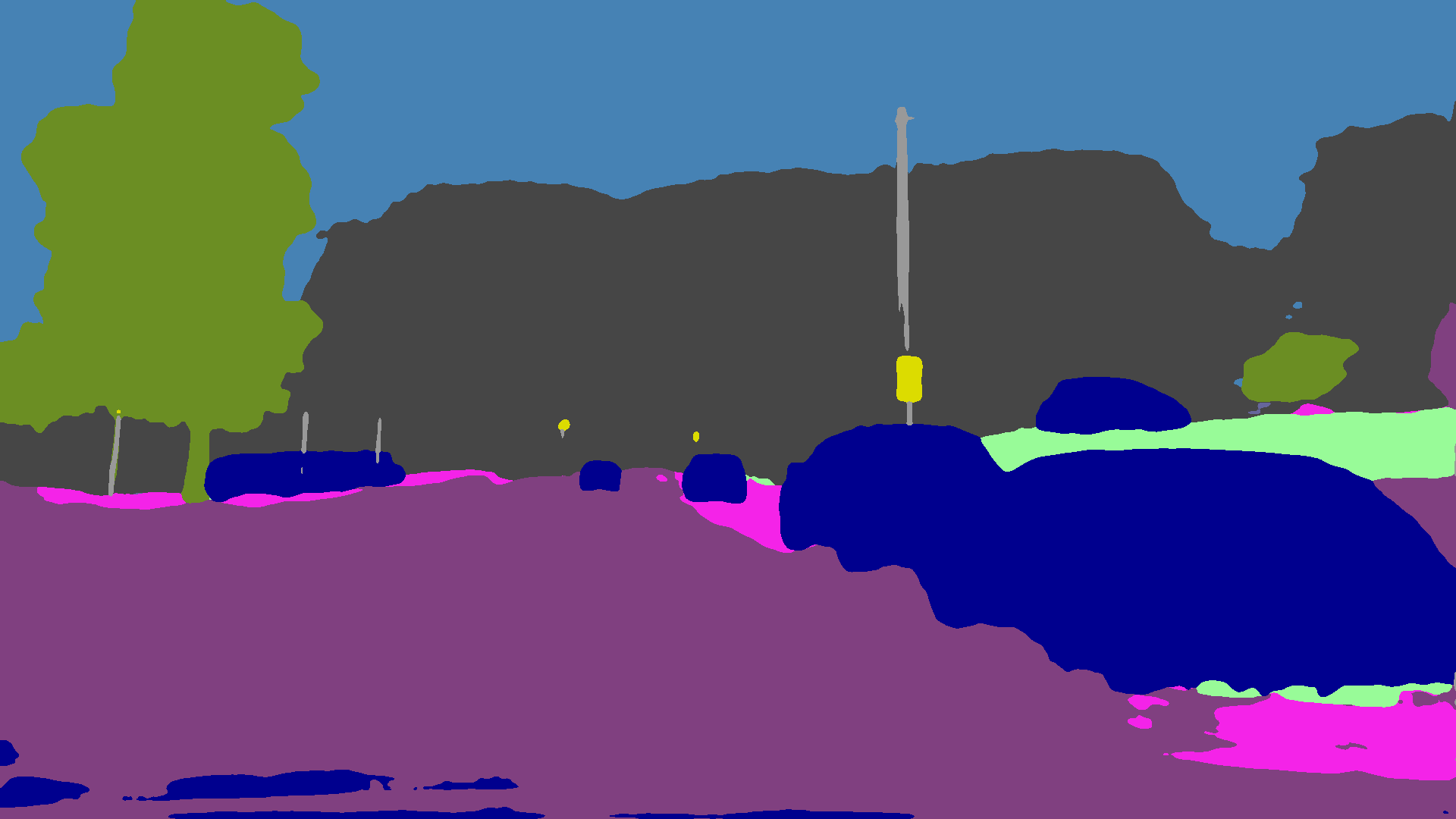}}\\ \vspace{-2.5mm}
  \subfloat{\rotatebox{90}{SFSU-synthetic~\cite{SFSU_synthetic}}} 
  \hfil
  \subfloat{\includegraphics[width=0.24\textwidth]{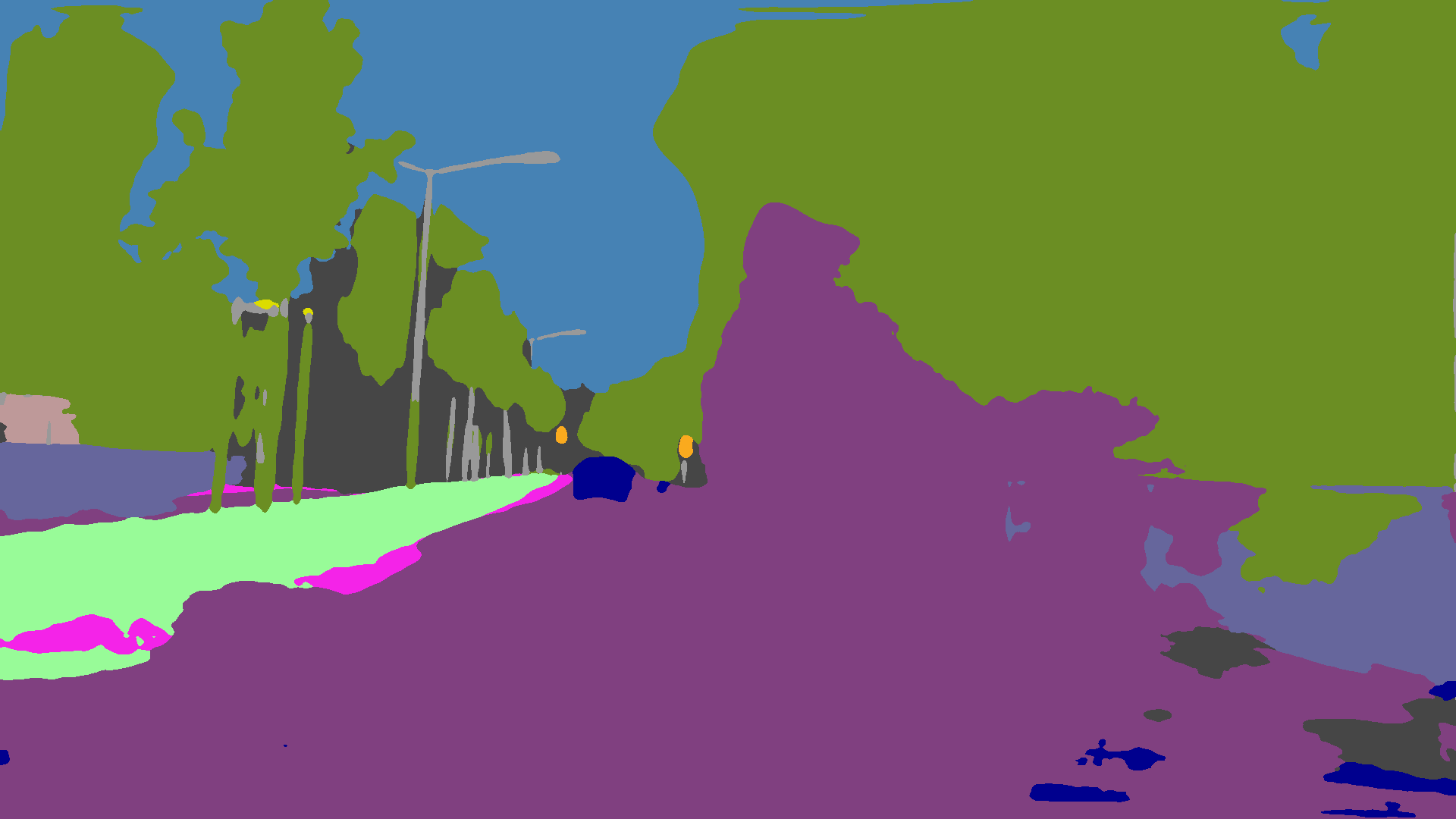}}
  \hfil
  \subfloat{\includegraphics[width=0.24\textwidth]{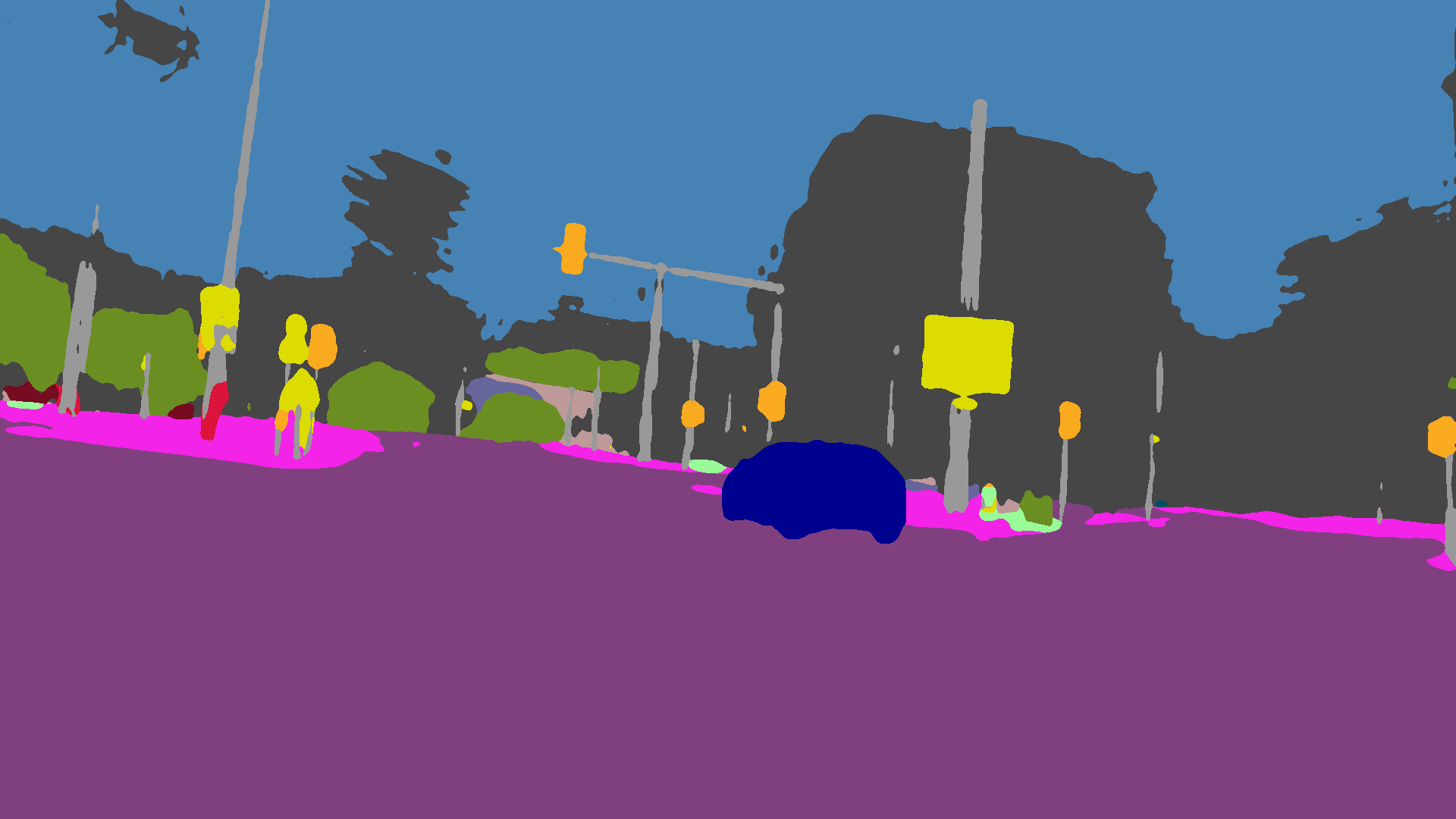}}
  \hfil
  \subfloat{\includegraphics[width=0.24\textwidth]{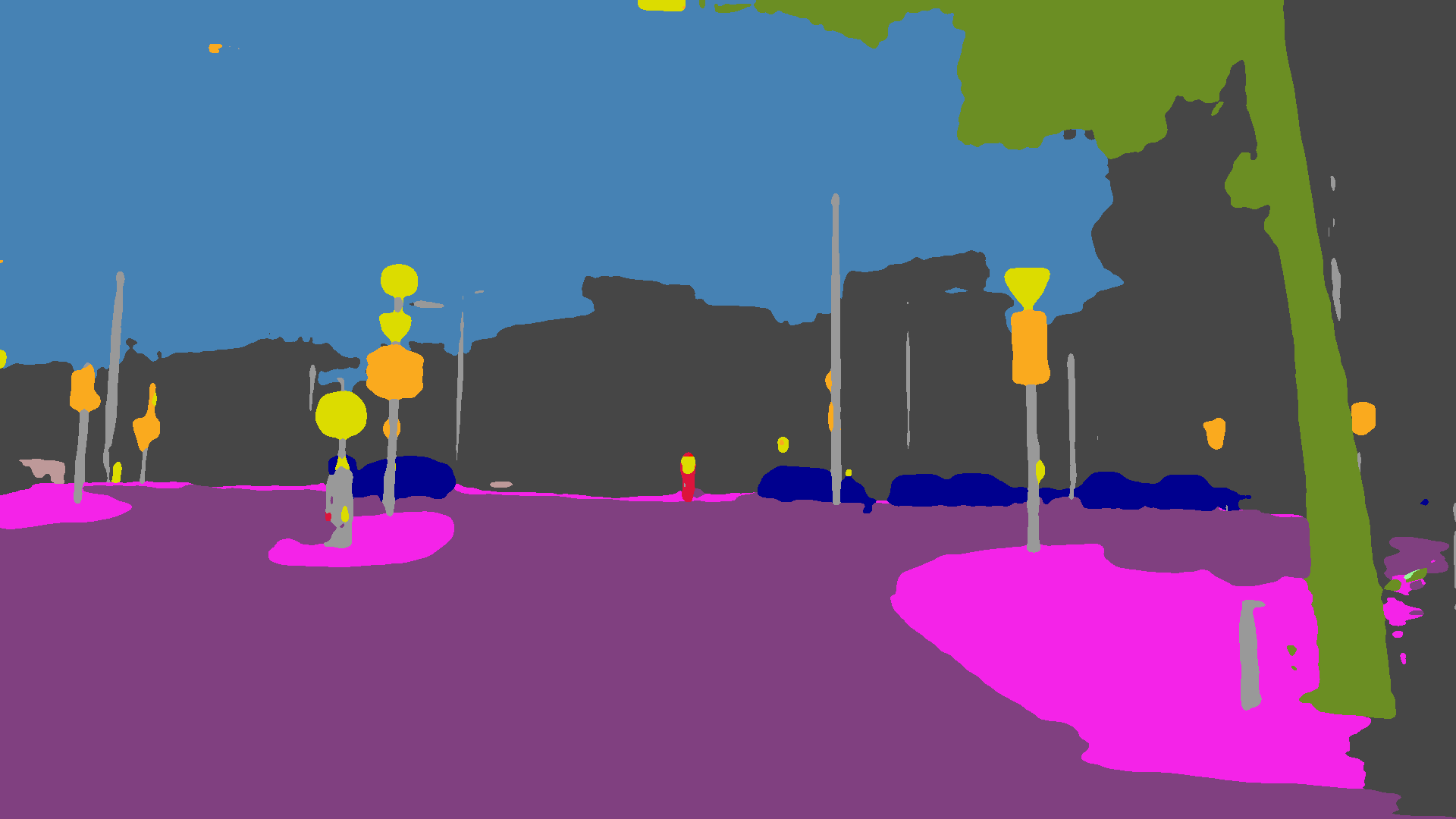}}
  \hfil 
  \subfloat{\includegraphics[width=0.24\textwidth]{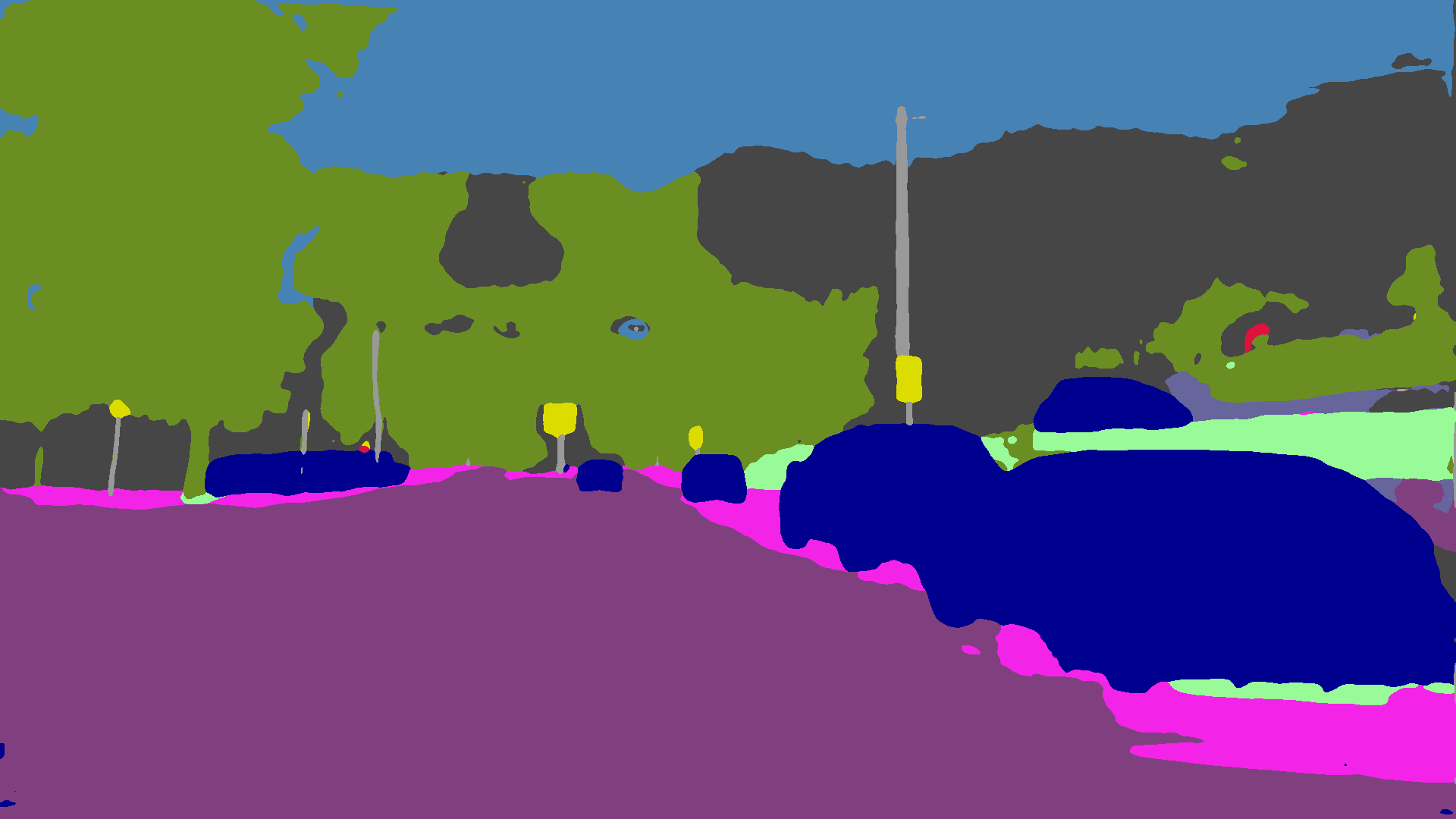}}\\  \vspace{-2.5mm}
    \subfloat{\rotatebox{90}{CMAda1 (ours)}} 
  \hfil
  \subfloat{\includegraphics[width=0.24\textwidth]{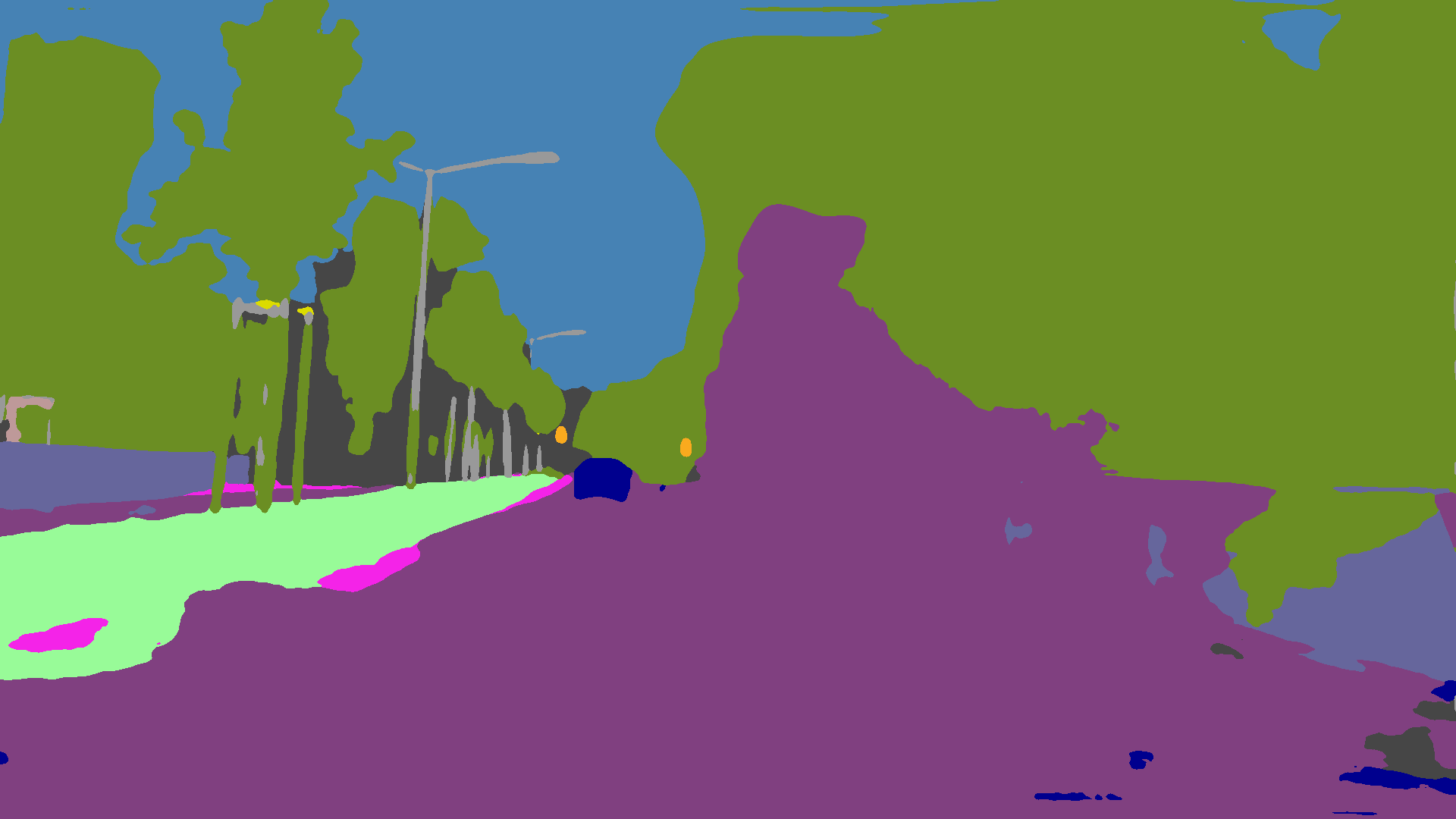}}
  \hfil
  \subfloat{\includegraphics[width=0.24\textwidth]{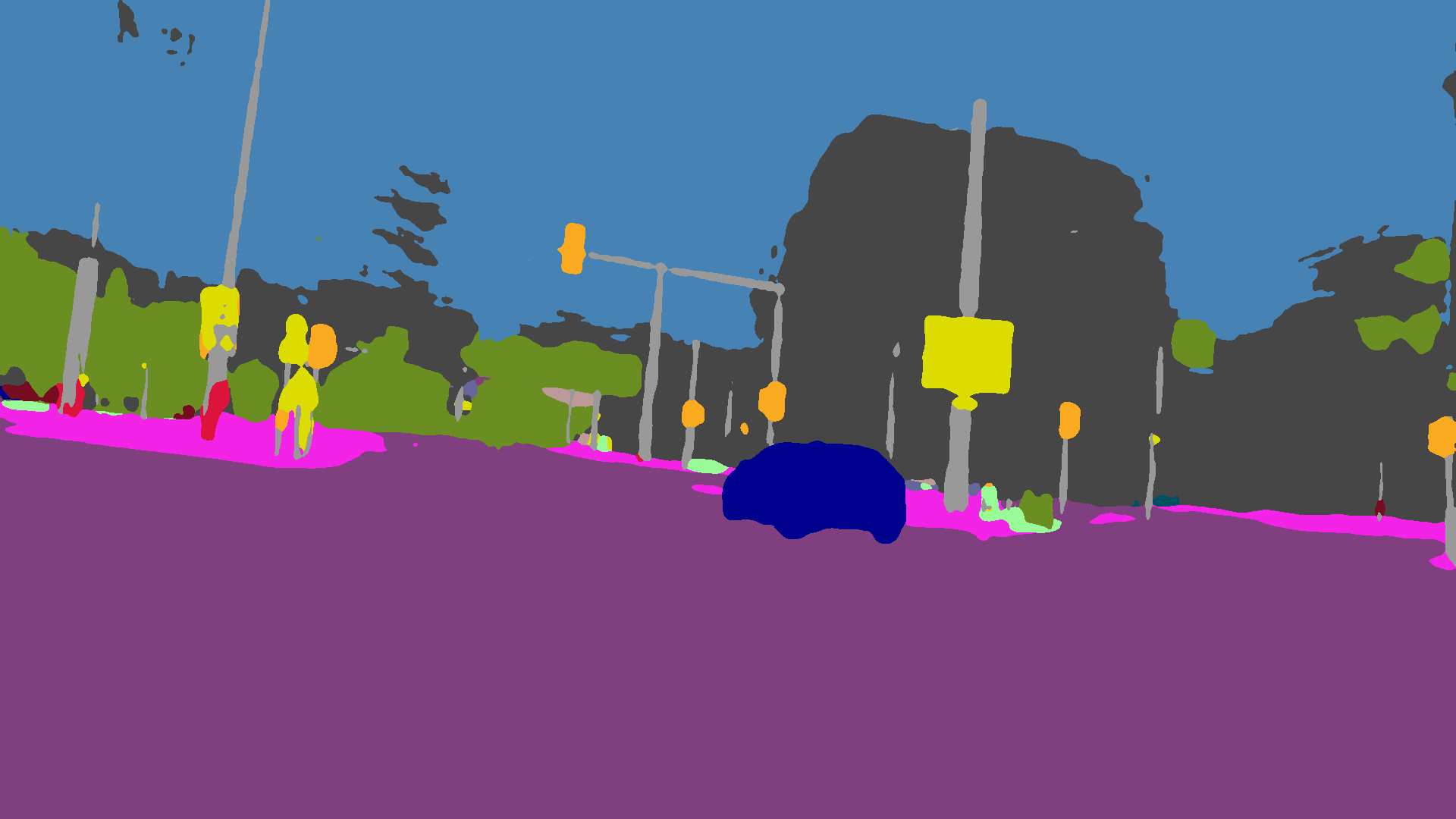}}
  \hfil
  \subfloat{\includegraphics[width=0.24\textwidth]{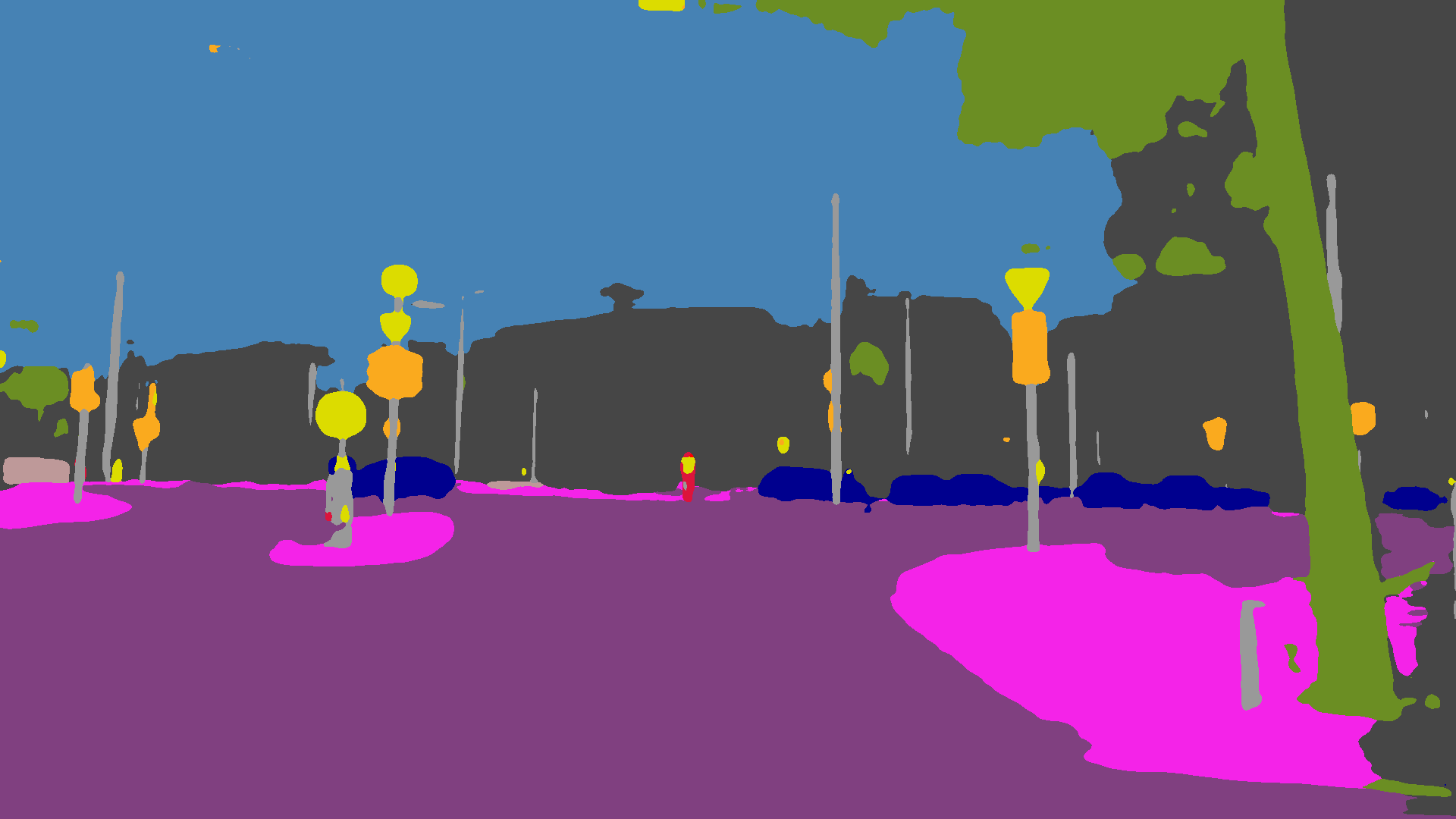}}
  \hfil 
  \subfloat{\includegraphics[width=0.24\textwidth]{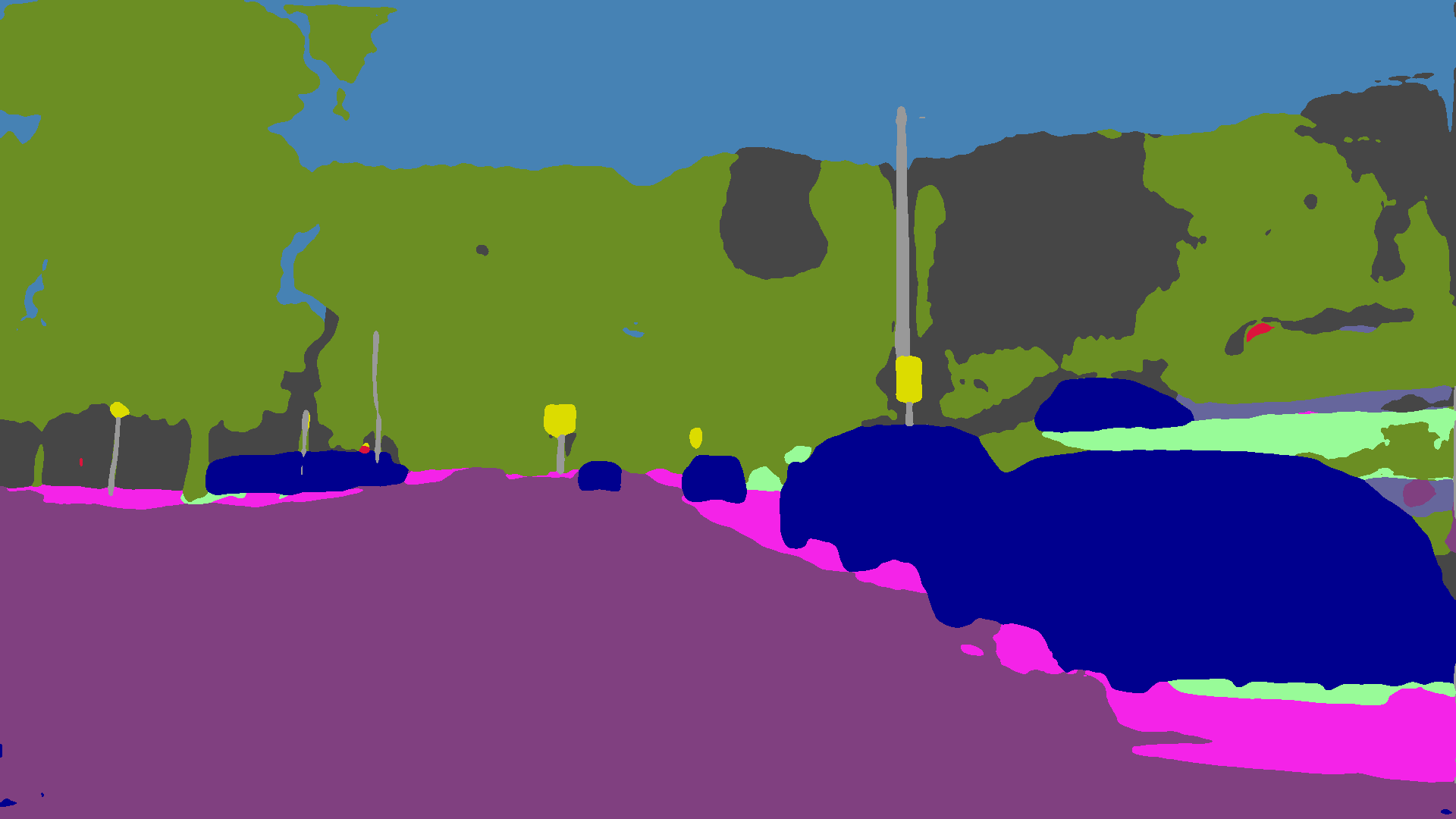}}\\  \vspace{-2.5mm}
   \subfloat{\rotatebox{90}{CMAda2 (ours)}}
  \hfil
  \subfloat{\includegraphics[width=0.24\textwidth]{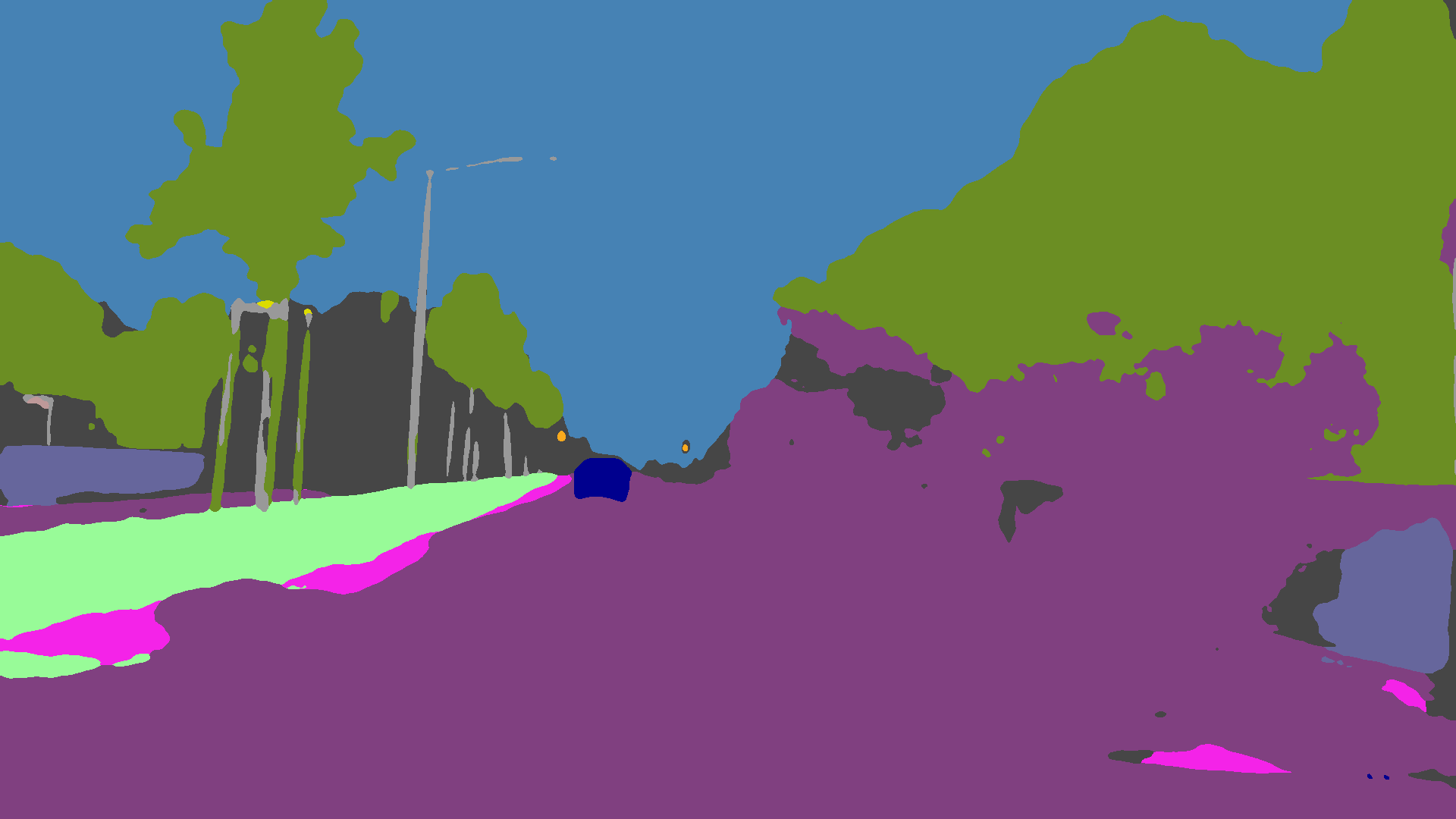}}
  \hfil
  \subfloat{\includegraphics[width=0.24\textwidth]{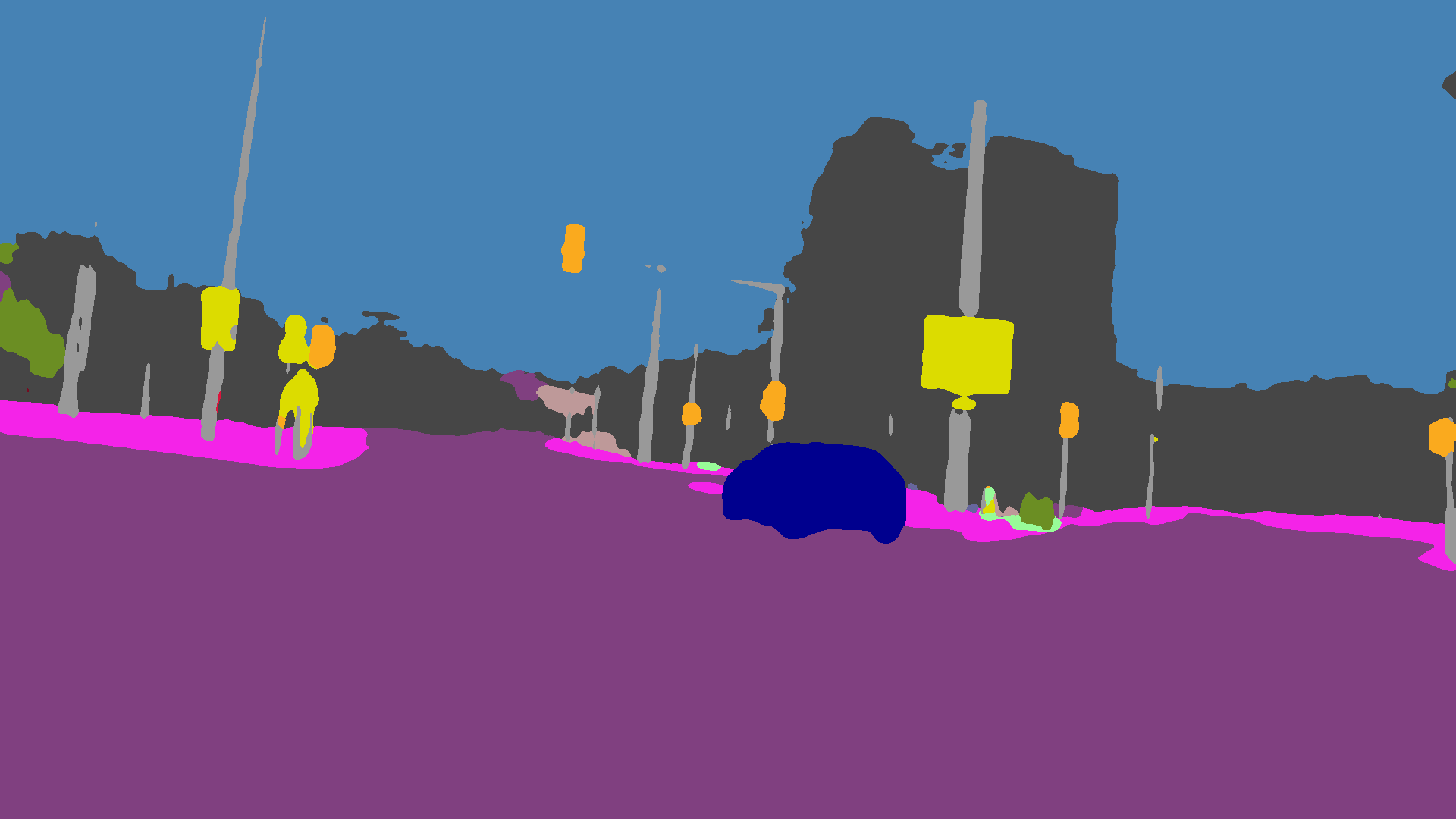}}
  \hfil
  \subfloat{\includegraphics[width=0.24\textwidth]{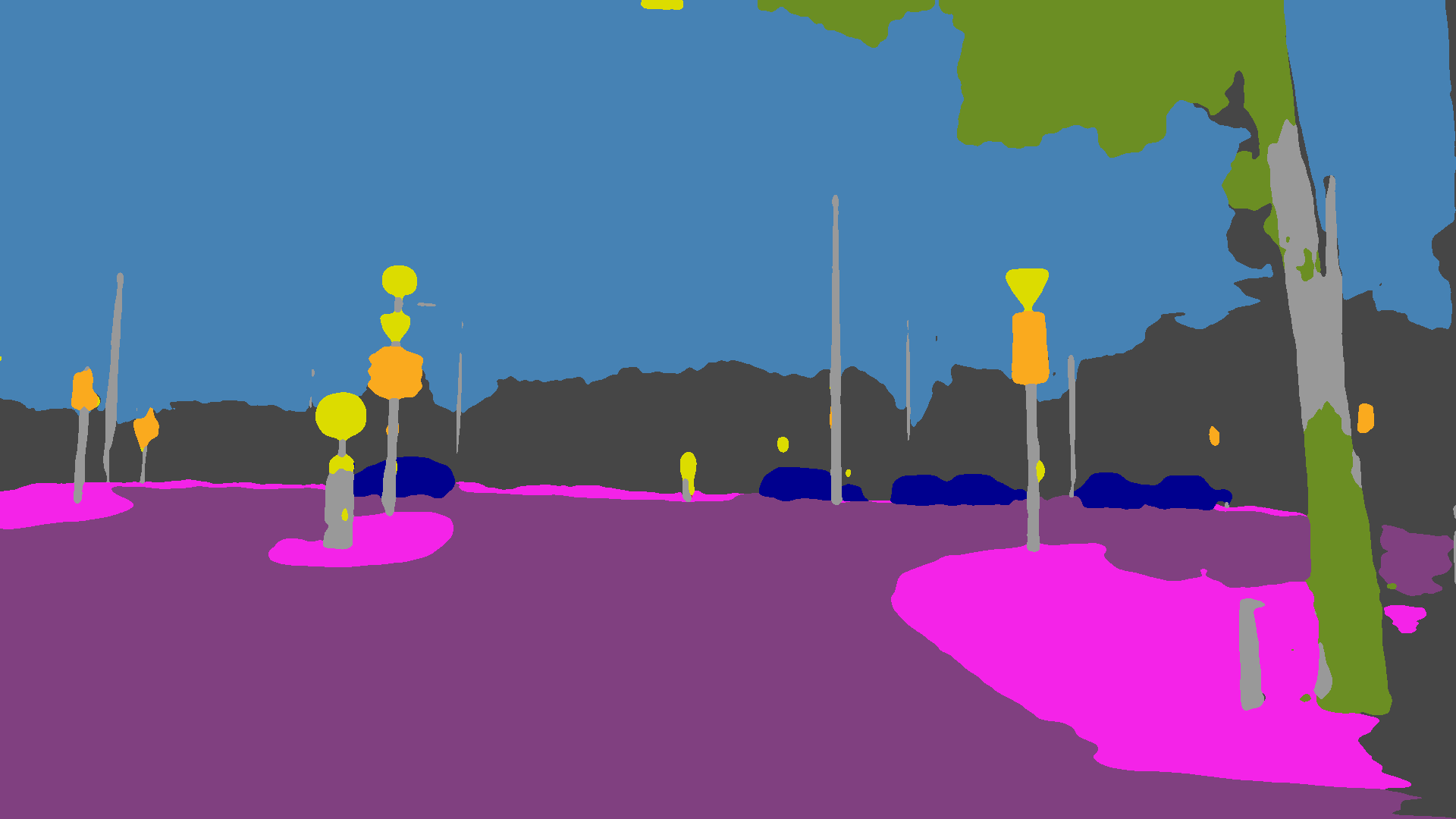}}
  \hfil 
  \subfloat{\includegraphics[width=0.24\textwidth]{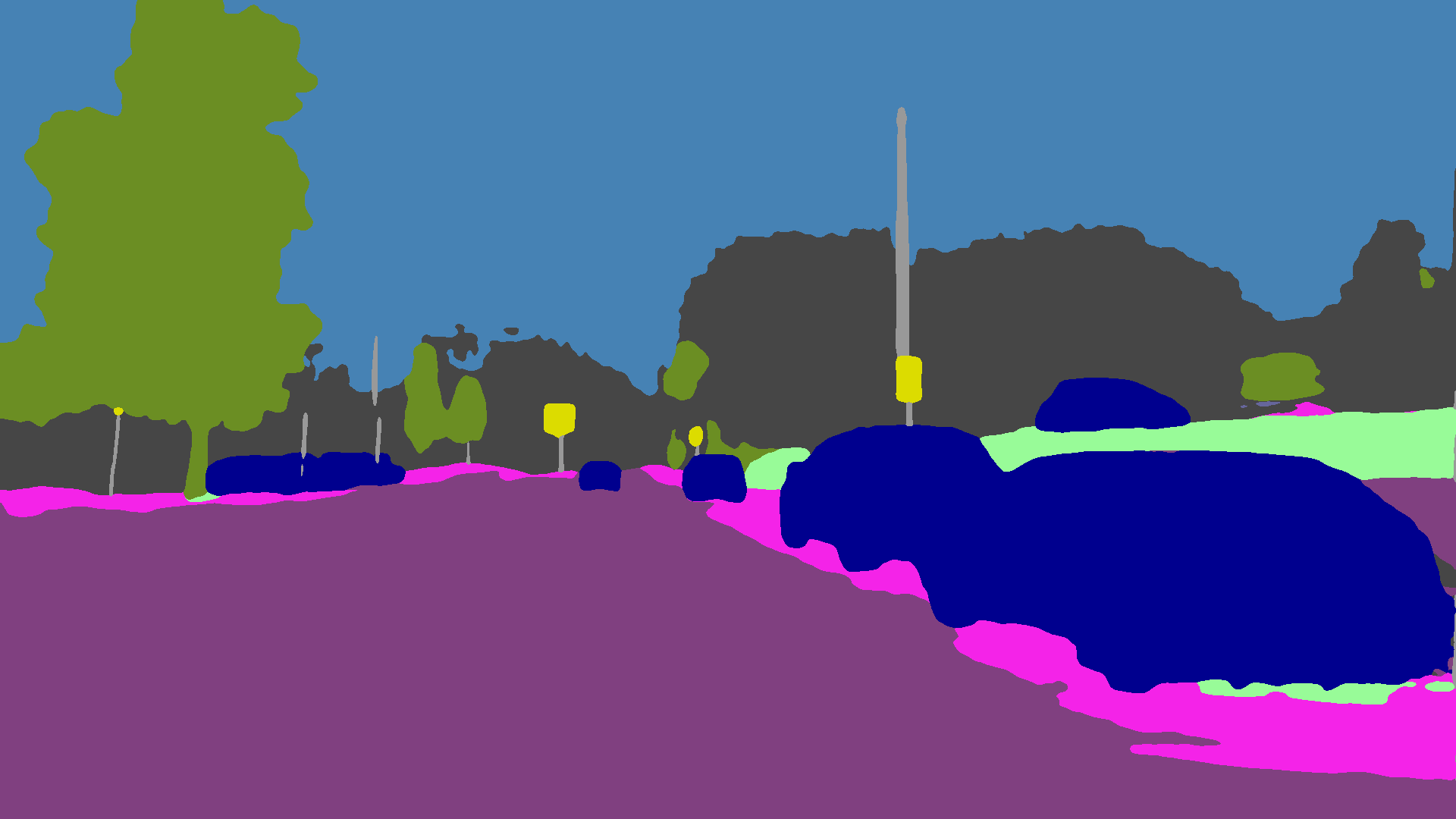}}\\  \vspace{-2.5mm}
   \subfloat{\rotatebox{90}{CMAda3+ (ours)}}
  \hfil
   \subfloat{\includegraphics[width=0.24\textwidth]{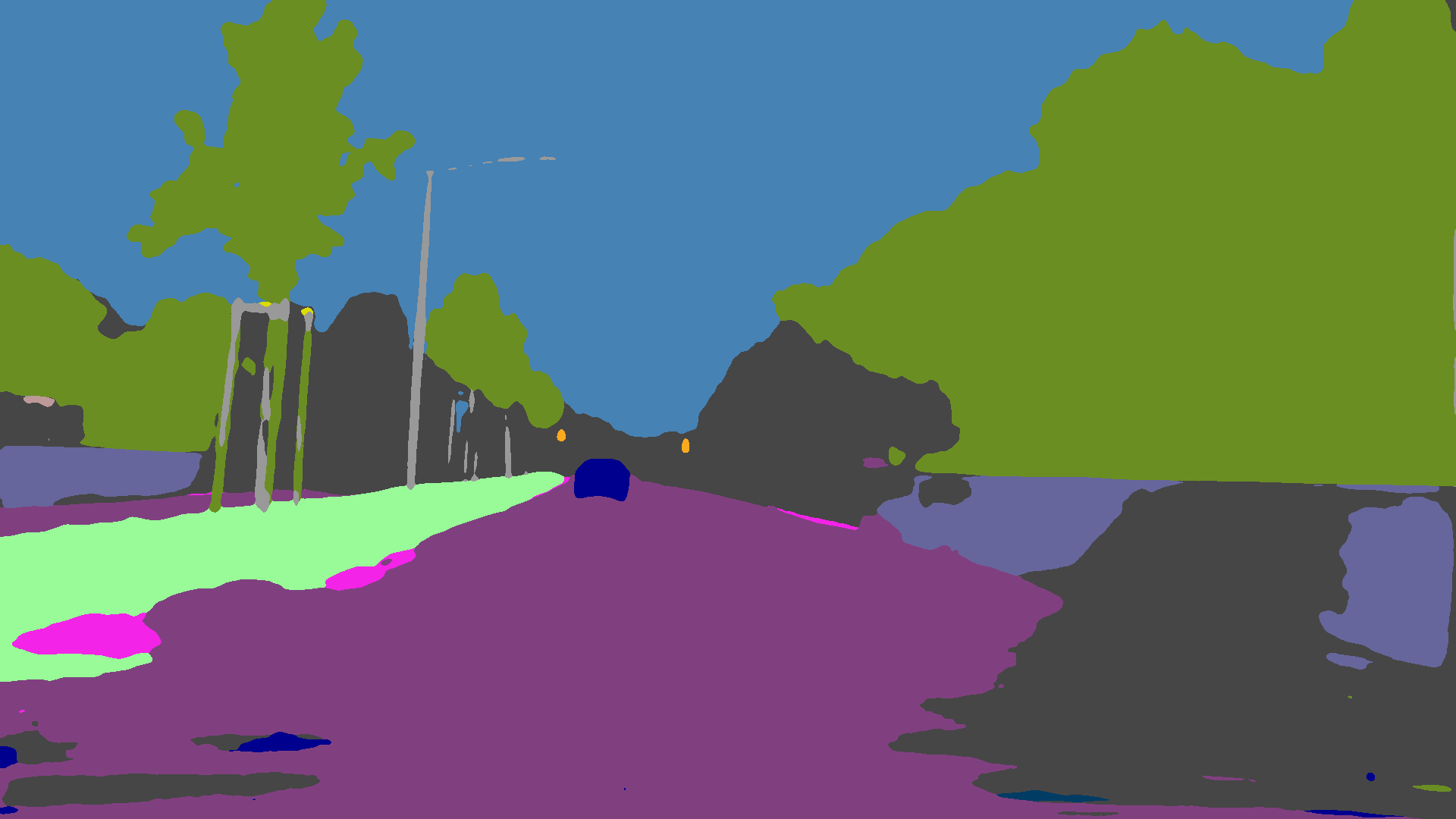}}
  \hfil
  \subfloat{\includegraphics[width=0.24\textwidth]{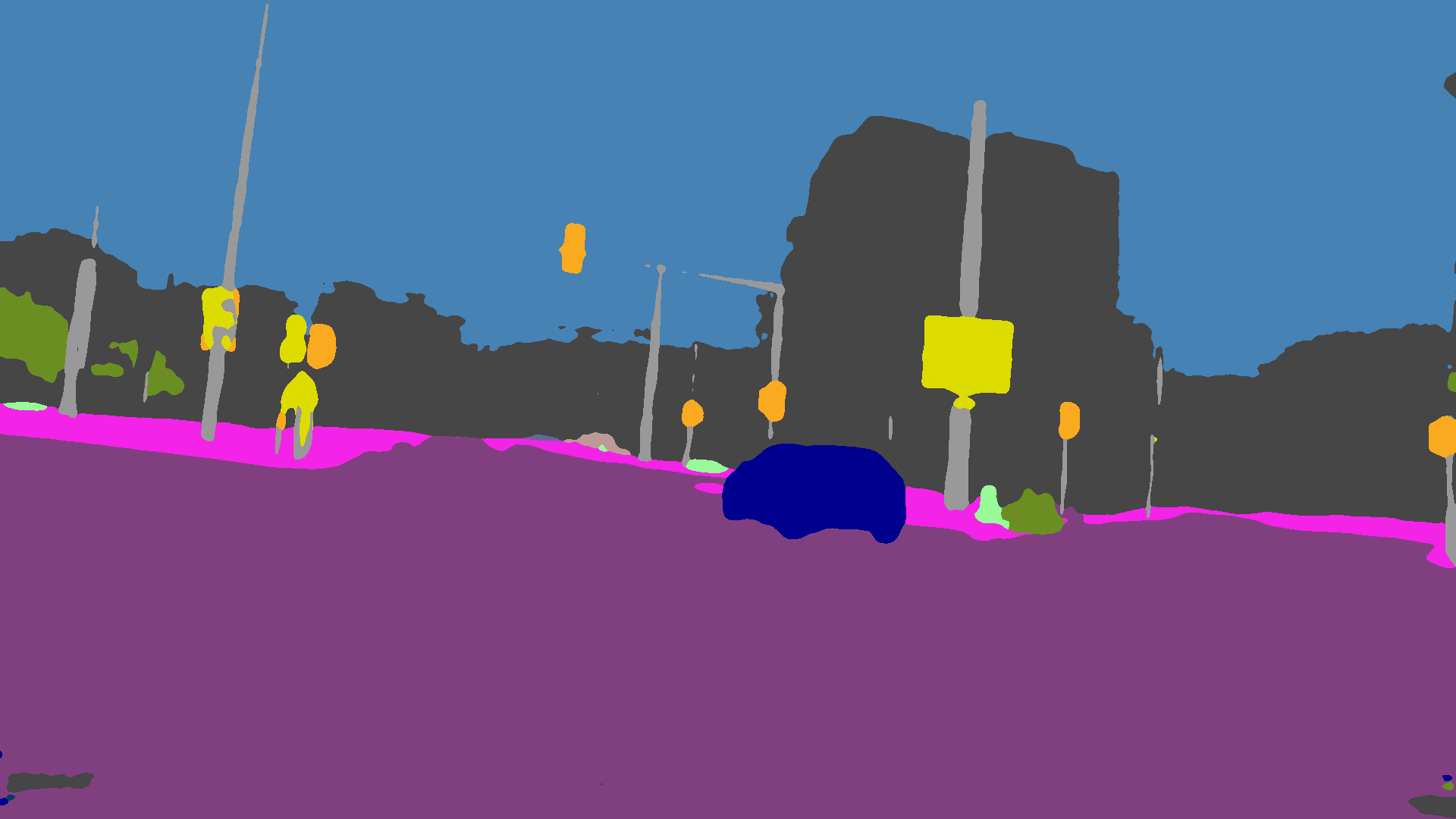}}
  \hfil
  \subfloat{\includegraphics[width=0.24\textwidth]{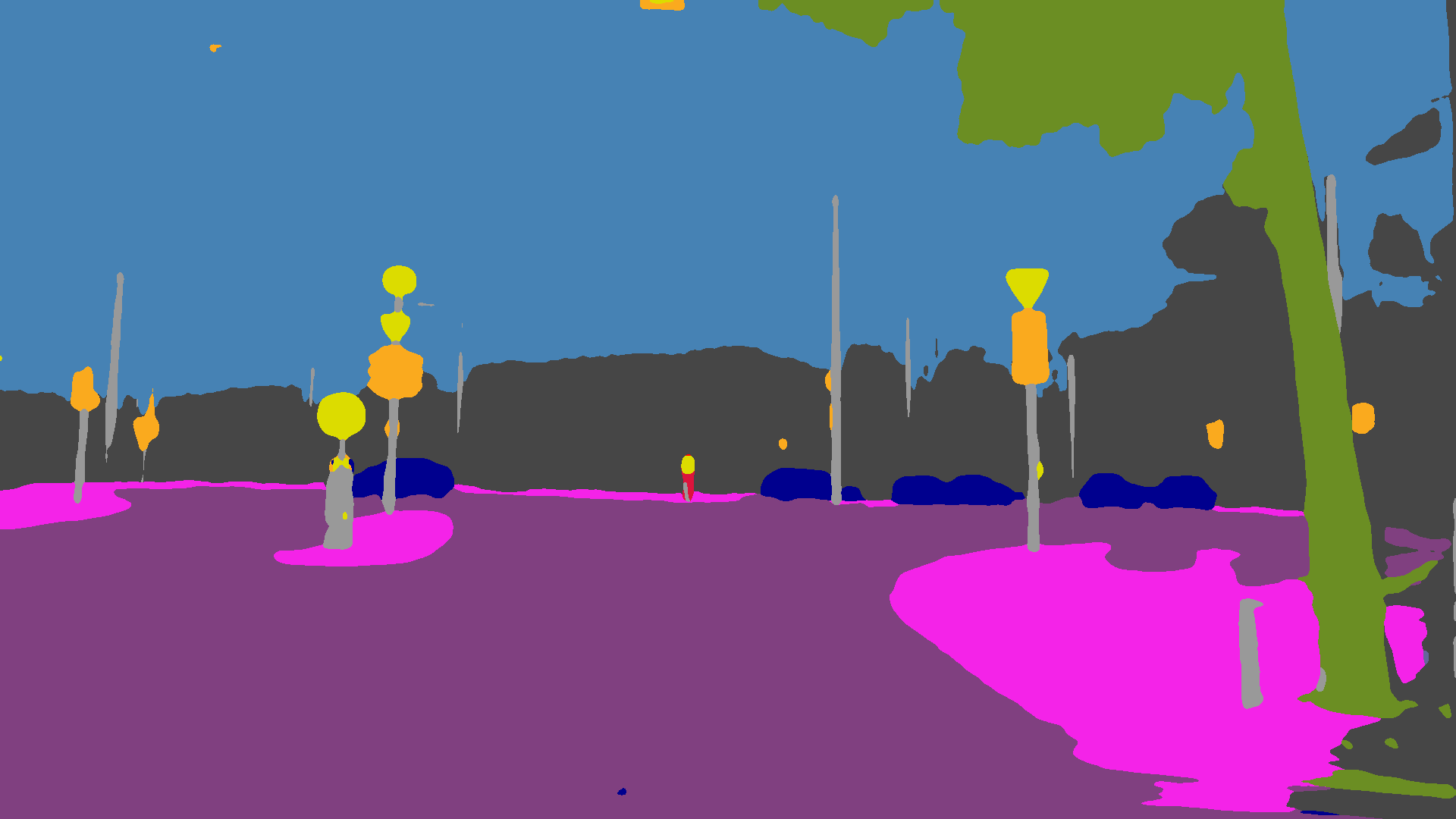}}
  \hfil 
  \subfloat{\includegraphics[width=0.24\textwidth]{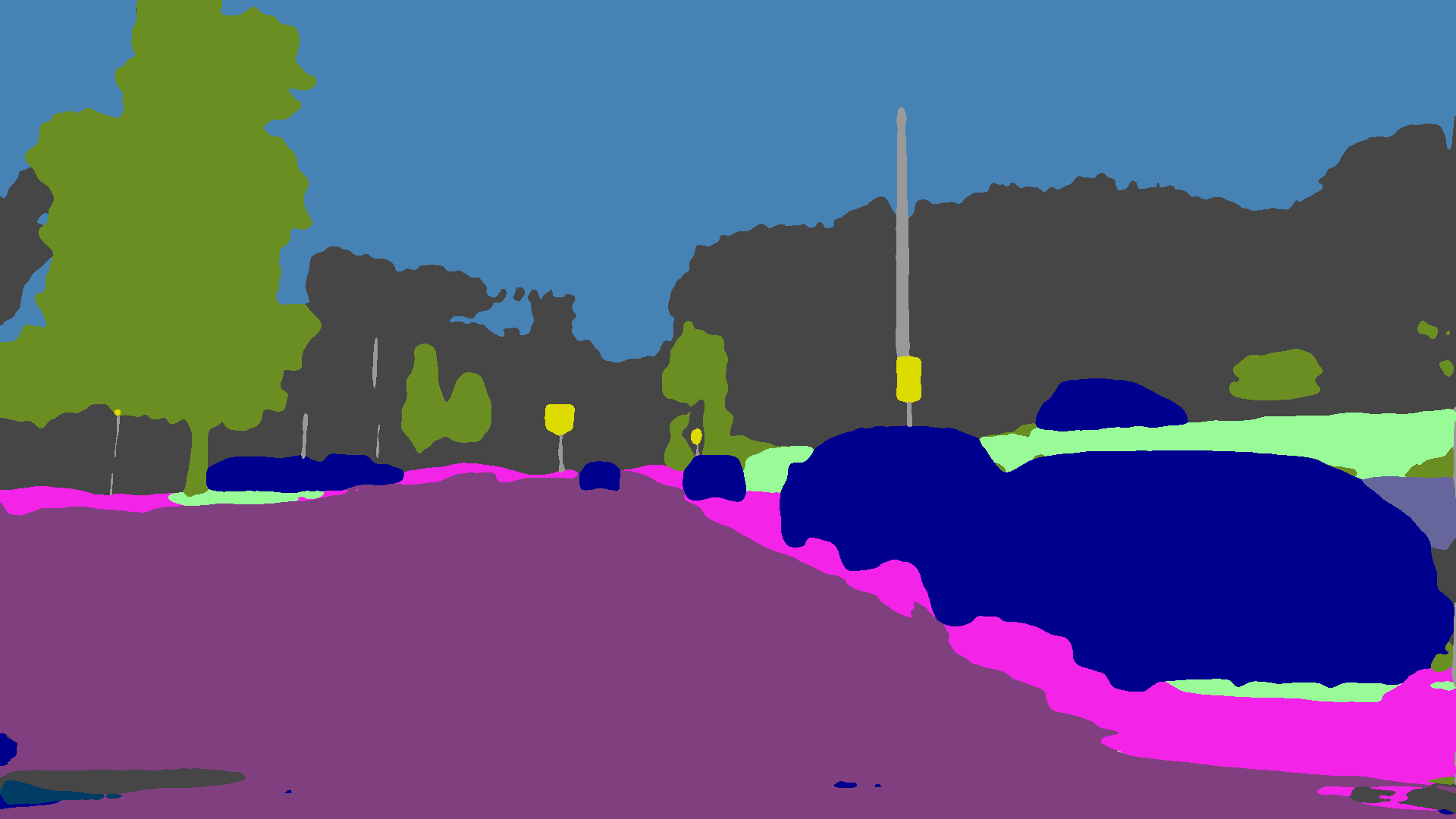}}\\  \vspace{-2.5mm}
     \subfloat{\rotatebox{90}{Ground Truth}}
  \hfil
   \subfloat{\includegraphics[width=0.24\textwidth]{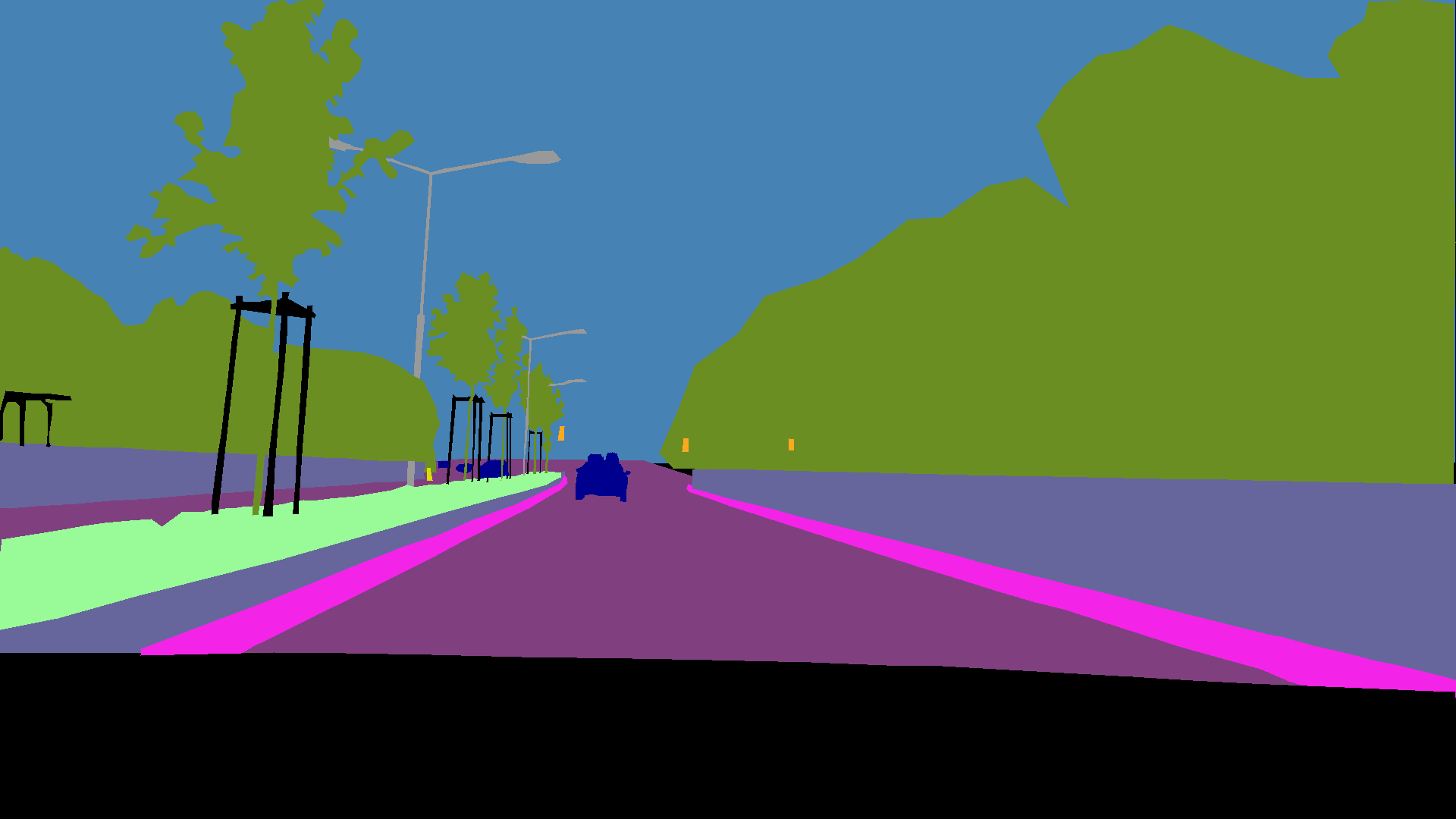}}
  \hfil
  \subfloat{\includegraphics[width=0.24\textwidth]{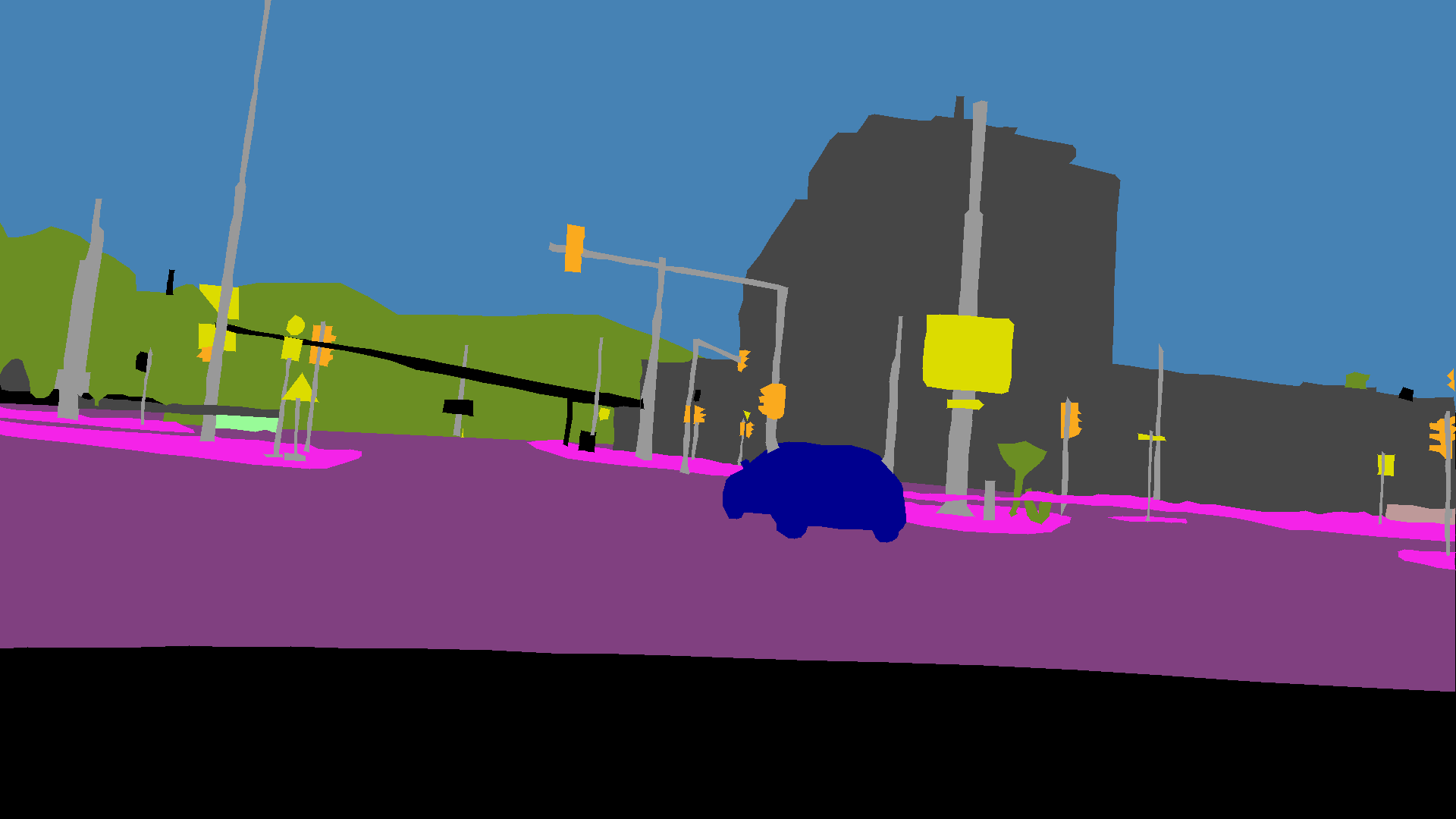}}
  \hfil
  \subfloat{\includegraphics[width=0.24\textwidth]{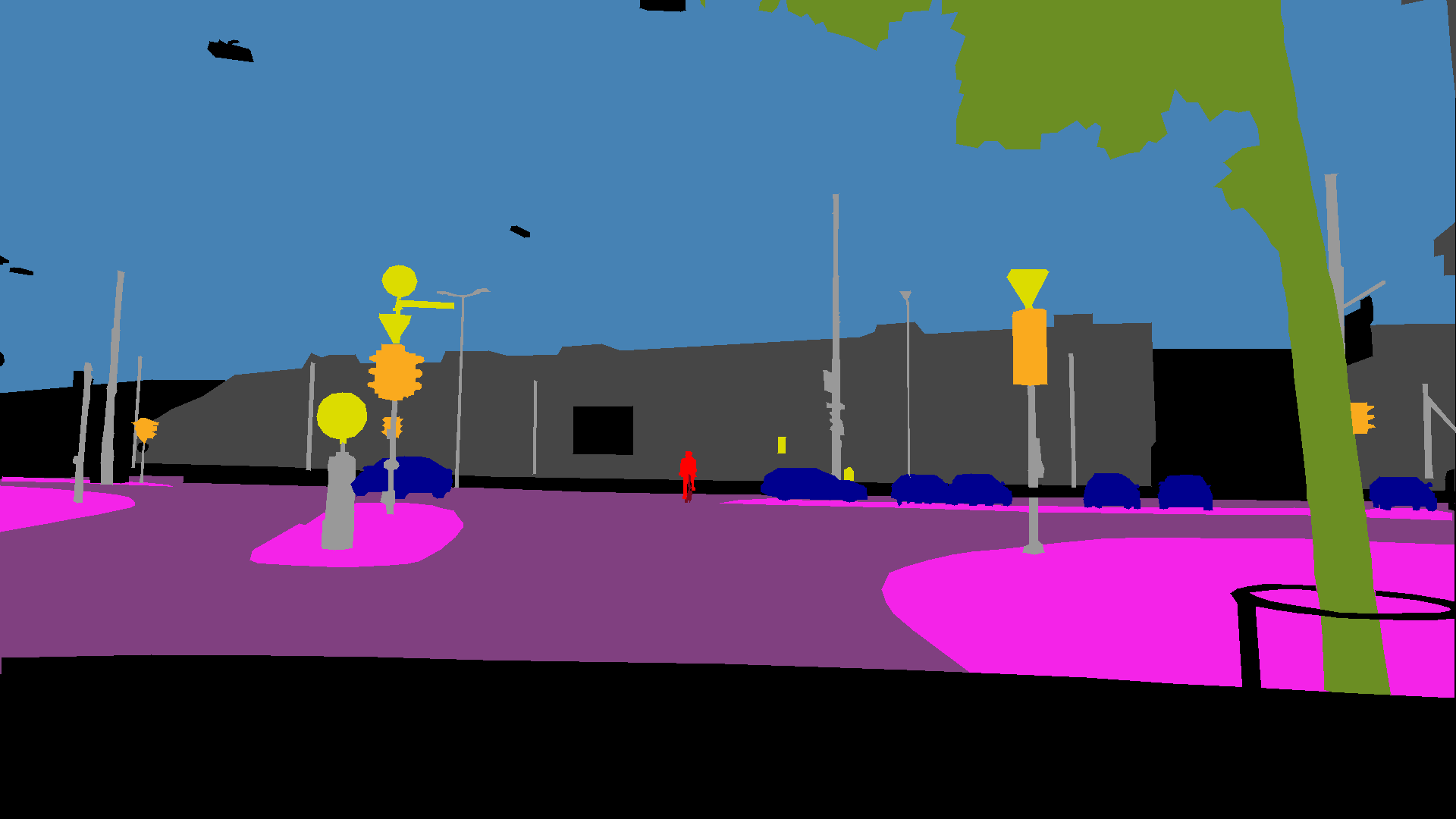}}
  \hfil 
  \subfloat{\includegraphics[width=0.24\textwidth]{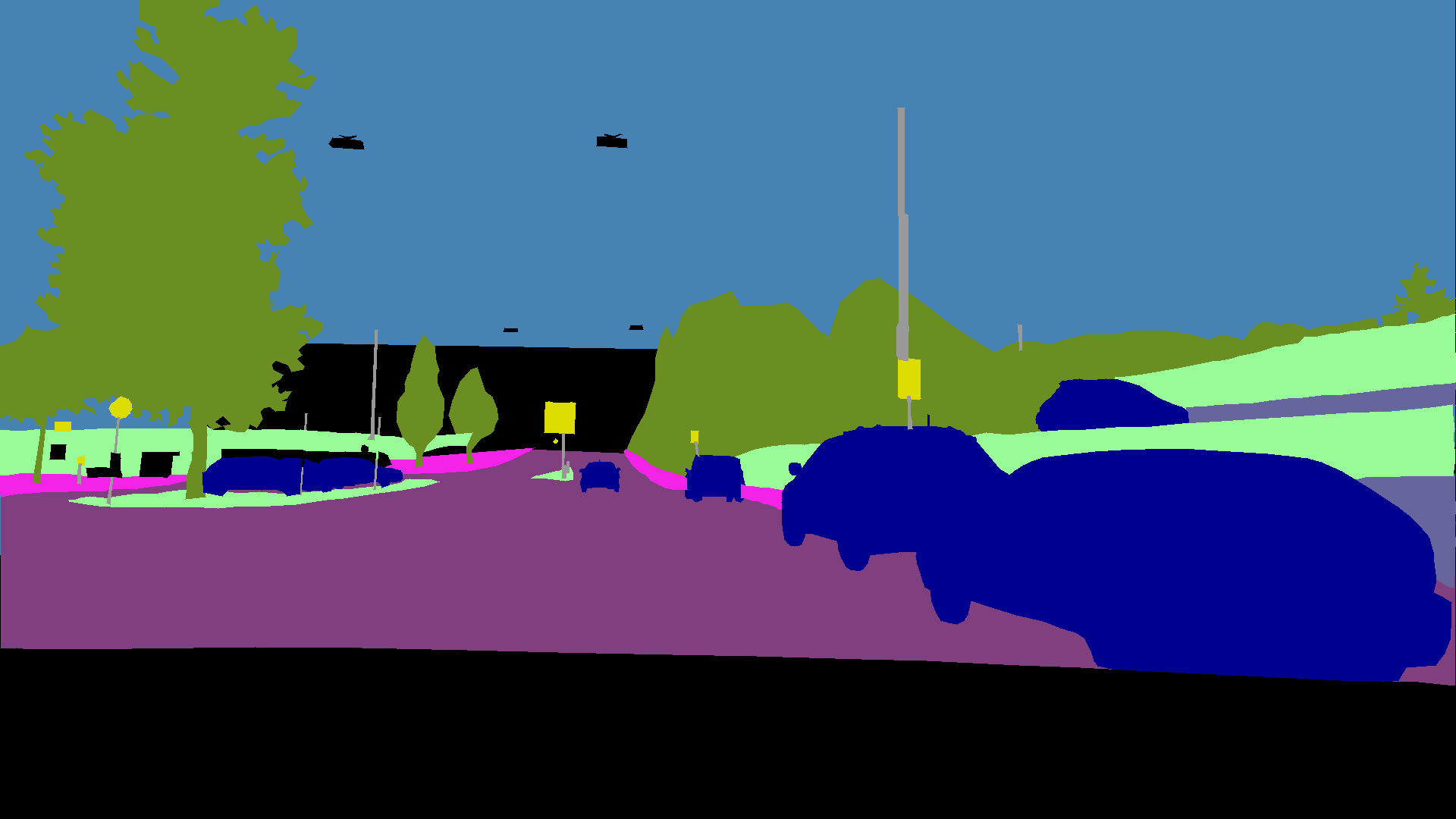}}\\ 
  \resizebox{\linewidth}{!}{
  \begin{tikzpicture}[tight background, scale=0.75, every node/.style={font=\large}]
      \draw[white, fill=void, draw=white] (0,0) rectangle (1 * 4, 1) node[pos=0.5] {Void};
      \draw[white, fill=road, draw=white] (1 * 4,0) rectangle (2 * 4, 1) node[pos=0.5] {Road};
      \draw[white, fill=sidewalk, draw=white] (2 * 4,0) rectangle (3 * 4, 1) node[pos=0.5] {Sidewalk};
      \draw[white, fill=building, draw=white] (3 * 4,0) rectangle (4 * 4, 1) node[pos=0.5] {Building};
      \draw[white, fill=wall, draw=white] (4 * 4,-0) rectangle (5 * 4, 1) node[pos=0.5] {Wall};
      \draw[black, fill=fence, draw=white] (5 * 4,-0) rectangle (6 * 4, 1) node[pos=0.5] {Fence};
      \draw[white, fill=pole, draw=white] (6 * 4,-0) rectangle (7 * 4, 1) node[pos=0.5] {Pole};
      \draw[white, fill=traffic light, draw=white] (7 * 4,-0) rectangle (8 * 4, 1) node[pos=0.5] {Traffic Light};
      \draw[black, fill=traffic sign, draw=white] (8 * 4,-0) rectangle (9 * 4, 1) node[pos=0.5] {Traffic Sign};
      \draw[white, fill=vegetation, draw=white] (9 * 4,-0) rectangle (10 * 4, 1) node[pos=0.5] {Vegetation};

      \draw[black, fill=terrain, draw=white] (0 * 4,-1) rectangle (1 * 4, 0) node[pos=0.5] {Terrain};
      \draw[white, fill=sky, draw=white] (1 * 4,-1) rectangle (4 * 2, 0) node[pos=0.5] {Sky};
      \draw[white, fill=person, draw=white] (2 * 4,-1) rectangle (3 * 4, 0) node[pos=0.5] {Person};
      \draw[white, fill=rider, draw=white] (3 * 4,-1) rectangle (4 * 4, 0) node[pos=0.5] {Rider};
      \draw[white, fill=car, draw=white] (4 * 4,-1) rectangle (5 * 4, 0) node[pos=0.5] {Car};
      \draw[white, fill=truck, draw=white] (5 * 4,-1) rectangle (6 * 4, 0) node[pos=0.5] {Truck};
      \draw[white, fill=bus, draw=white] (6 * 4,-1) rectangle (7 * 4, 0) node[pos=0.5] {Bus};
      \draw[white, fill=train, draw=white] (7 * 4,-1) rectangle (8 * 4, 0) node[pos=0.5] {Train};
      \draw[white, fill=motorcycle, draw=white] (8 * 4,-1) rectangle (9 * 4, 0) node[pos=0.5] {Motorcycle};
      \draw[white, fill=bicycle, draw=white] (9 * 4,-1) rectangle (10 * 4, 0) node[pos=0.5] {Bicycle};
  \end{tikzpicture}}
\caption{Qualitative results for semantic segmentation on \emph{Foggy Zurich-test}}
\label{fig:sem:seg}
\end{figure*}

\subsubsection{Benefit of Fog Densification}
\label{sec:benefit:fog:densification}

The final component of our proposed pipeline that we evaluate is our fog densification method, introduced in Section~\ref{sec:CMAda+}. Table~\ref{table:experiments:main} shows the results of CMAda2+ and CMAda3+ on the three test datasets, along with the results of their counterparts CMAda2 and CMAda3. CMAda2+ and CMAda2 use the same training parameters. The same holds for CMAda3+ and CMAda3.
Applying our fog densification to the real foggy training sets used in CMAda significantly improves performance for both numbers of adaptation stages that are examined. For instance, CMAda3+ outperforms CMAda3 by $3.1\%$, $2.4\%$ and $0.9\%$ on \emph{Foggy Zurich-test}, \emph{Foggy Driving-dense} and \emph{Foggy Driving} respectively.
This is because without fog densification, the images in the synthetic dataset $\mathcal{D}_{\text{syn}}^z$ of each adaptation stage (defined in \eqref{eq:dataset:syn}) have the exact same fog density $\beta_z$ as images in the target domain of that stage, whereas the images in the real dataset $\mathcal{D}_{\text{real}}^z$ have \emph{lower} fog density than $\beta_z$ (cf.\ \eqref{eq:dataset:real}). This lower fog density of the real training images facilitates the self-learning, bootstrapping strategy. However, it also creates a domain gap between training and test images due to the difference in their fog density. On the contrary, the dataset with densified fog defined in \eqref{eq:dataset:real+} matches the target fog density of the test images, which helps close this domain gap and significantly boosts the performance of CMAda.

\subsubsection{Qualitative Results and Discussion}
\label{sec:qualitative:results}

In Figure~\ref{fig:sem:seg}, we show segmentation results on \emph{Foggy Zurich-test} generated with our best-performing method CMAda3+, our conference paper method CMAda2 and the single-stage version CMAda1 using only synthetic training data from \emph{Foggy Cityscapes-DBF}, compared to the method of~\cite{SFSU_synthetic} that only uses synthetic data from Foggy Cityscapes~\cite{SFSU_synthetic} and the clear-weather RefineNet model~\cite{refinenet}. This visual comparison demonstrates that our multiple-stage methods CMAda3+ and CMAda2 yield significantly better results and generally capture the road layout more accurately than the two competing approaches and our single-stage method CMAda1. Moreover, the more stages CMAda involves, the more accurate the segmentation result is in general. For instance, on the leftmost image of Figure~\ref{fig:sem:seg}, CMAda3+ segments the wall and the vegetation on the right side much better than the other methods and only misclassifies some parts of them as \emph{building}, which is a much less detrimental error from a driving perspective than confusing these classes with \emph{road}, as is the case for the other methods. Similarly, the buildings and the tree trunk in the third image are better segmented by CMAda3+.

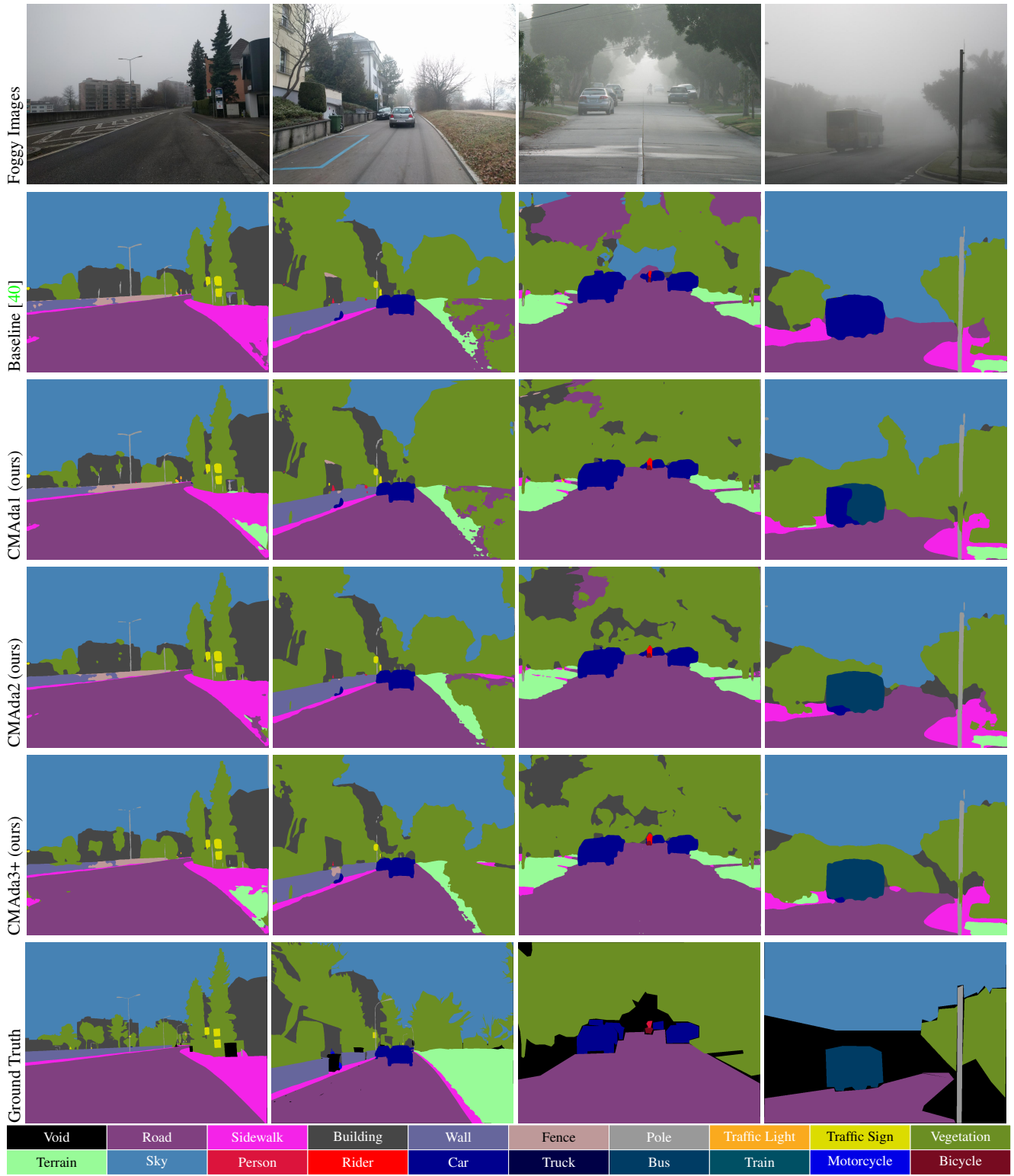
\begin{figure*}[!tb]
  \centering
  \addtocounter{subfigure}{-20}
    \subfloat{\rotatebox{90}{Foggy Images}}  
  \hfil
  \subfloat{\includegraphics[width=0.24\textwidth]{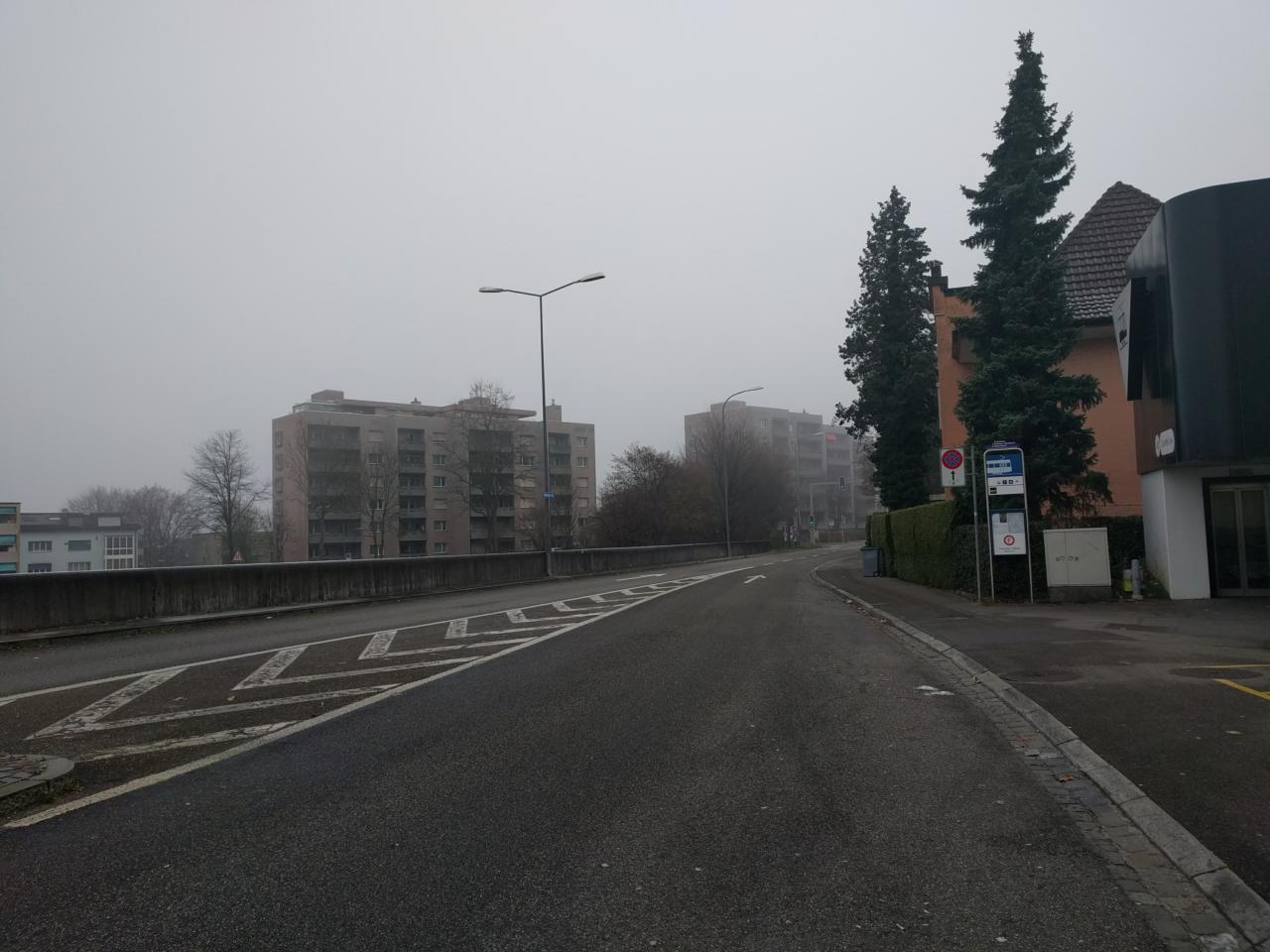}}
  \hfil
  \subfloat{\includegraphics[width=0.24\textwidth]{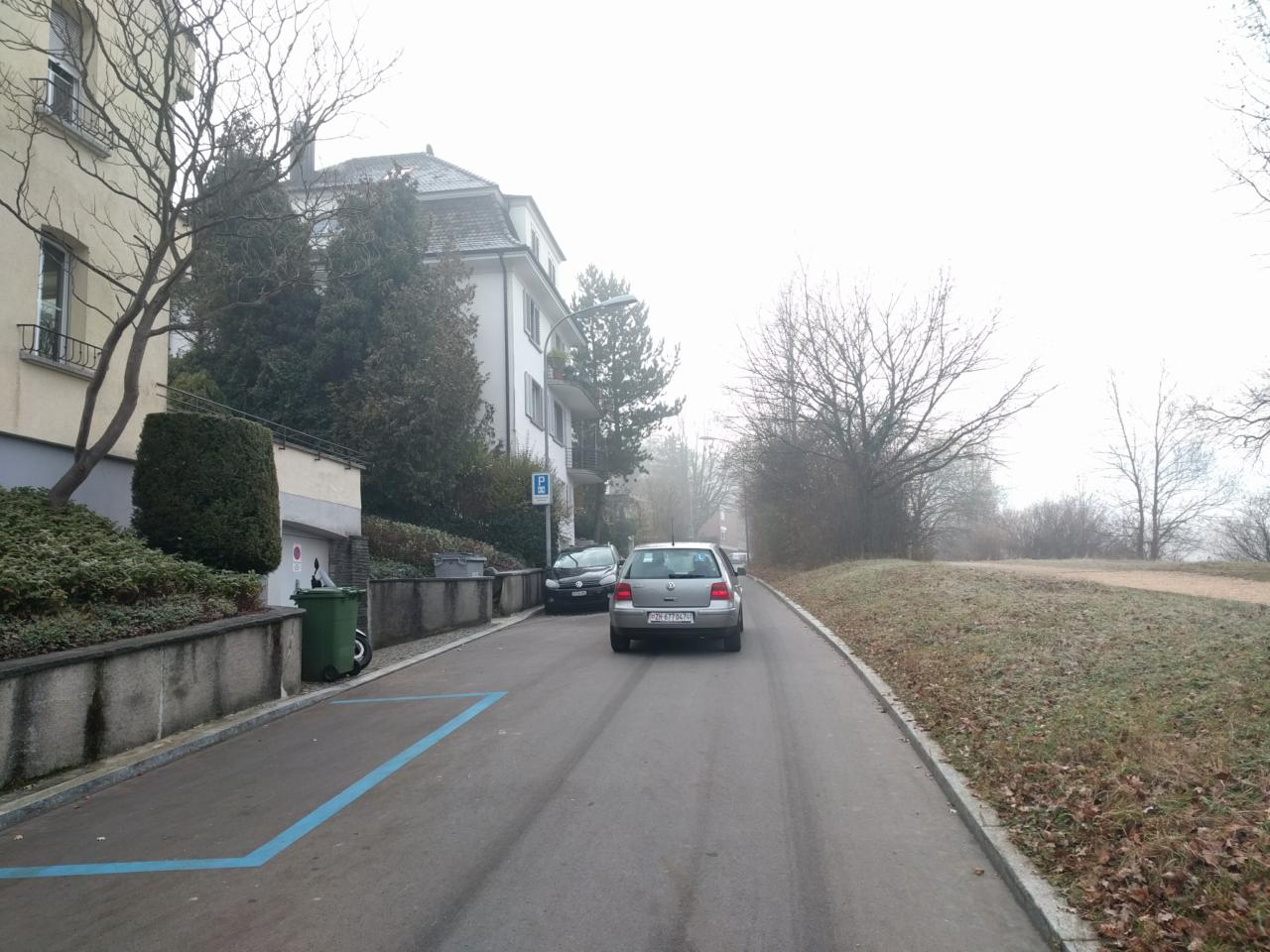}}
  \hfil
  \subfloat{\includegraphics[width=0.24\textwidth]{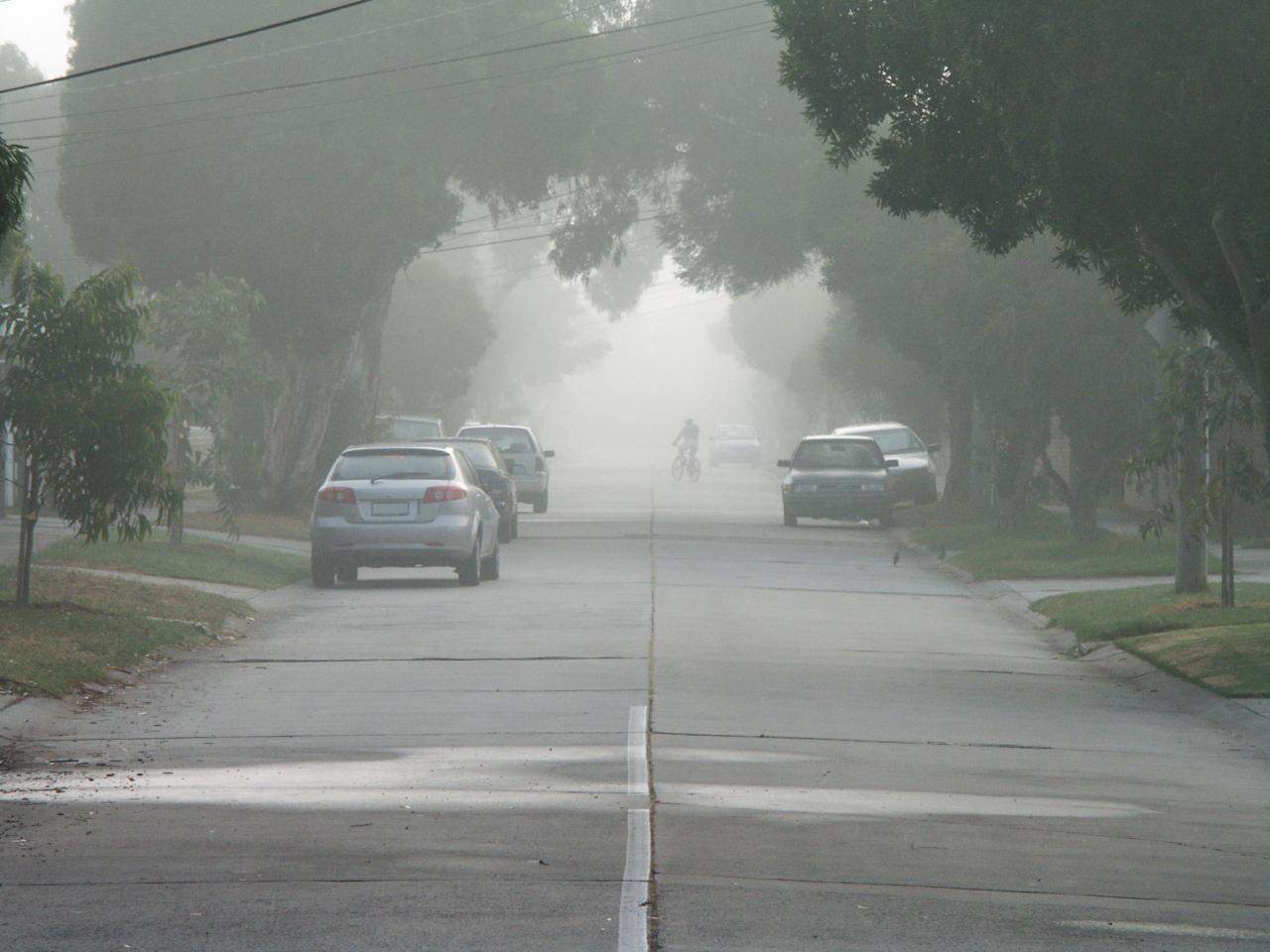}}
  \hfil
  \subfloat{\includegraphics[width=0.24\textwidth,height=0.18\textwidth]{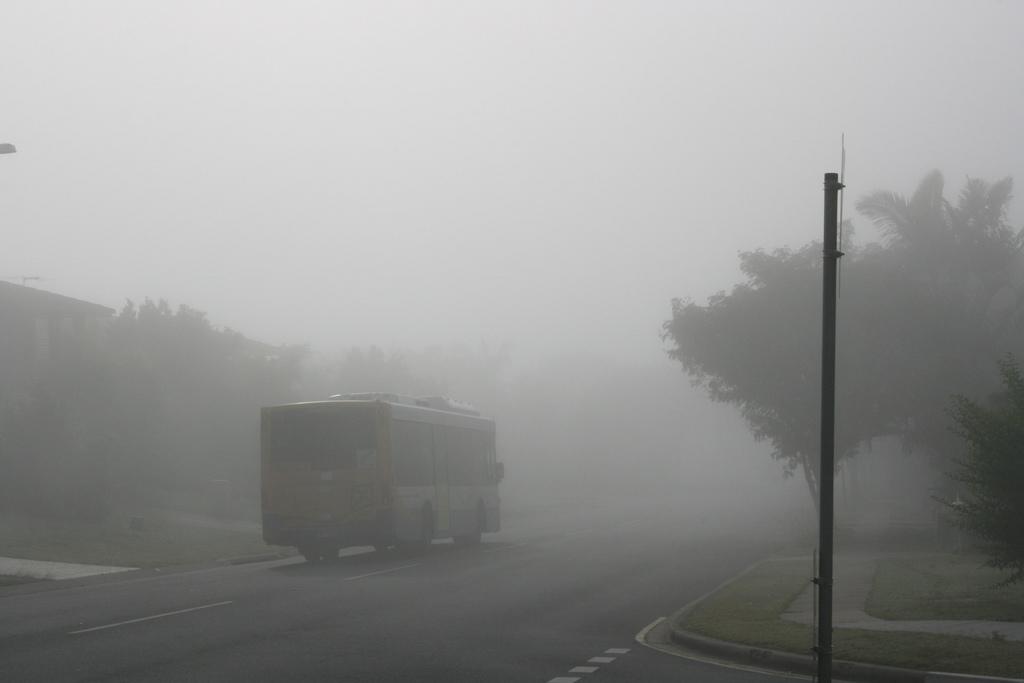}}\\  \vspace{-2.5mm}
  \subfloat{\rotatebox{90}{Baseline~\cite{refinenet}}}
      \hfil
  \subfloat{\includegraphics[width=0.24\textwidth]{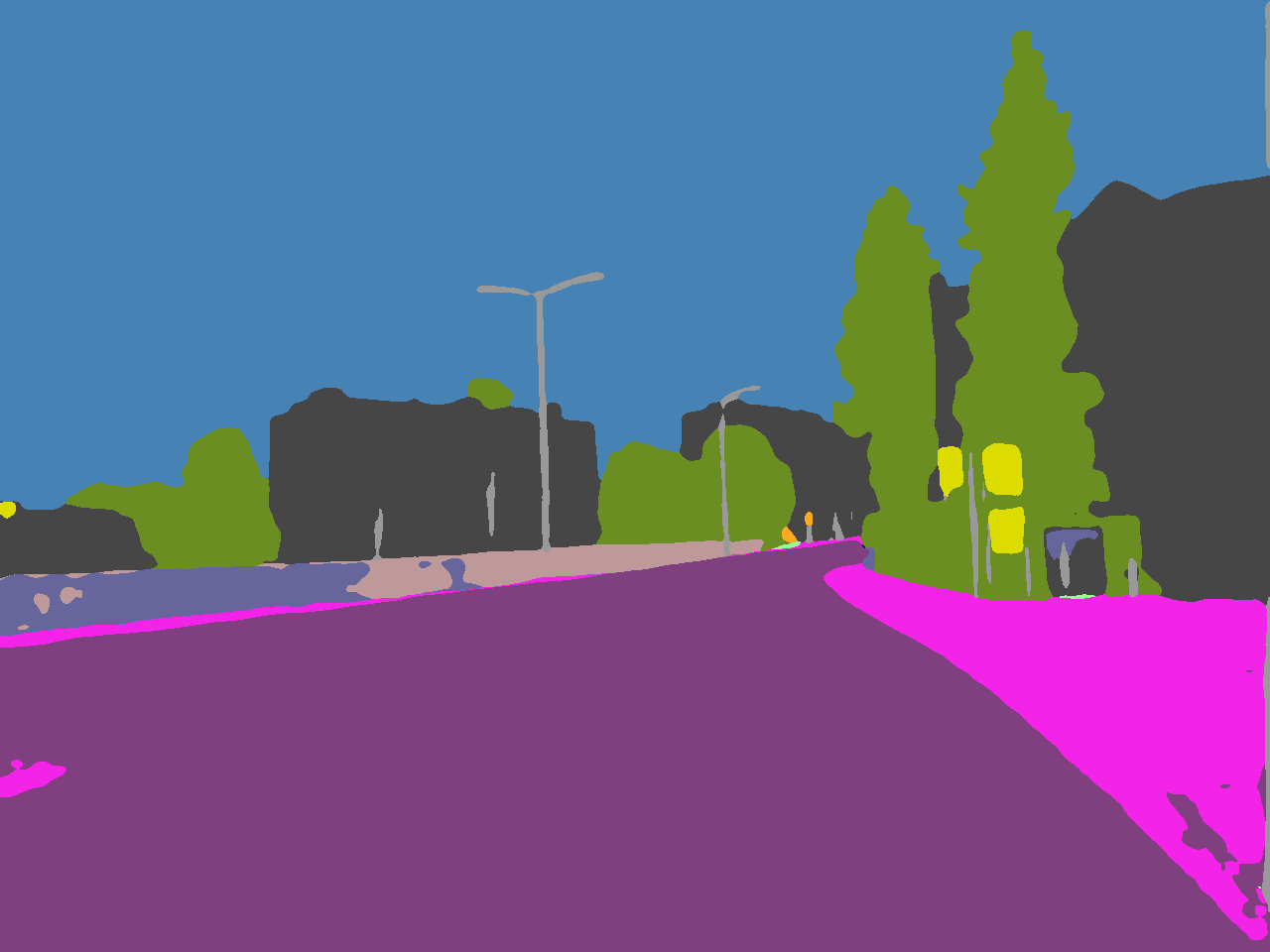}}
  \hfil
  \subfloat{\includegraphics[width=0.24\textwidth]{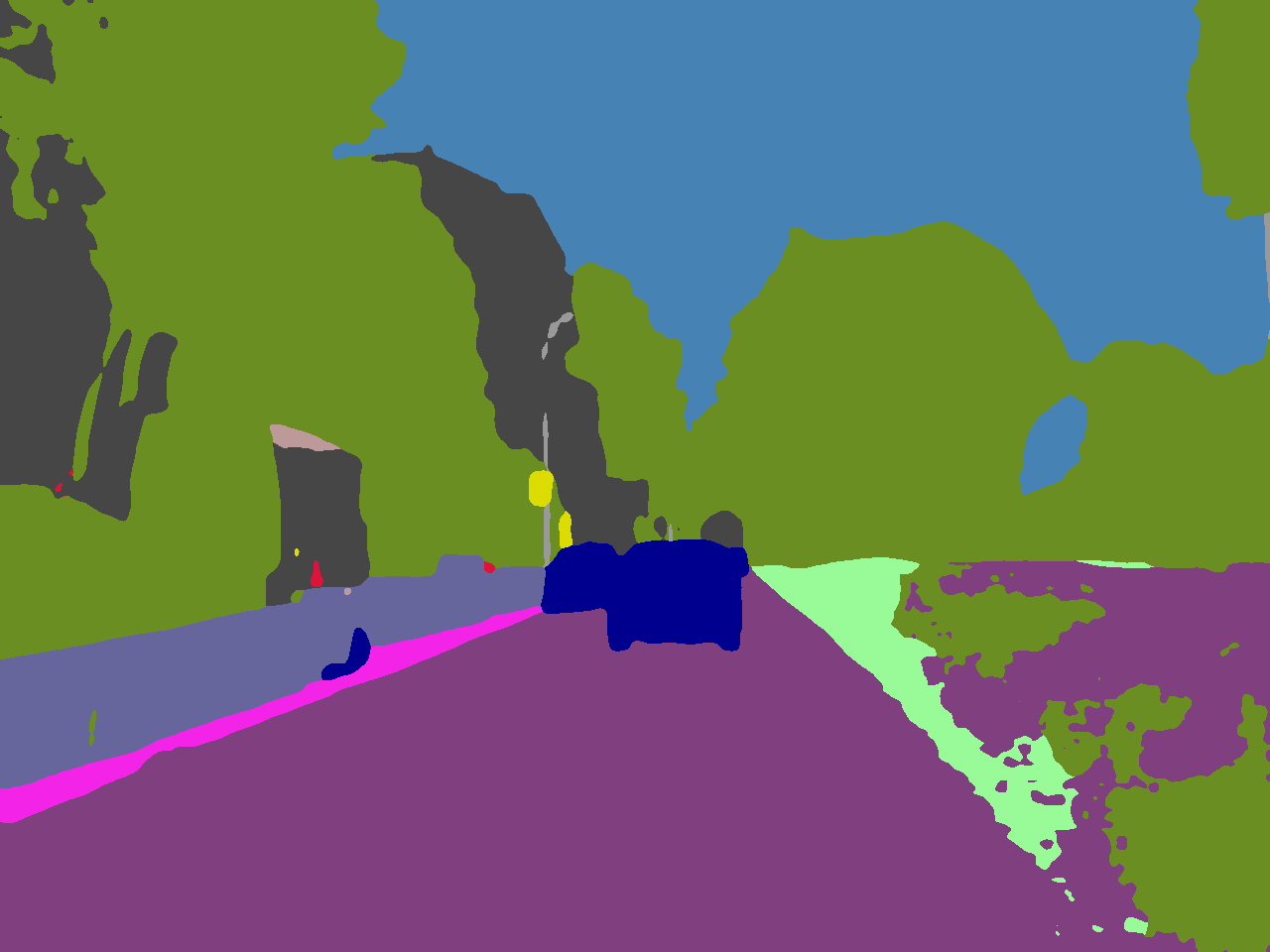}}
  \hfil
  \subfloat{\includegraphics[width=0.24\textwidth]{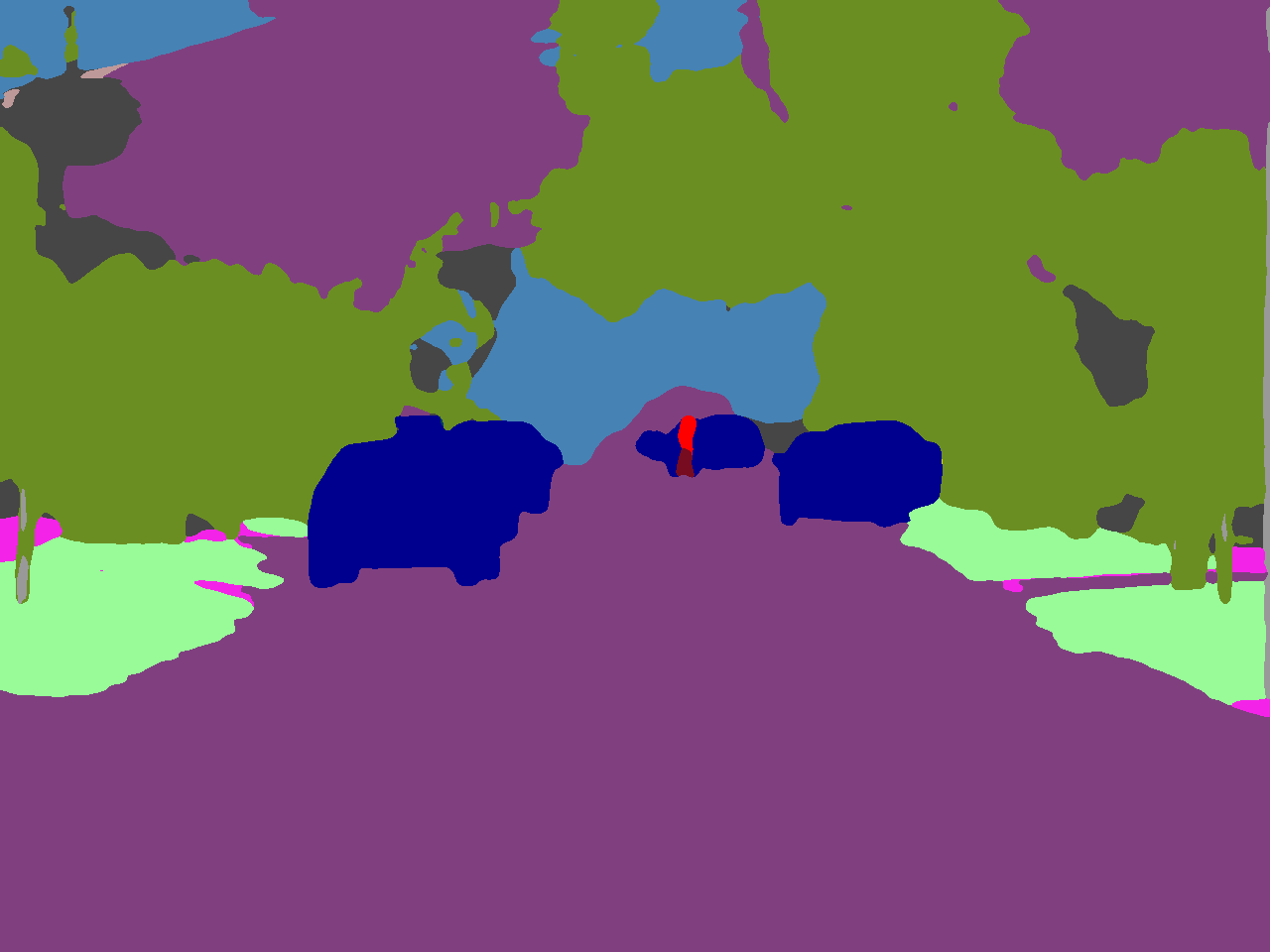}}
  \hfil
  \subfloat{\includegraphics[width=0.24\textwidth,height=0.18\textwidth]{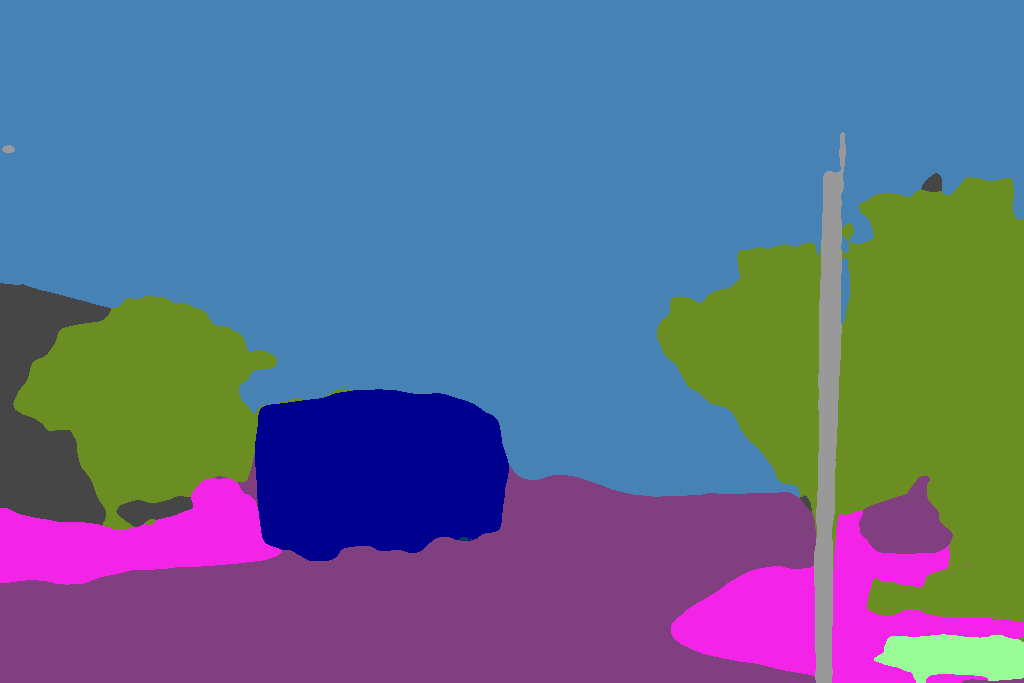}}\\ \vspace{-2.5mm}
  \subfloat{\rotatebox{90}{CMAda1 (ours)}} 
  \hfil
  \subfloat{\includegraphics[width=0.24\textwidth]{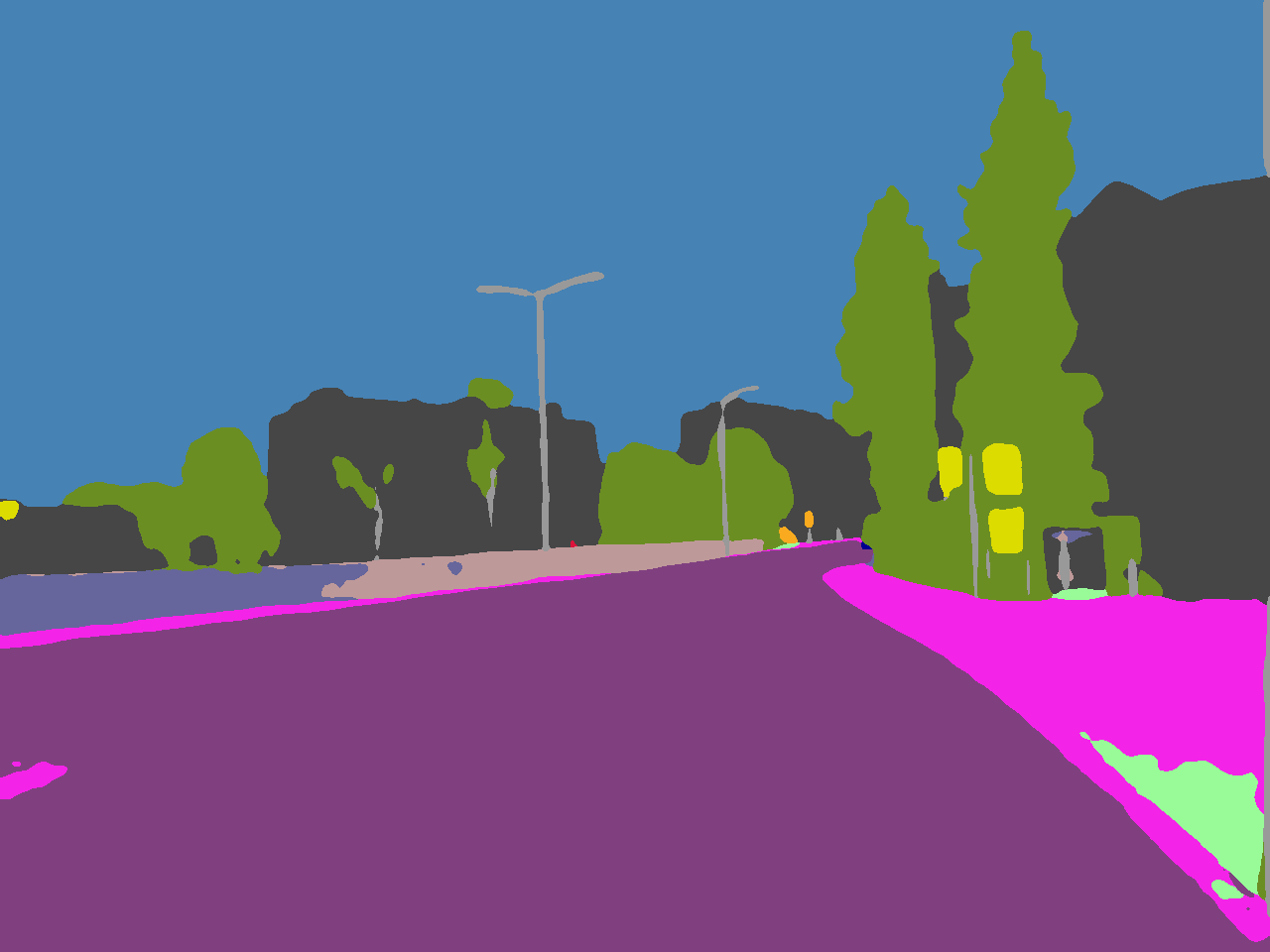}}
  \hfil
  \subfloat{\includegraphics[width=0.24\textwidth]{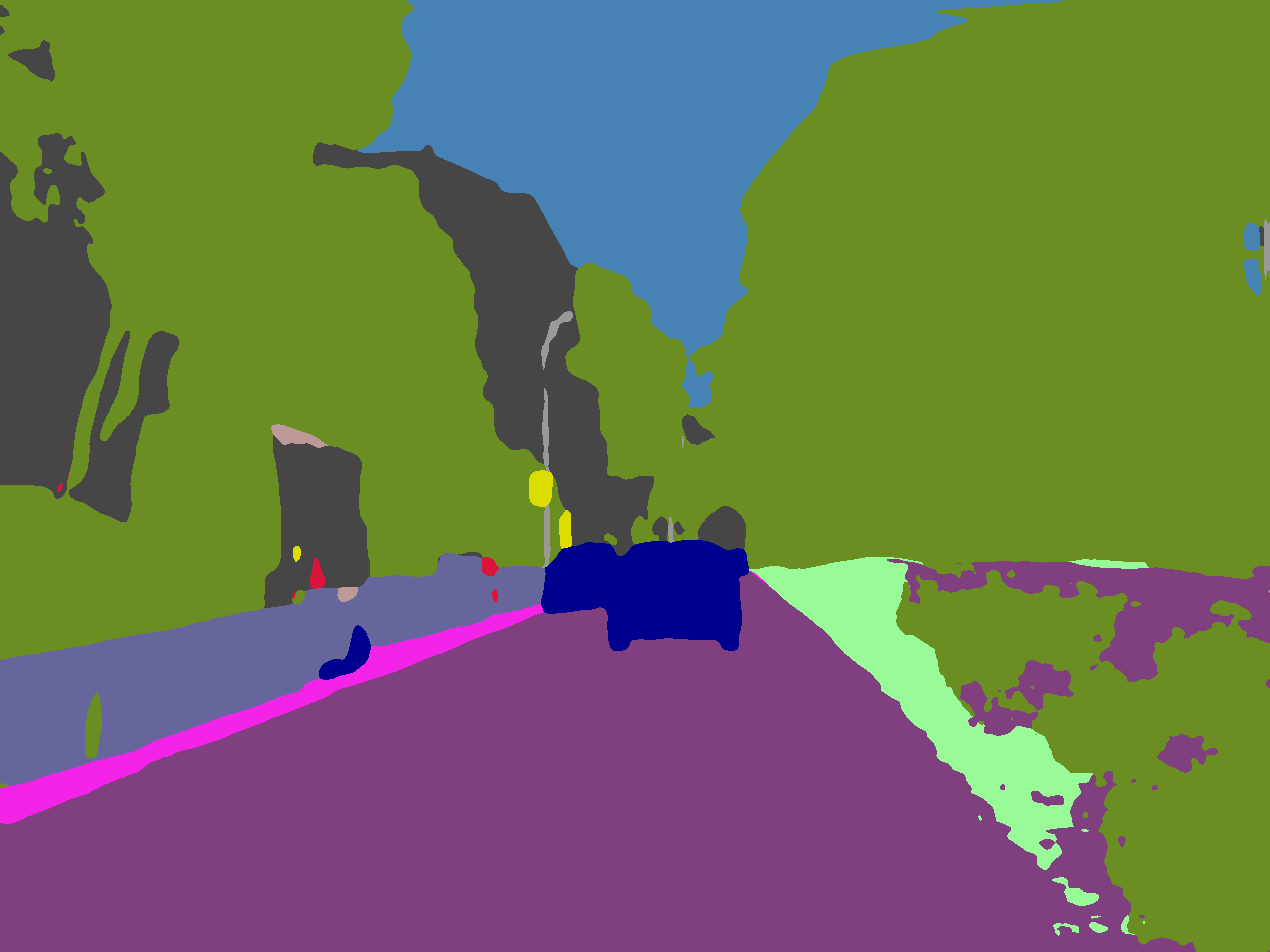}}
  \hfil
  \subfloat{\includegraphics[width=0.24\textwidth]{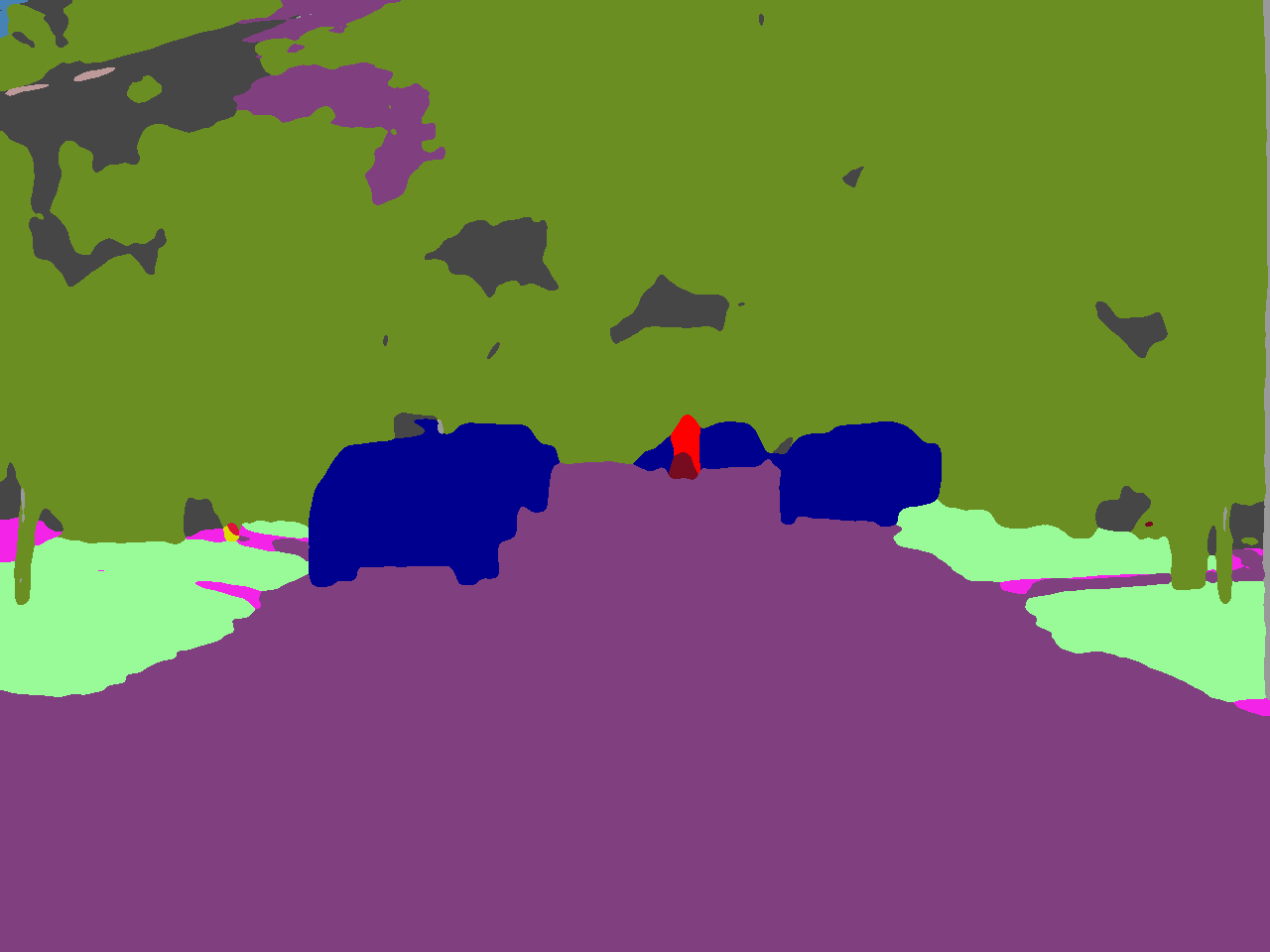}}
  \hfil 
  \subfloat{\includegraphics[width=0.24\textwidth,height=0.18\textwidth]{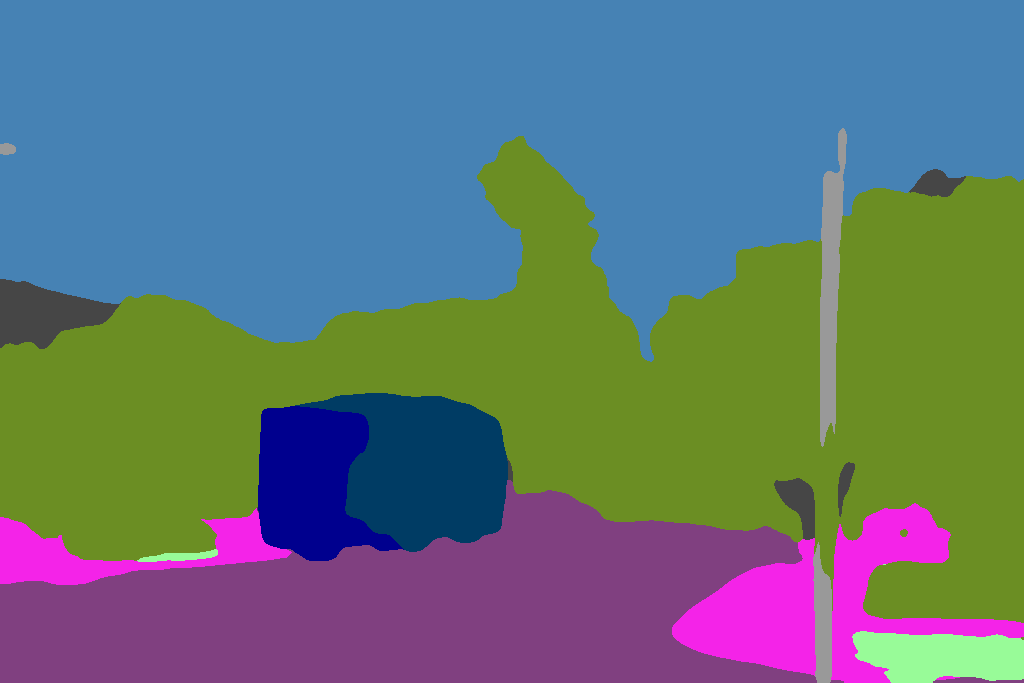}}\\  \vspace{-2.5mm}
   \subfloat{\rotatebox{90}{CMAda2 (ours)}}
       \hfil
  \subfloat{\includegraphics[width=0.24\textwidth]{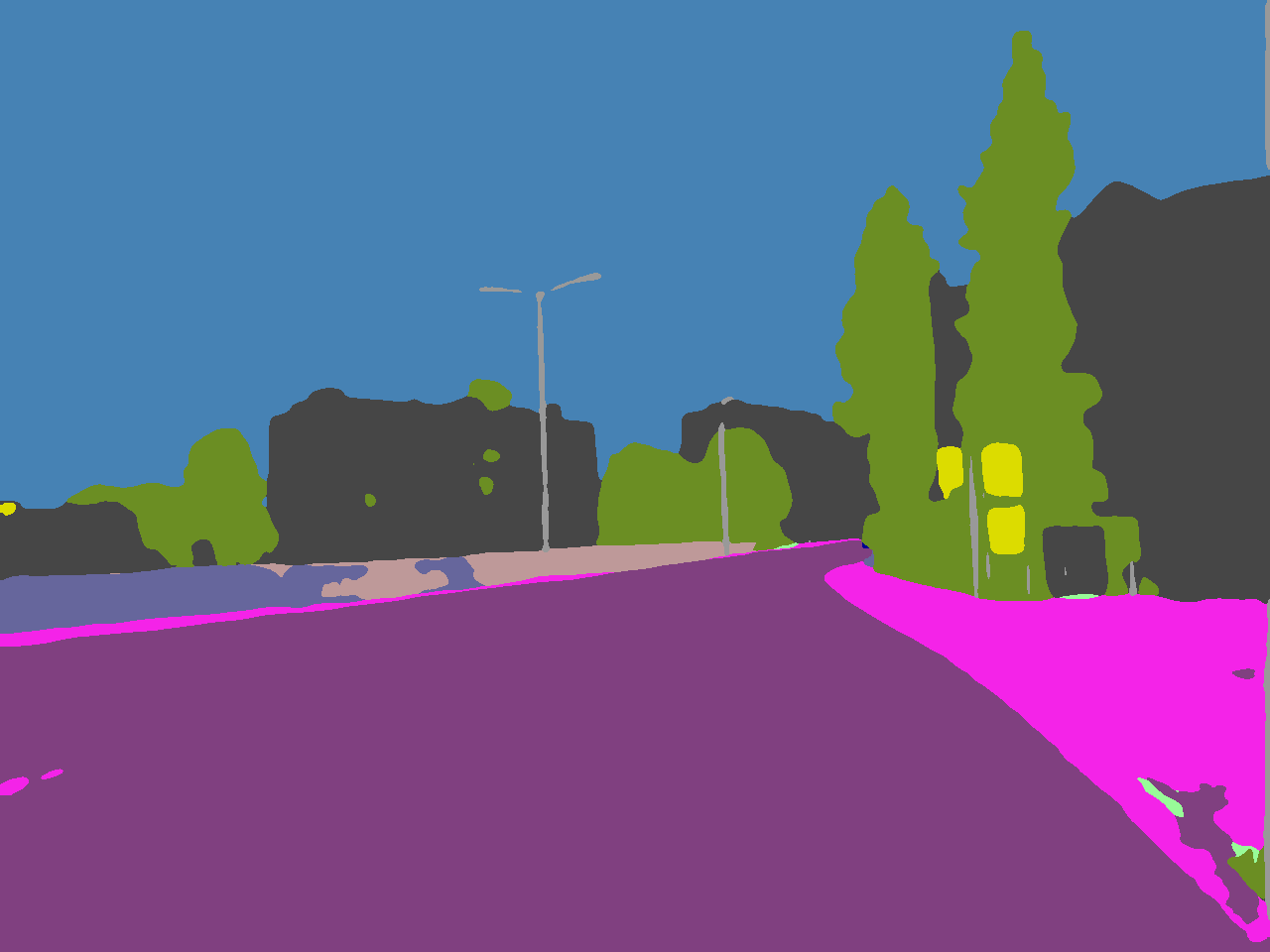}}
  \hfil
  \subfloat{\includegraphics[width=0.24\textwidth]{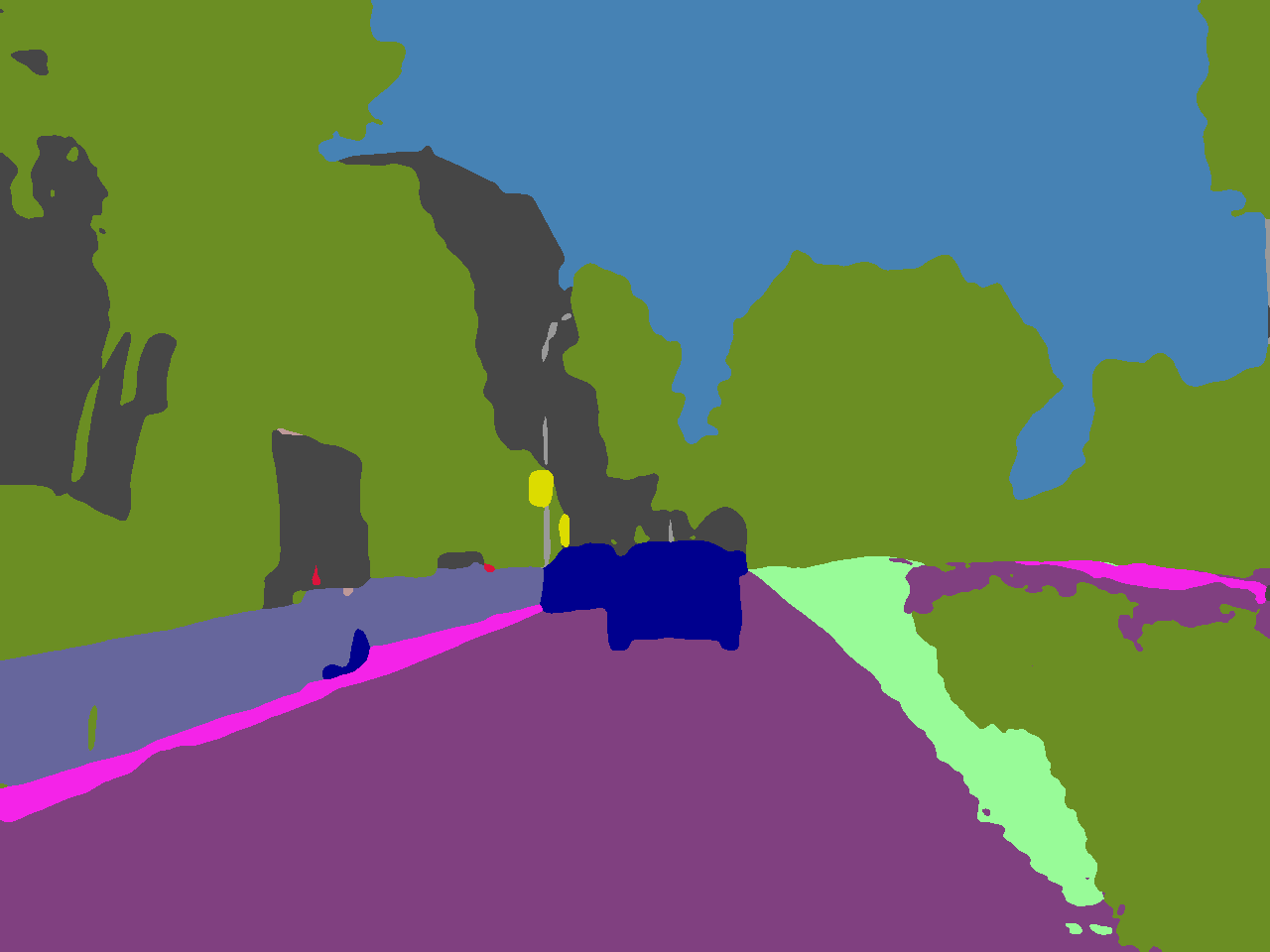}}
  \hfil
  \subfloat{\includegraphics[width=0.24\textwidth]{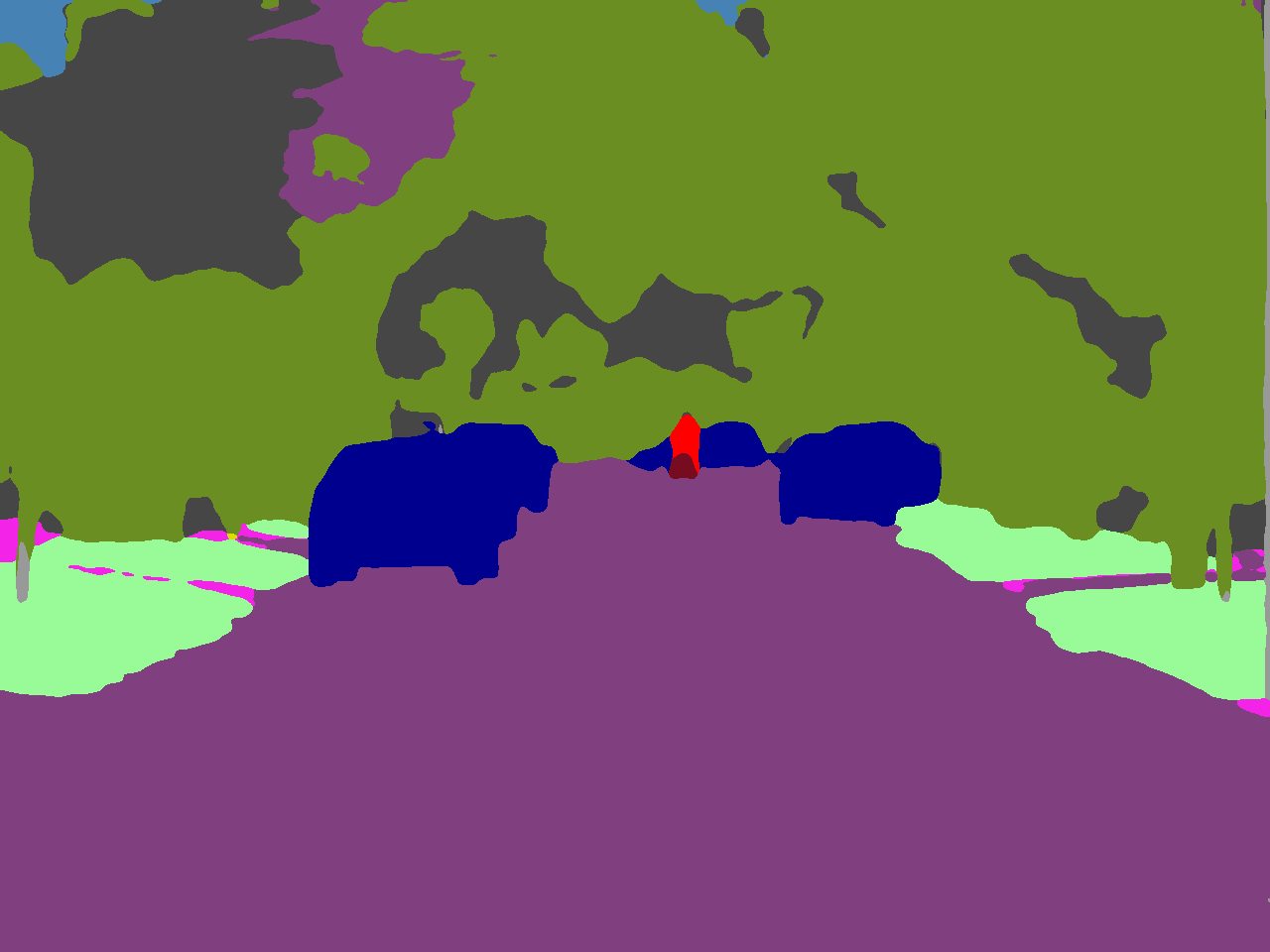}}
  \hfil 
  \subfloat{\includegraphics[width=0.24\textwidth,height=0.18\textwidth]{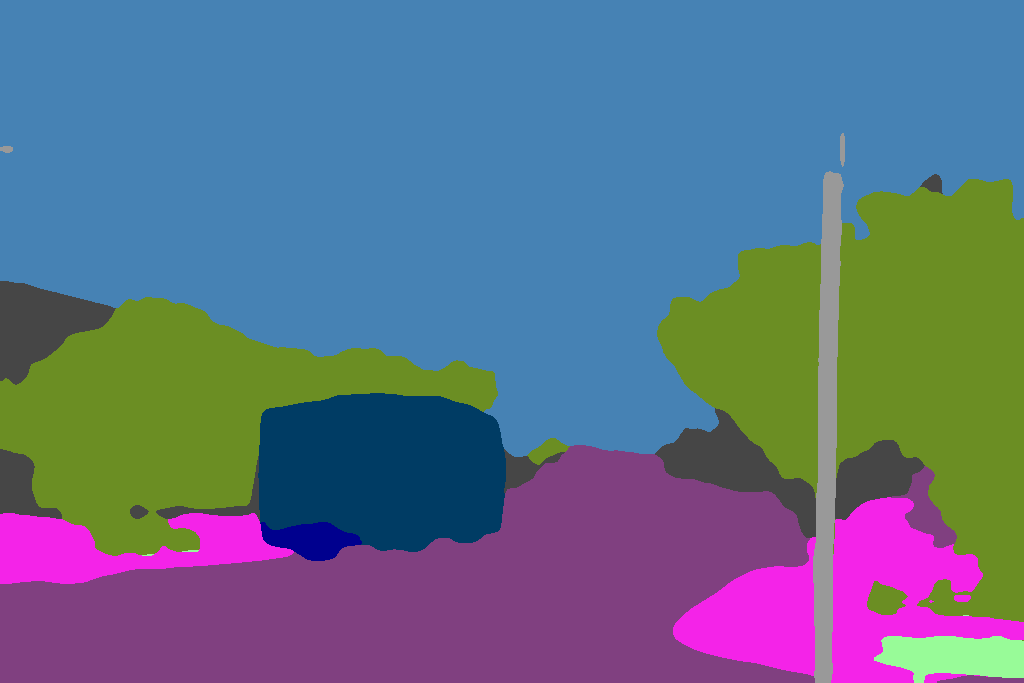}}\\  \vspace{-2.5mm}
   \subfloat{\rotatebox{90}{CMAda3+ (ours)}}
  \hfil
   \subfloat{\includegraphics[width=0.24\textwidth]{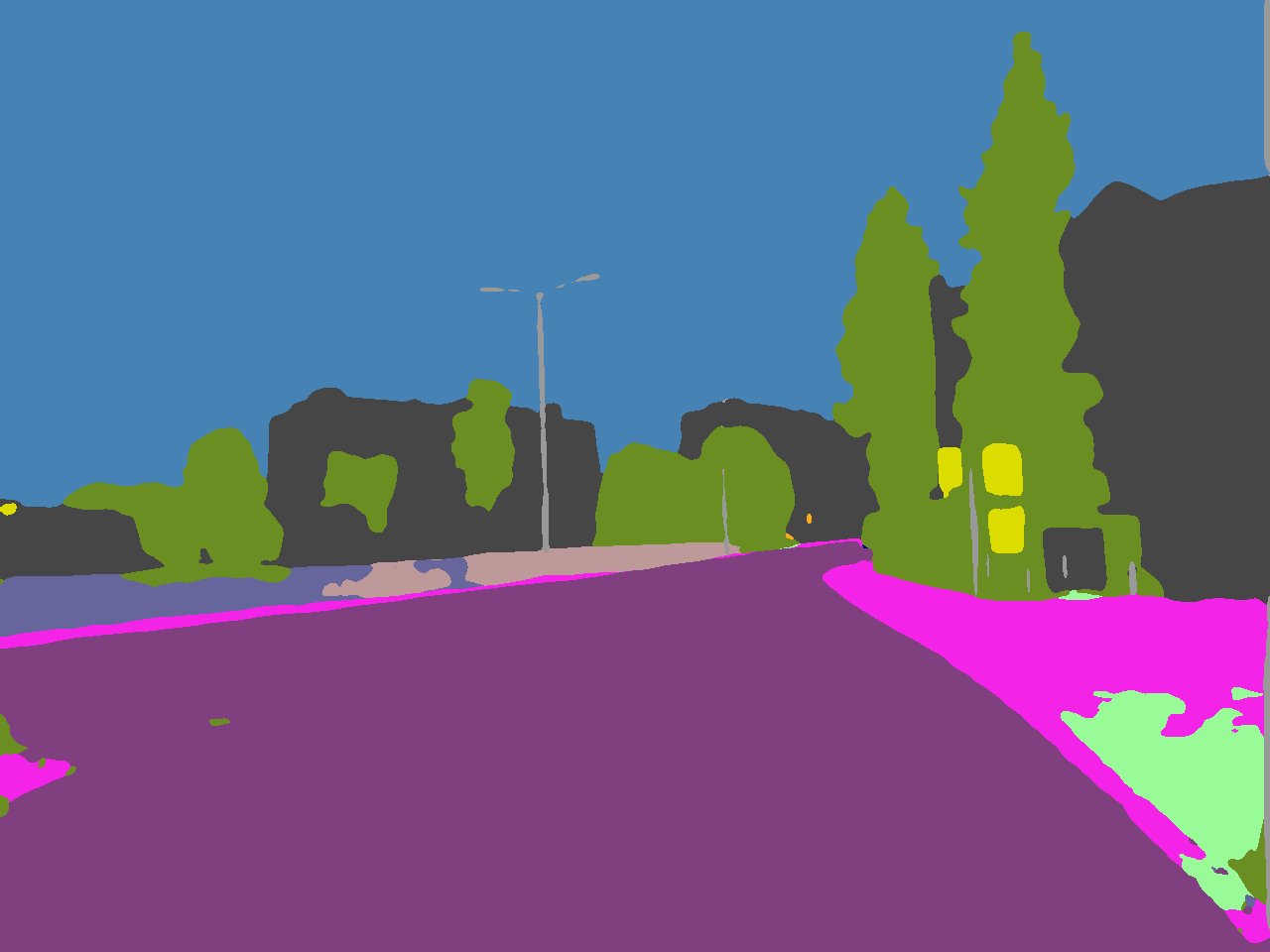}}
  \hfil
  \subfloat{\includegraphics[width=0.24\textwidth]{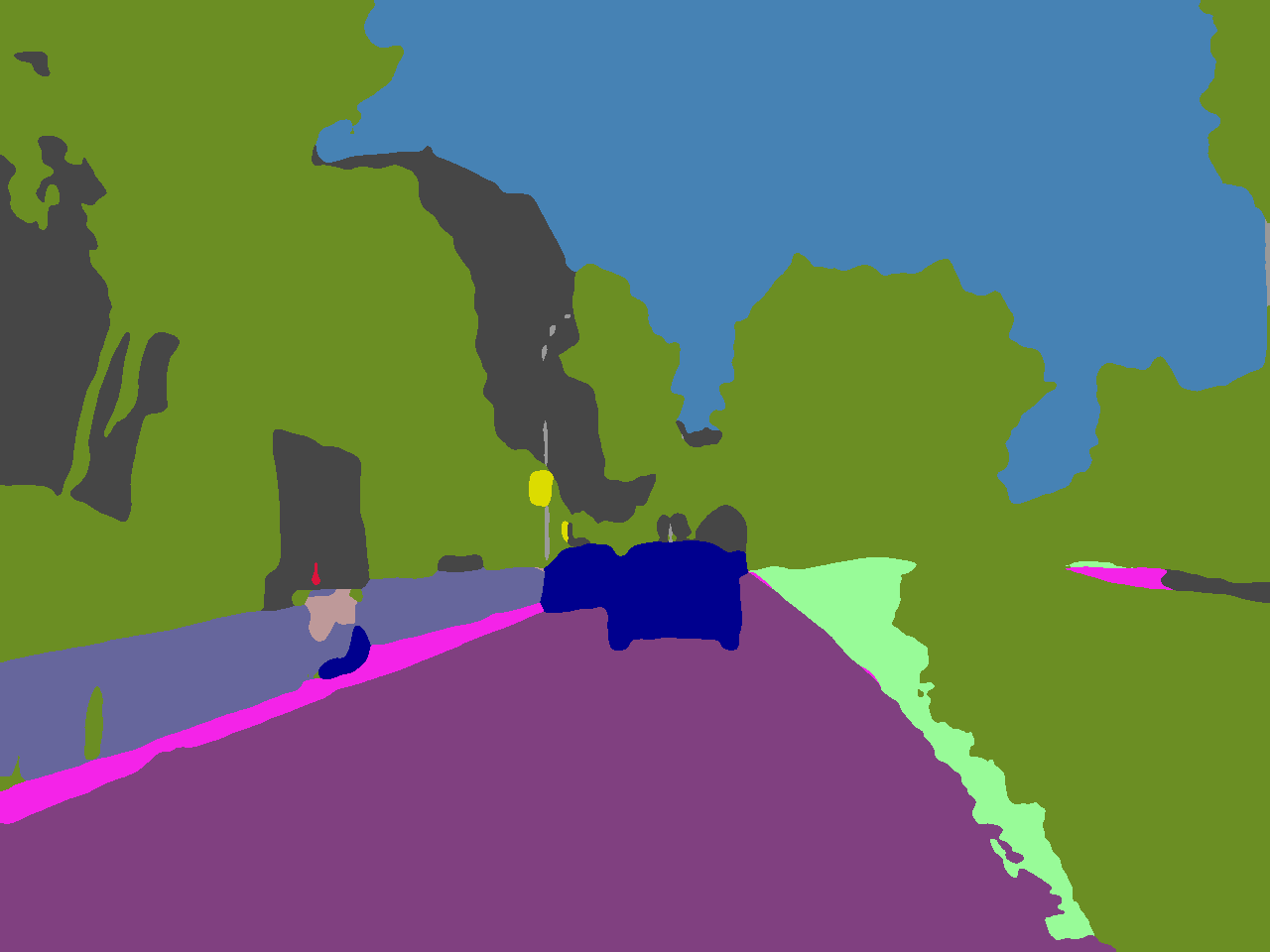}}
  \hfil
  \subfloat{\includegraphics[width=0.24\textwidth]{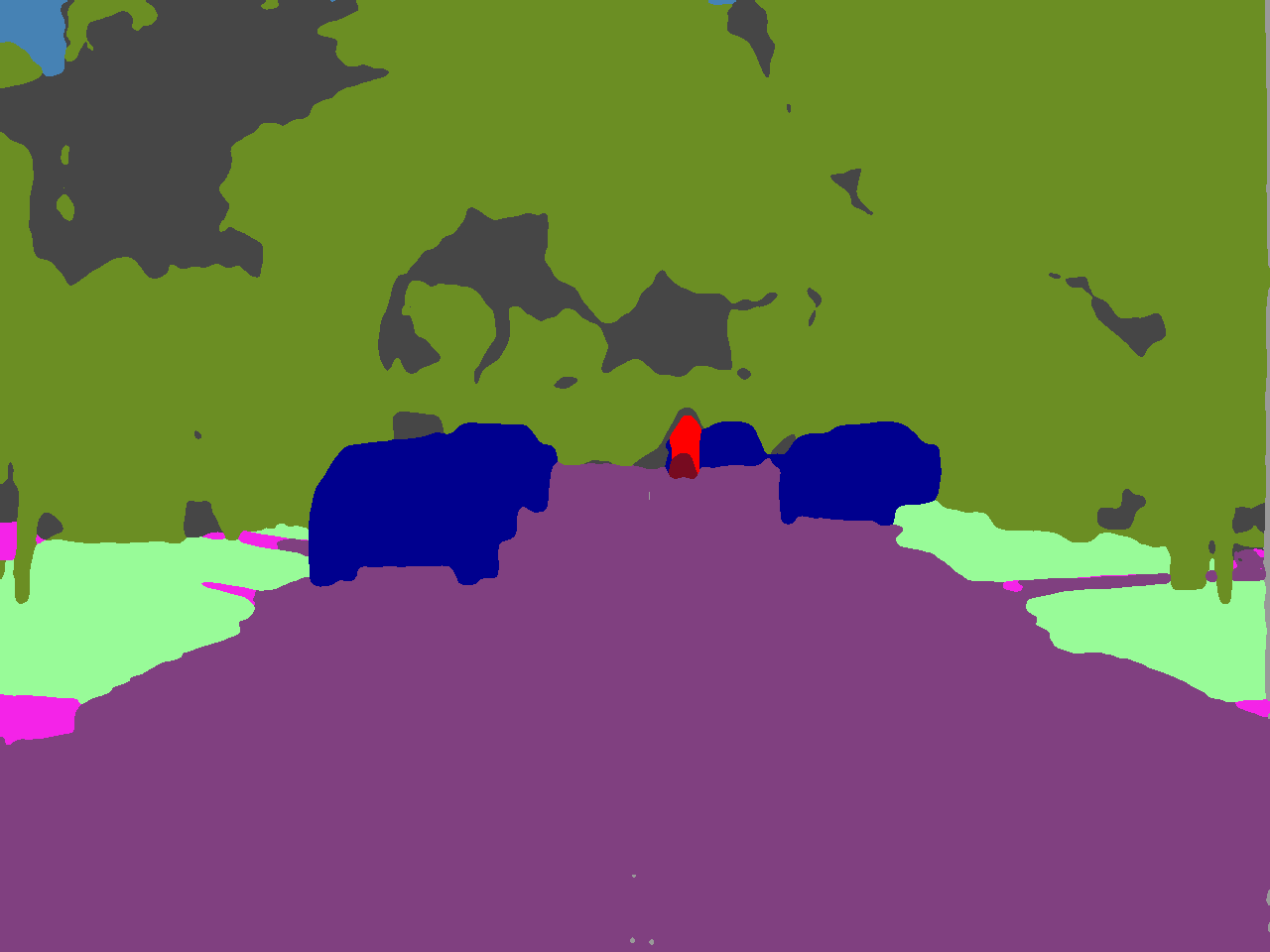}}
  \hfil 
  \subfloat{\includegraphics[width=0.24\textwidth,height=0.18\textwidth]{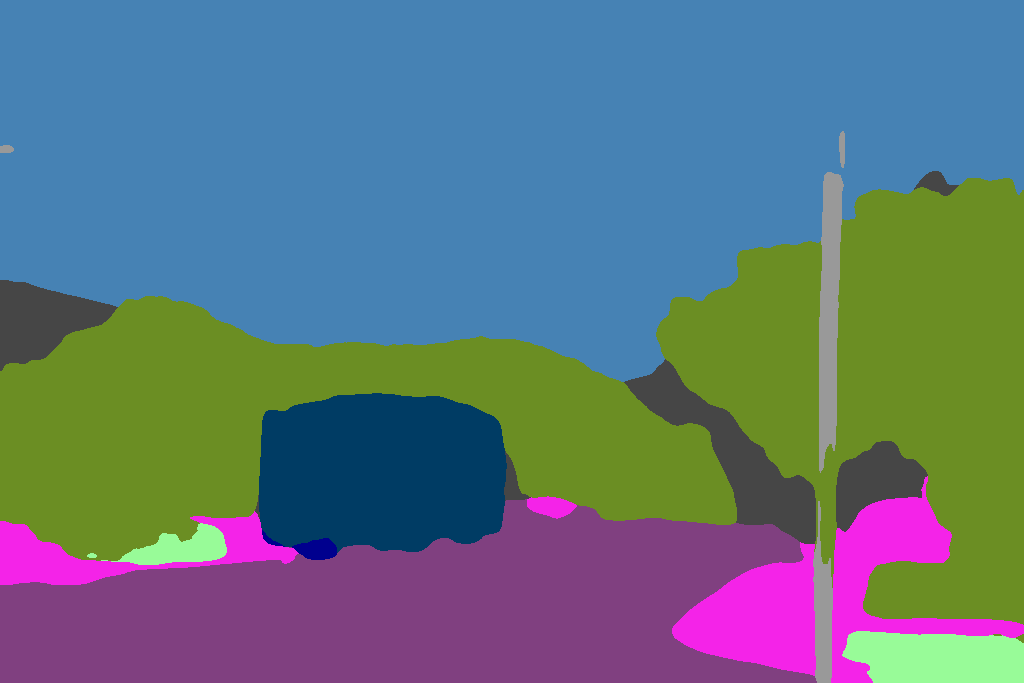}}\\  \vspace{-2.5mm} 
     \subfloat{\rotatebox{90}{Ground Truth}}
  \hfil
   \subfloat{\includegraphics[width=0.24\textwidth]{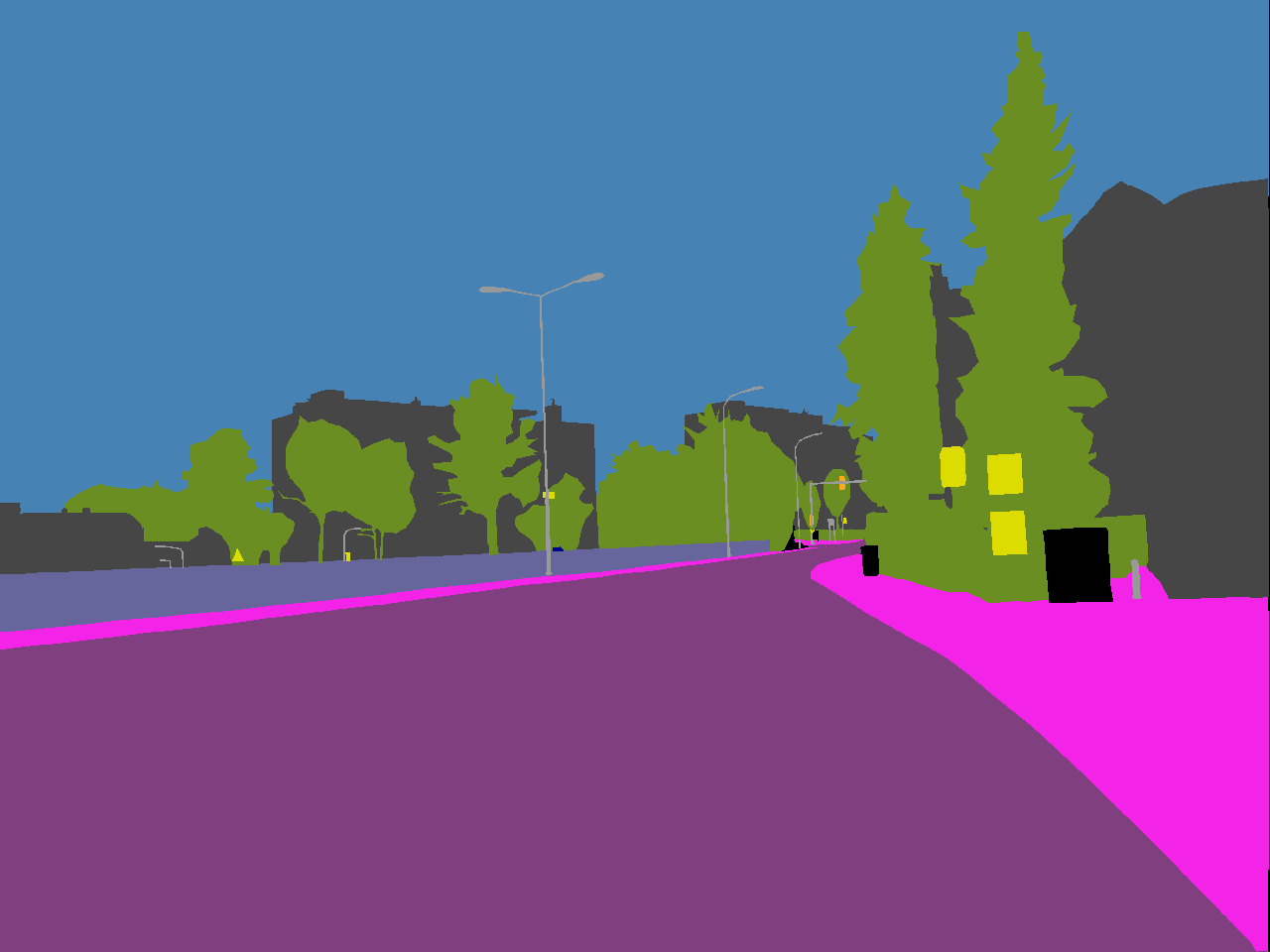}}
  \hfil
  \subfloat{\includegraphics[width=0.24\textwidth]{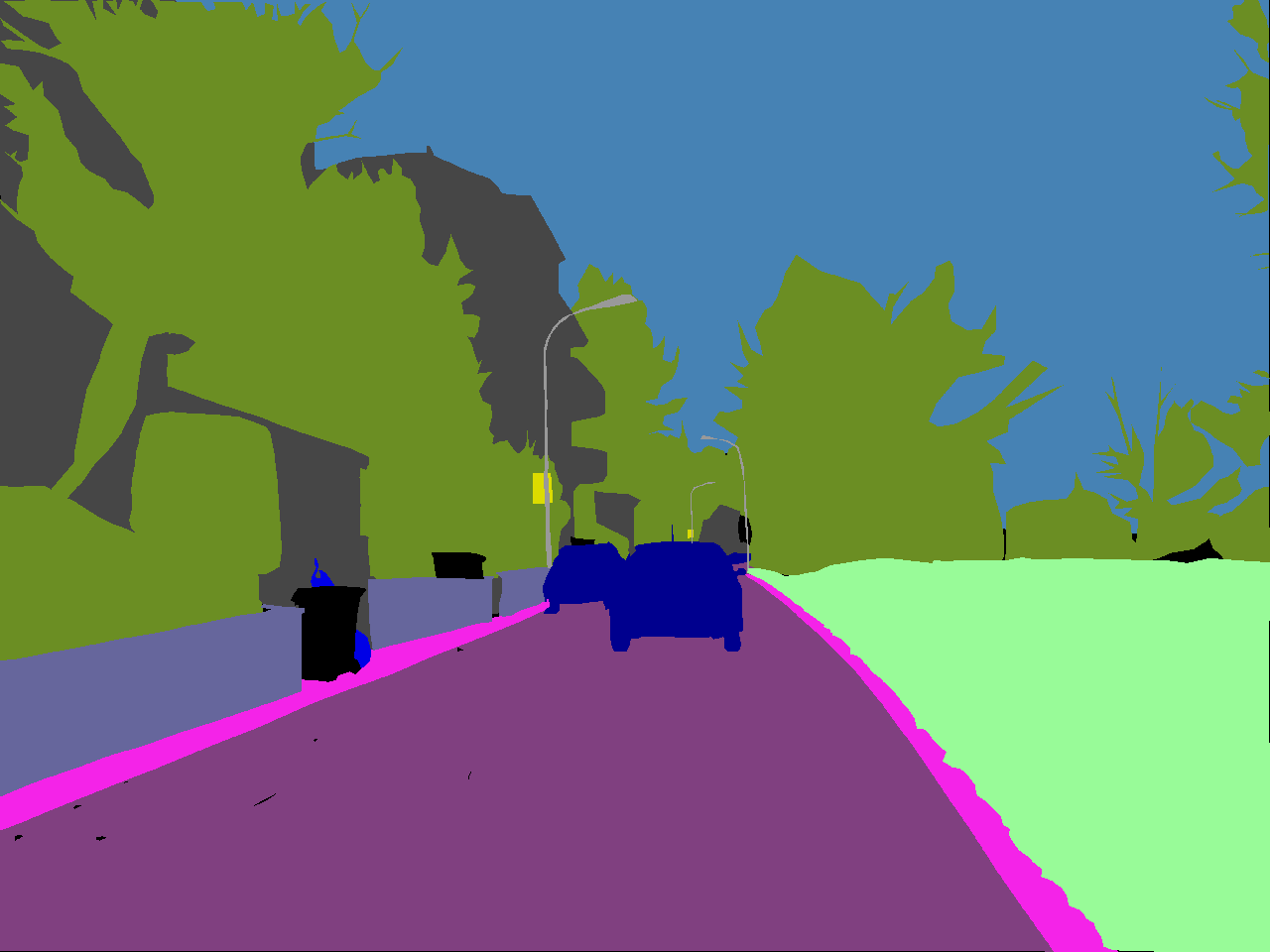}}
  \hfil
  \subfloat{\includegraphics[width=0.24\textwidth]{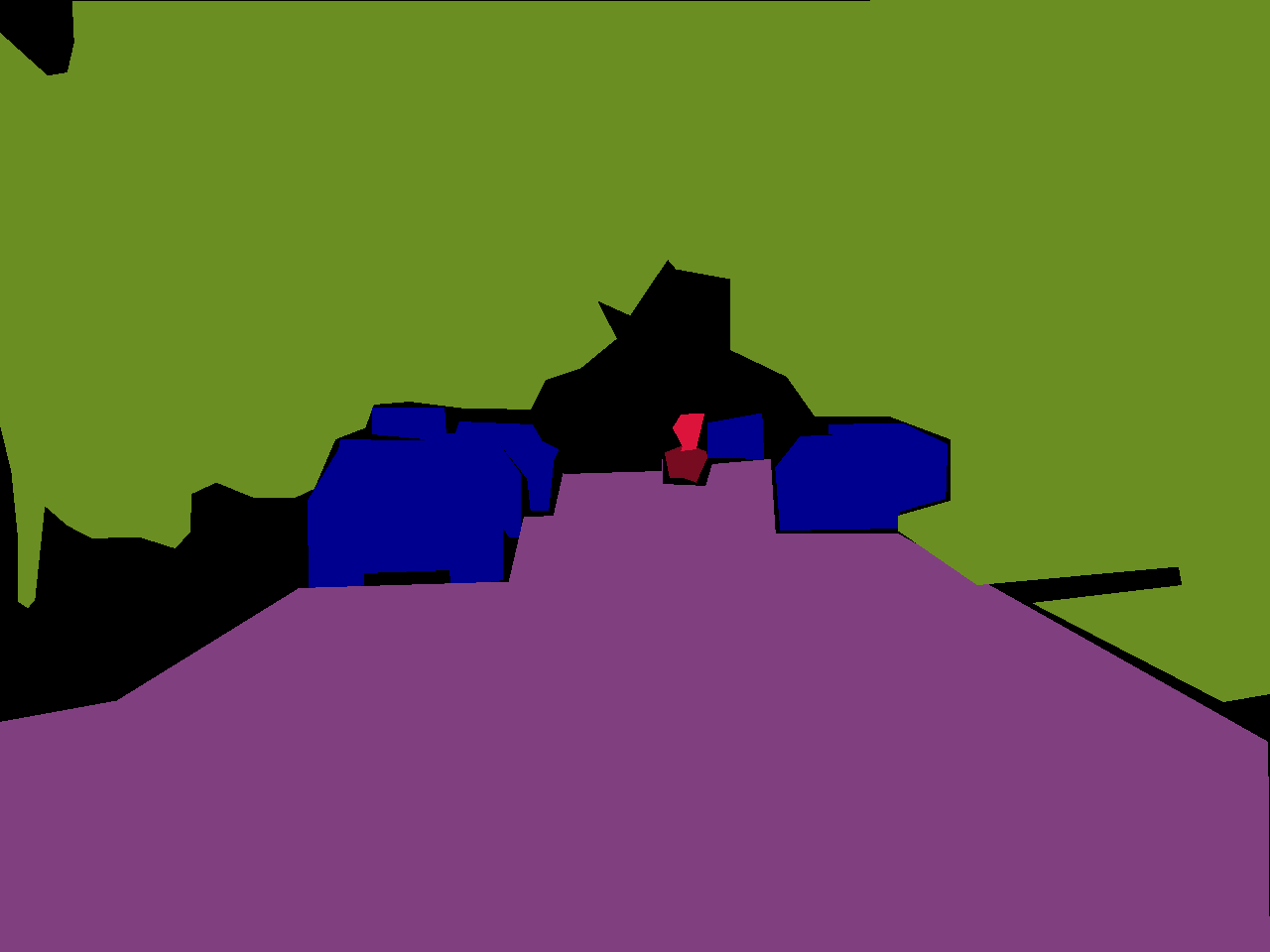}}
  \hfil 
  \subfloat{\includegraphics[width=0.24\textwidth,height=0.18\textwidth]{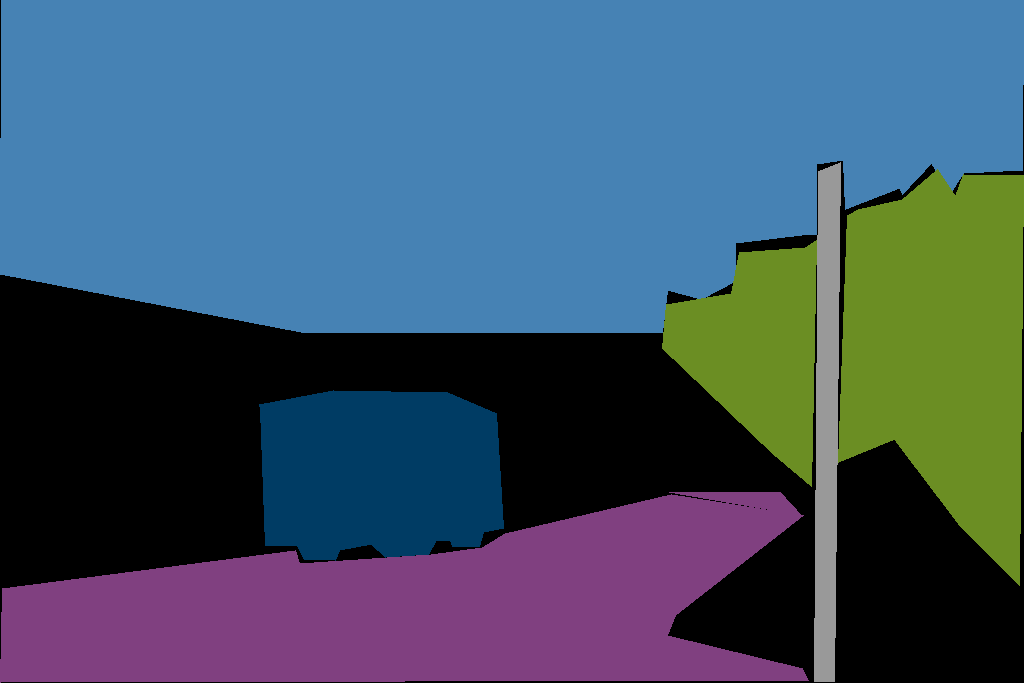}}\\ 
  \resizebox{\linewidth}{!}{
  \begin{tikzpicture}[tight background, scale=0.75, every node/.style={font=\large}]
      \draw[white, fill=void, draw=white] (0,0) rectangle (1 * 4, 1) node[pos=0.5] {Void};
      \draw[white, fill=road, draw=white] (1 * 4,0) rectangle (2 * 4, 1) node[pos=0.5] {Road};
      \draw[white, fill=sidewalk, draw=white] (2 * 4,0) rectangle (3 * 4, 1) node[pos=0.5] {Sidewalk};
      \draw[white, fill=building, draw=white] (3 * 4,0) rectangle (4 * 4, 1) node[pos=0.5] {Building};
      \draw[white, fill=wall, draw=white] (4 * 4,-0) rectangle (5 * 4, 1) node[pos=0.5] {Wall};
      \draw[black, fill=fence, draw=white] (5 * 4,-0) rectangle (6 * 4, 1) node[pos=0.5] {Fence};
      \draw[white, fill=pole, draw=white] (6 * 4,-0) rectangle (7 * 4, 1) node[pos=0.5] {Pole};
      \draw[white, fill=traffic light, draw=white] (7 * 4,-0) rectangle (8 * 4, 1) node[pos=0.5] {Traffic Light};
      \draw[black, fill=traffic sign, draw=white] (8 * 4,-0) rectangle (9 * 4, 1) node[pos=0.5] {Traffic Sign};
      \draw[white, fill=vegetation, draw=white] (9 * 4,-0) rectangle (10 * 4, 1) node[pos=0.5] {Vegetation};

      \draw[black, fill=terrain, draw=white] (0 * 4,-1) rectangle (1 * 4, 0) node[pos=0.5] {Terrain};
      \draw[white, fill=sky, draw=white] (1 * 4,-1) rectangle (4 * 2, 0) node[pos=0.5] {Sky};
      \draw[white, fill=person, draw=white] (2 * 4,-1) rectangle (3 * 4, 0) node[pos=0.5] {Person};
      \draw[white, fill=rider, draw=white] (3 * 4,-1) rectangle (4 * 4, 0) node[pos=0.5] {Rider};
      \draw[white, fill=car, draw=white] (4 * 4,-1) rectangle (5 * 4, 0) node[pos=0.5] {Car};
      \draw[white, fill=truck, draw=white] (5 * 4,-1) rectangle (6 * 4, 0) node[pos=0.5] {Truck};
      \draw[white, fill=bus, draw=white] (6 * 4,-1) rectangle (7 * 4, 0) node[pos=0.5] {Bus};
      \draw[white, fill=train, draw=white] (7 * 4,-1) rectangle (8 * 4, 0) node[pos=0.5] {Train};
      \draw[white, fill=motorcycle, draw=white] (8 * 4,-1) rectangle (9 * 4, 0) node[pos=0.5] {Motorcycle};
      \draw[white, fill=bicycle, draw=white] (9 * 4,-1) rectangle (10 * 4, 0) node[pos=0.5] {Bicycle};
  \end{tikzpicture}}
\caption{Qualitative semantic segmentation results on images from \emph{Foggy Driving} with varying fog density. Foggy images in the top row are sorted from left to right in ascending order of estimated fog density using our estimator}
\label{fig:sem:seg:fog:density}
\end{figure*}

To further demonstrate the behavior of CMAda, we also show semantic segmentation results of the clear-weather RefineNet model~\cite{refinenet} and the three aforementioned variants of our method for variable fog density in Figure~\ref{fig:sem:seg:fog:density}. In particular, we have applied our fog density estimator to \emph{Foggy Driving} and use four images therefrom for which the estimated fog density ranges from very low to very high. First, we observe that the clear-weather baseline performs comparably well for very light fog due to the small domain shift from clear weather, but for higher fog densities CMAda variants outperform this baseline. The advantage gets more pronounced as fog density increases. Second, comparing the different CMAda variants, we conclude that having more adaptation stages leads to increasing returns as fog density increases. For instance, the bus in the highly foggy rightmost image is correctly recognized only after all three adaptation stages have been applied.
 
While we observe a significant improvement with CMAda, semantic segmentation performance on foggy scenes is still much worse than the reported performance by existing papers on clear-weather scenes. Foggy scenes are indeed more challenging than clear-weather scenes with respect to understanding their semantics. There are more underlying causal factors of variation that generated foggy data, which requires either more training data or more intelligent learning approaches to disentangle the increased degrees of freedom. While our method shows considerable improvement by transferring semantic knowledge from clear-weather to fog, the models are adapted in an ``unsupervised'' manner, \ie{}without using human annotations of real foggy data. Incorporating a moderate amount of human annotations of real foggy scenes into our learning approach is a promising research direction, if significantly better results are desired. 

Our method involves two data streams: partially synthetic data with annotations and real data without annotations. Learning from the real data stream is based on a ``self-learning'' mechanism, which creates a risk of entering a negative reinforcement loop by adapting to mistakes made at previous stages. In practice, we find that our training process is stable. 
In order to further investigate this, we follow the  literature~\cite{data:distillation:cvpr18} to identify and exclude the erroneous predictions from training. In particular, the confidence scores of the predictions are used as a proxy for prediction quality and we generate pseudo-labels only for pixels where this confidence is higher than a defined threshold. This prediction selection step, however, does not provide clear benefit and thus is not included in our approach. 

We believe that the low risk of entering the negative reinforcement loop and the steady improvement of our method can be ascribed to two factors: 1) the accurate human annotations of the partially synthetic data stream restrict the space of adapted models, ruling out solutions that would create severe errors in the inferred labels of the real data; and 2) each adaptation stage is initialized with the solution of the previous stage, which helps smoothly traverse the model space from the initial clear-weather model to the target foggy model.

\begin{figure}[!tb]
  \centering
  \addtocounter{subfigure}{-20}
    \subfloat{\rotatebox{90}{Images}}  
      \hfil
  \subfloat{\includegraphics[width=0.19\textwidth,height=0.1425\textwidth]{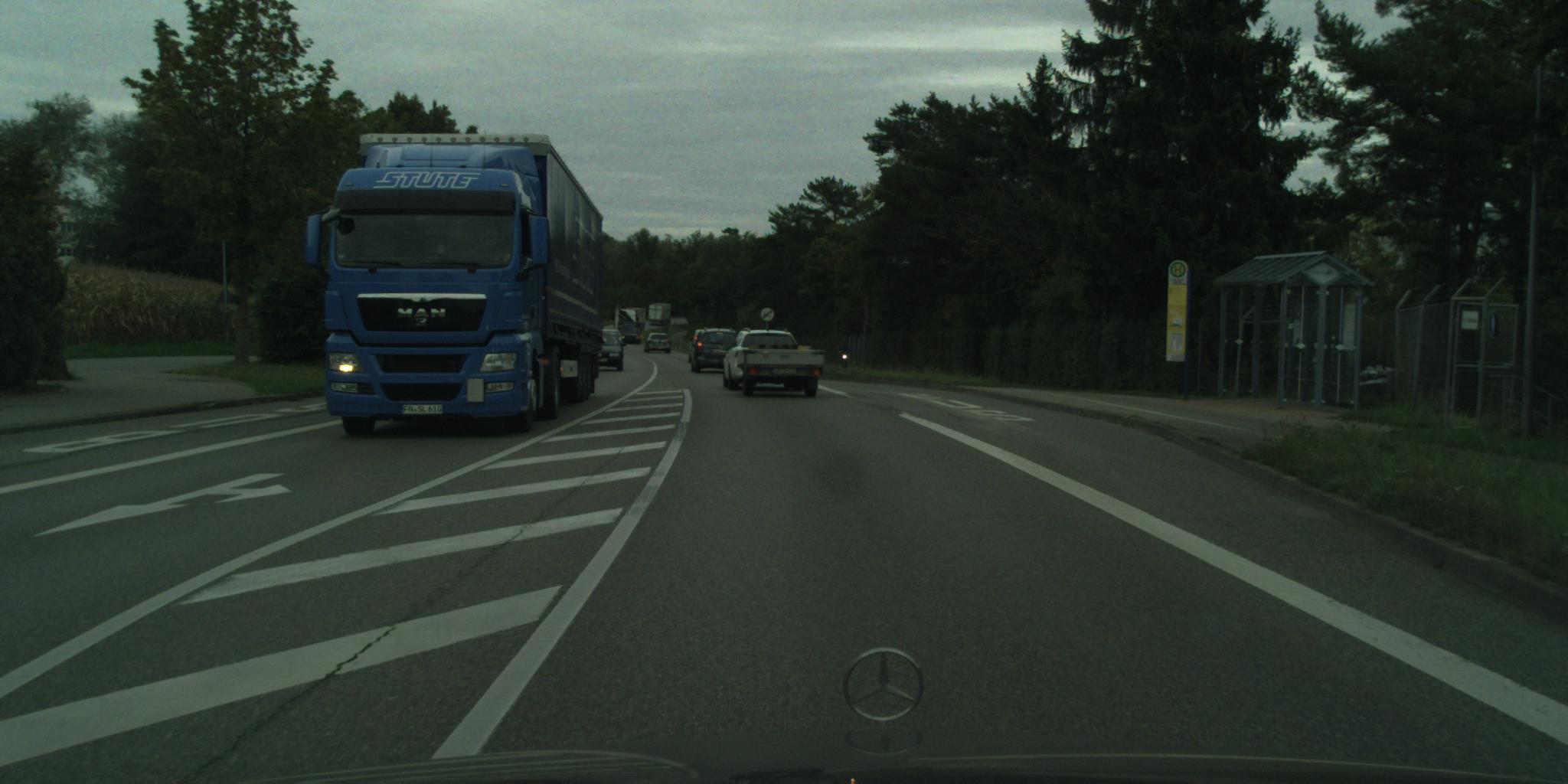}}
  \hfil
  \subfloat{\includegraphics[width=0.19\textwidth,height=0.1425\textwidth]{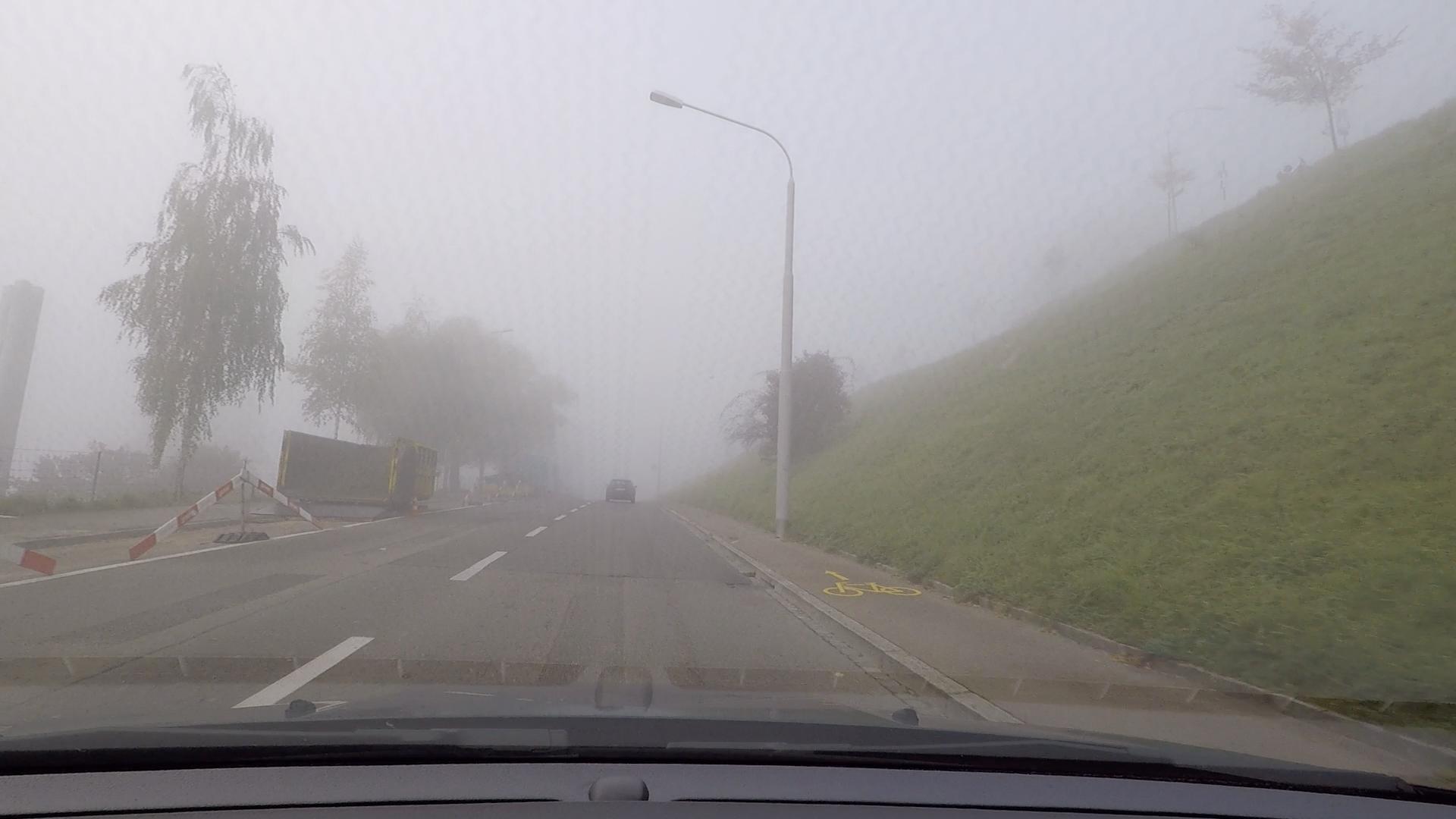}}\\  \vspace{-2.5mm}
  \subfloat{\rotatebox{90}{Baseline~\cite{refinenet}}}
    \hfil
  \subfloat{\includegraphics[width=0.19\textwidth,height=0.1425\textwidth]{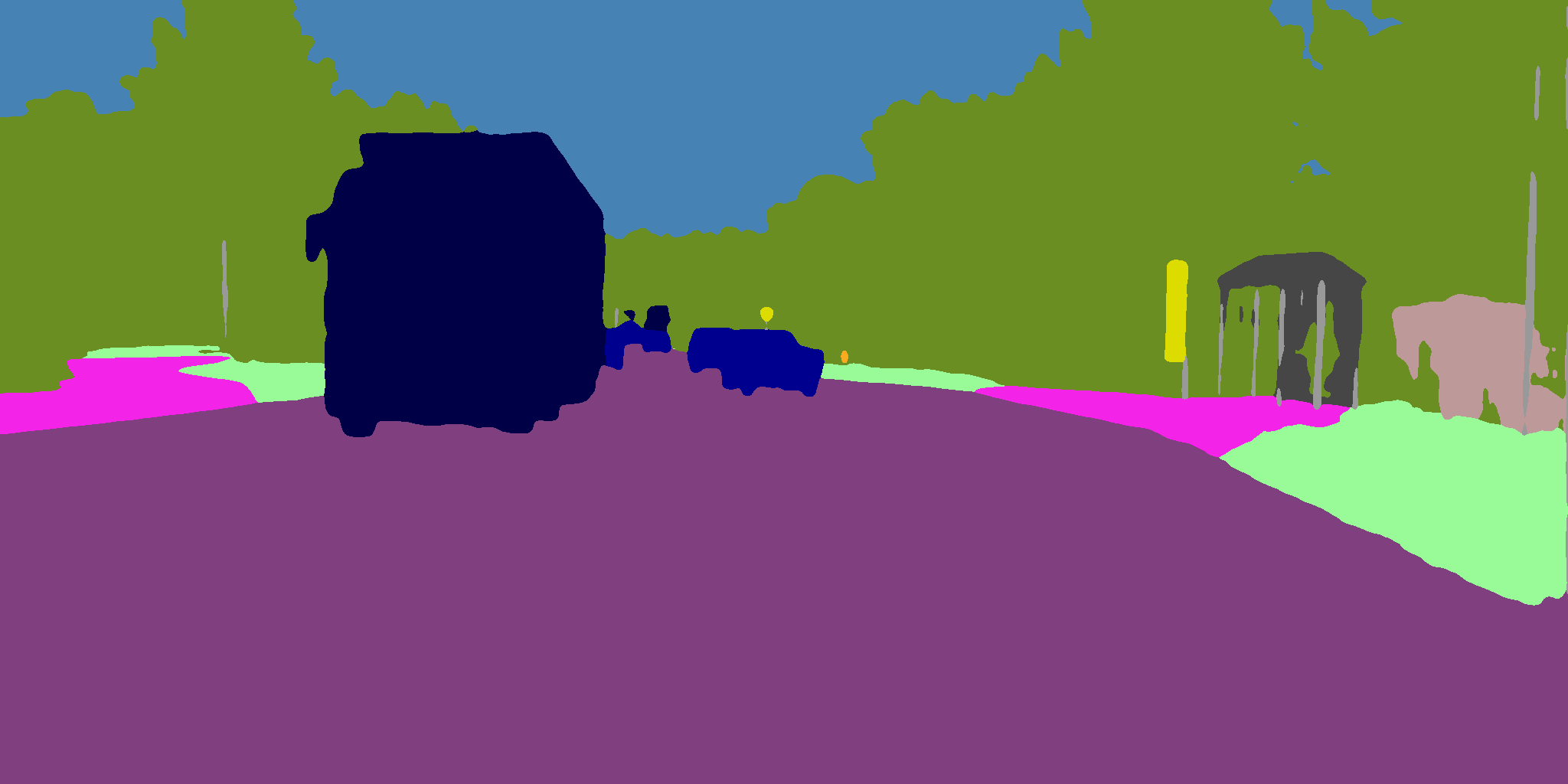}}
  \hfil
  \subfloat{\includegraphics[width=0.19\textwidth,height=0.1425\textwidth]{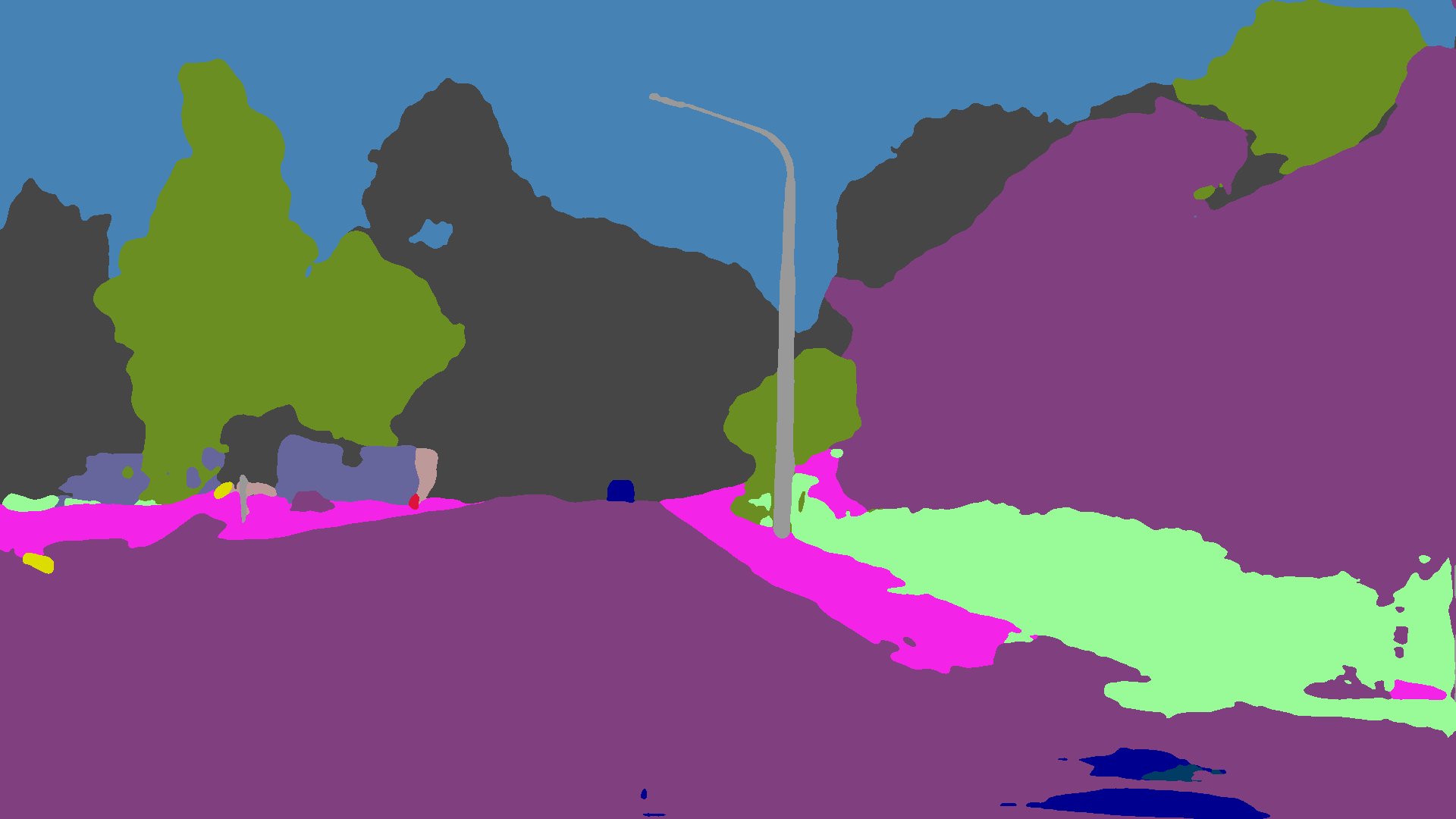}}\\ \vspace{-2.5mm}
  \subfloat{\rotatebox{90}{CMAda1}} 
      \hfil
  \subfloat{\includegraphics[width=0.19\textwidth,height=0.1425\textwidth]{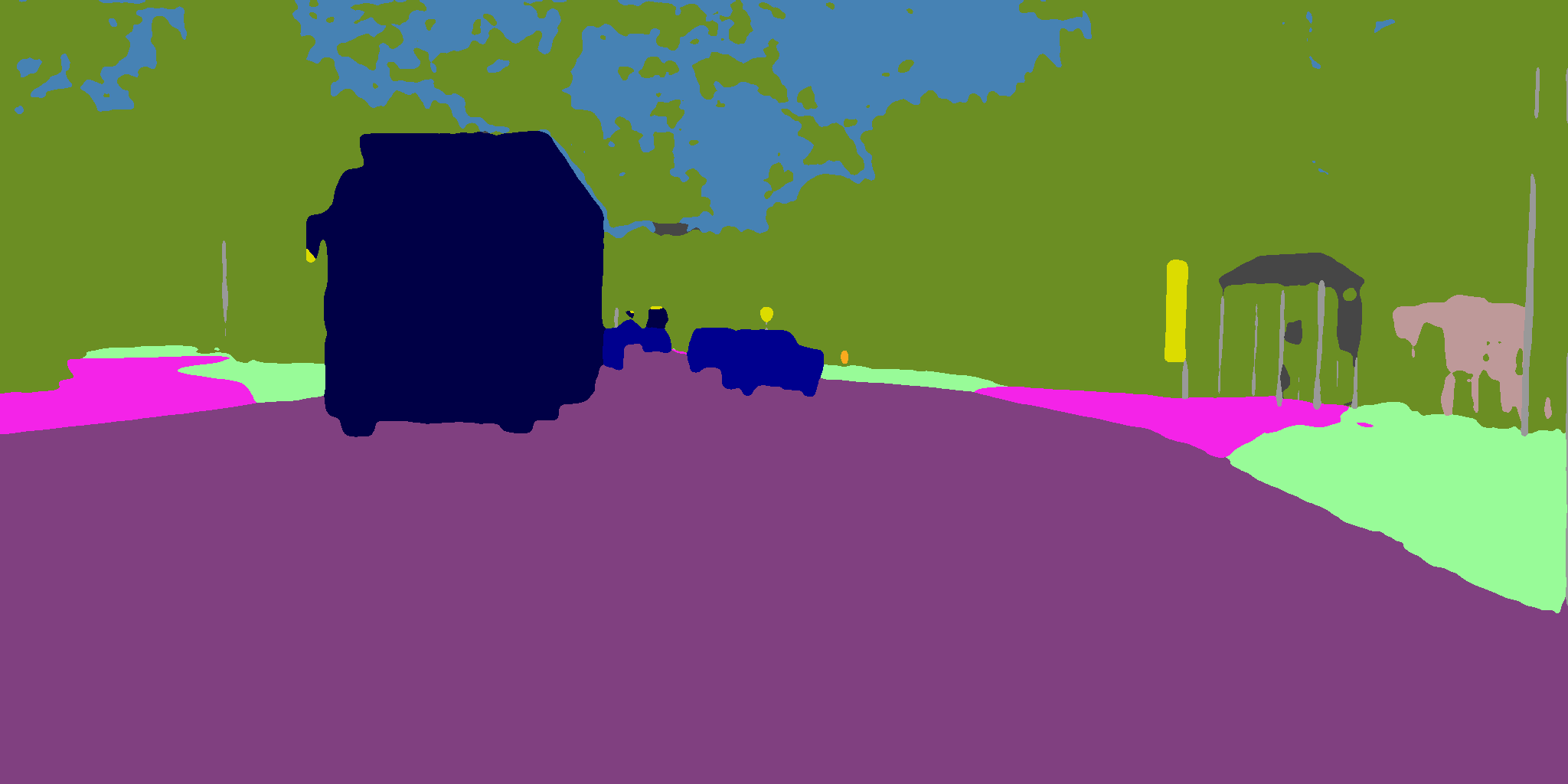}}
  \hfil 
  \subfloat{\includegraphics[width=0.19\textwidth,height=0.1425\textwidth]{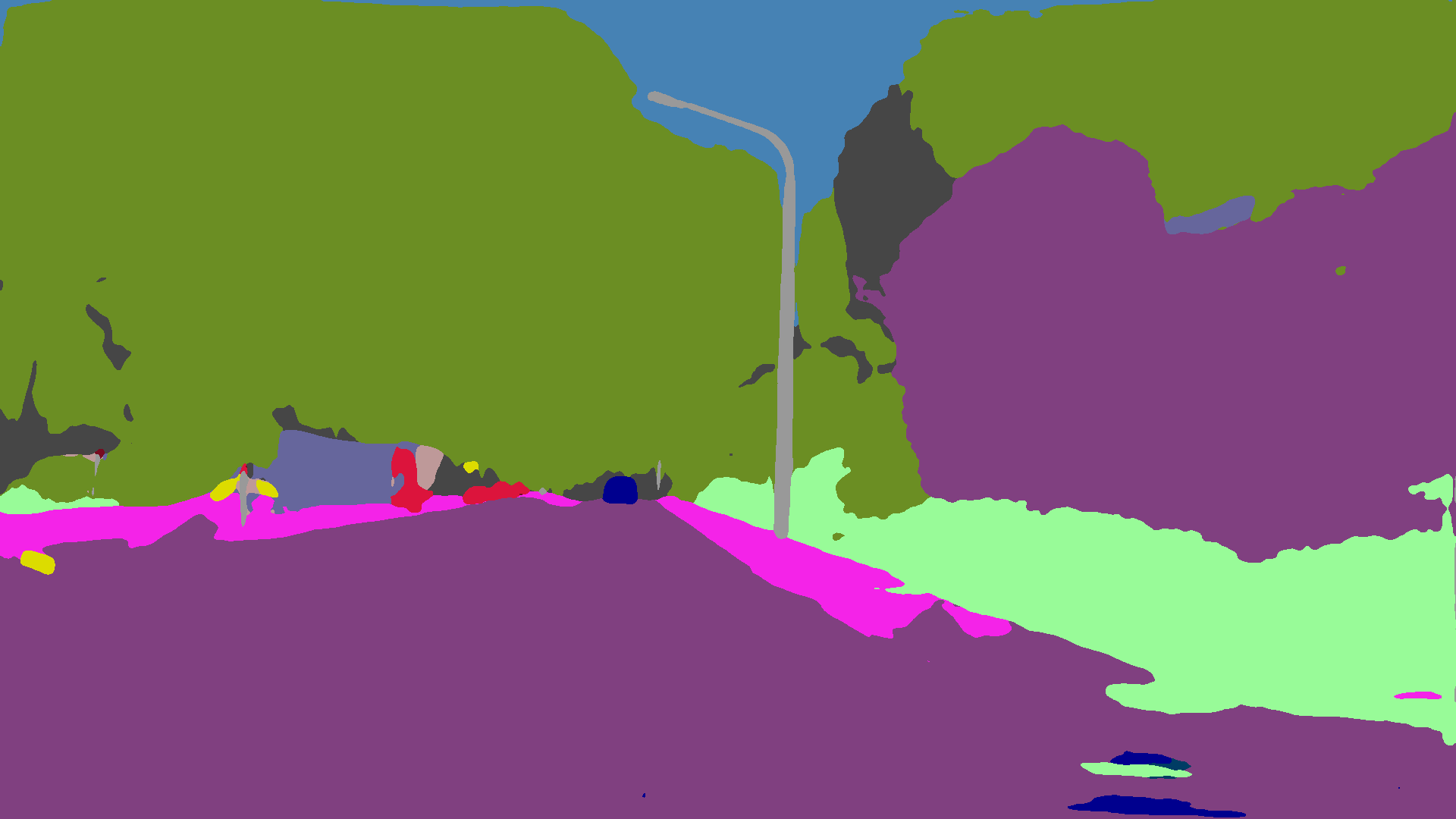}}\\  \vspace{-2.5mm}
   \subfloat{\rotatebox{90}{CMAda2}}
       \hfil
  \subfloat{\includegraphics[width=0.19\textwidth,height=0.1425\textwidth]{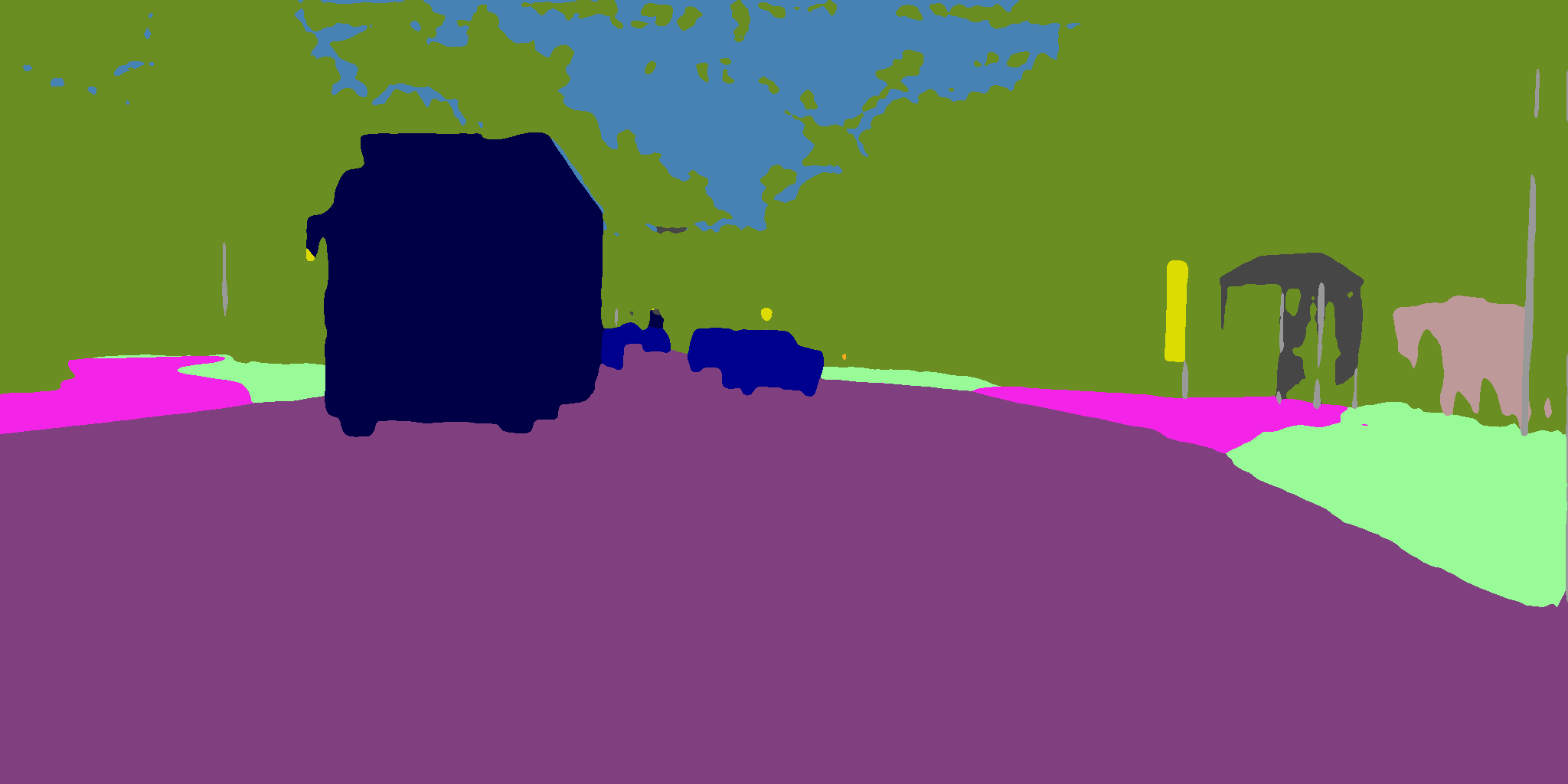}}
  \hfil 
  \subfloat{\includegraphics[width=0.19\textwidth,height=0.1425\textwidth]{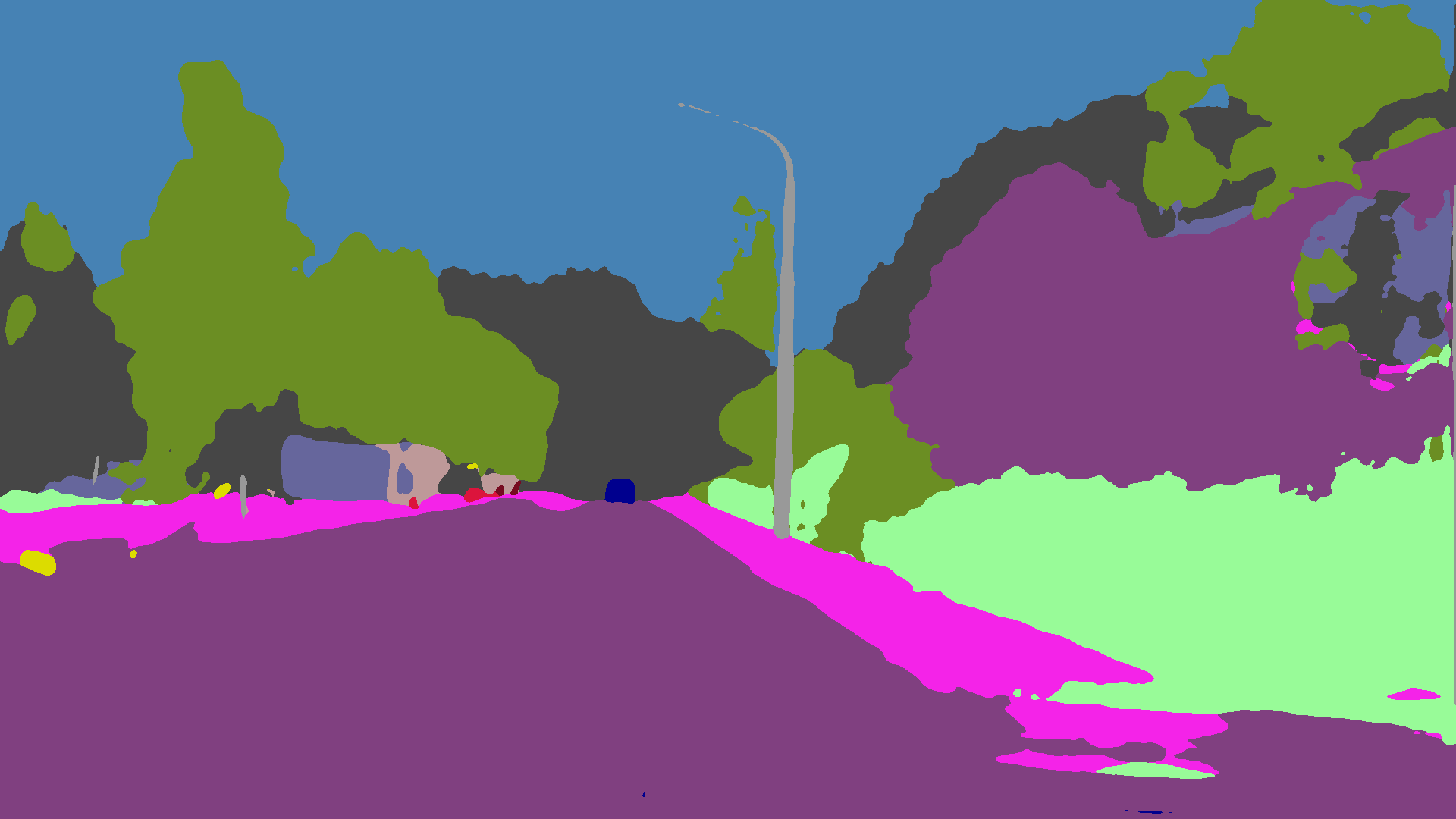}}\\  \vspace{-2.5mm}
   \subfloat{\rotatebox{90}{CMAda3+}}
       \hfil
  \subfloat{\includegraphics[width=0.19\textwidth,height=0.1425\textwidth]{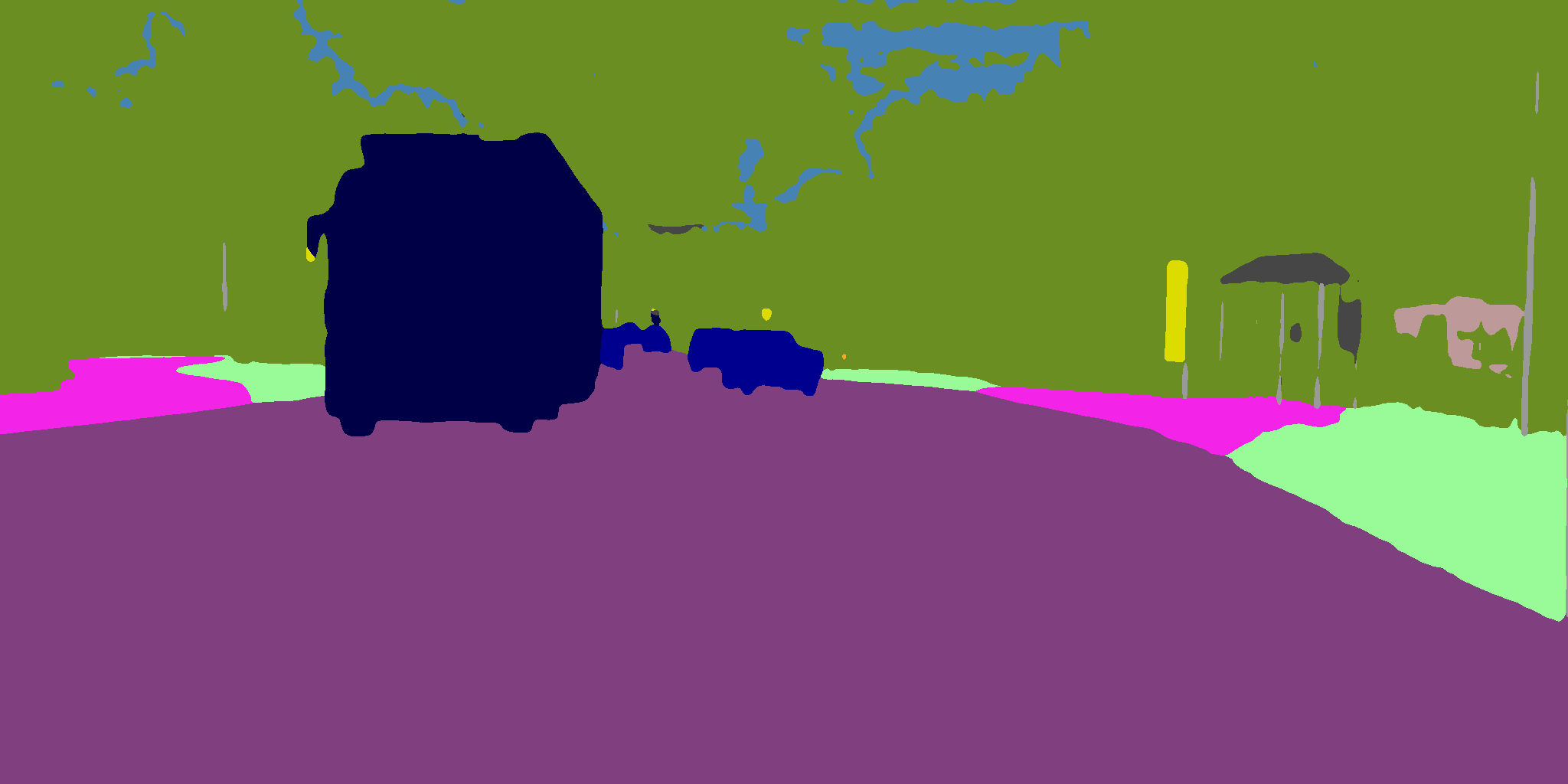}}
  \hfil 
  \subfloat{\includegraphics[width=0.19\textwidth,height=0.1425\textwidth]{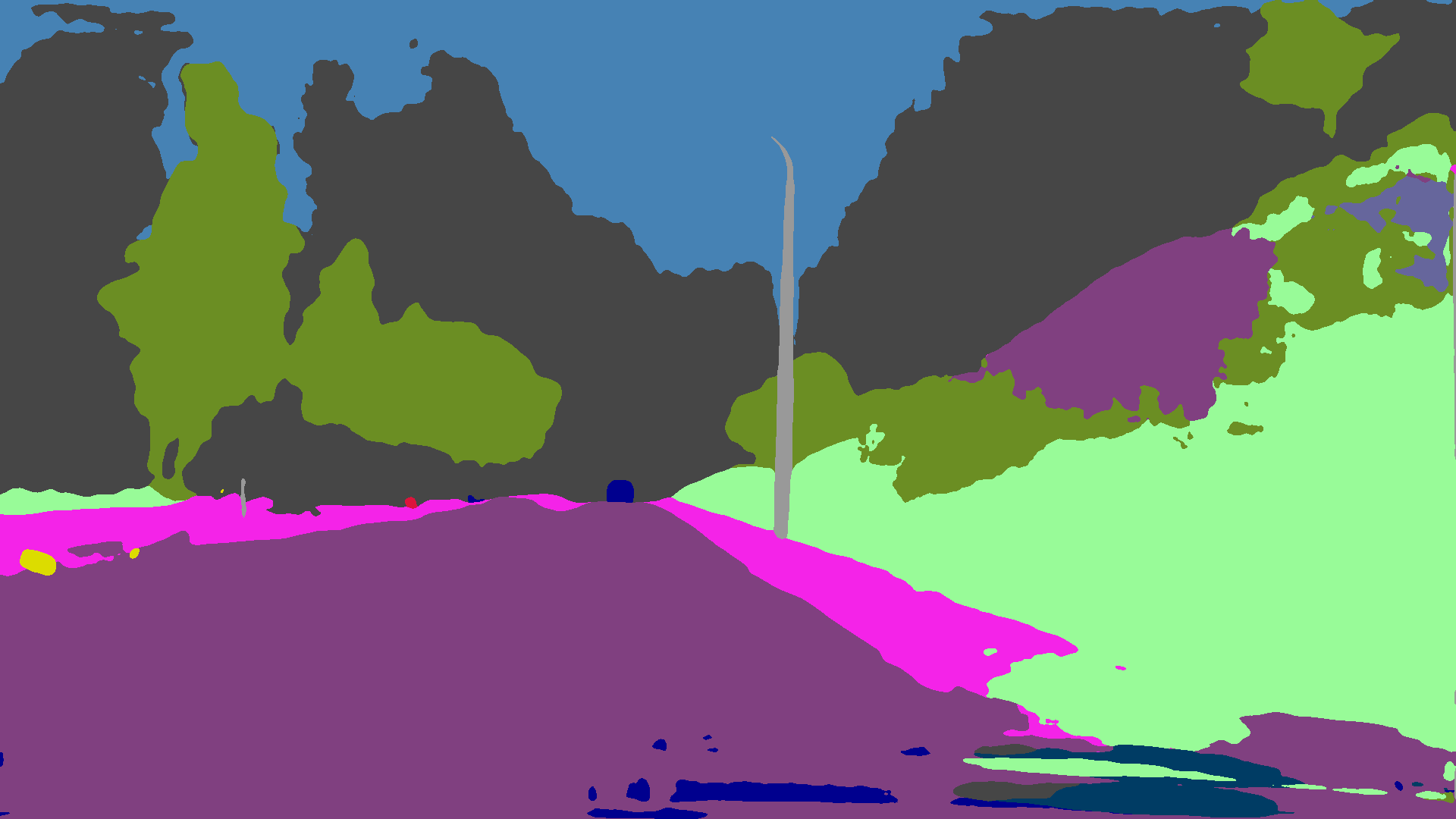}}\\  \vspace{-2.5mm} 
    \subfloat{\rotatebox{90}{Model Selection}}
       \hfil
  \subfloat{\includegraphics[width=0.19\textwidth,height=0.1425\textwidth]{./lindau_000002_000019_refinenet.png}}
  \hfil
  \subfloat{\includegraphics[width=0.19\textwidth,height=0.1425\textwidth]{./1_1_35_cma3+.png}}\\  \vspace{-2.5mm} 
     \subfloat{\rotatebox{90}{Ground Truth}}
         \hfil
  \subfloat{\includegraphics[width=0.19\textwidth,height=0.1425\textwidth]{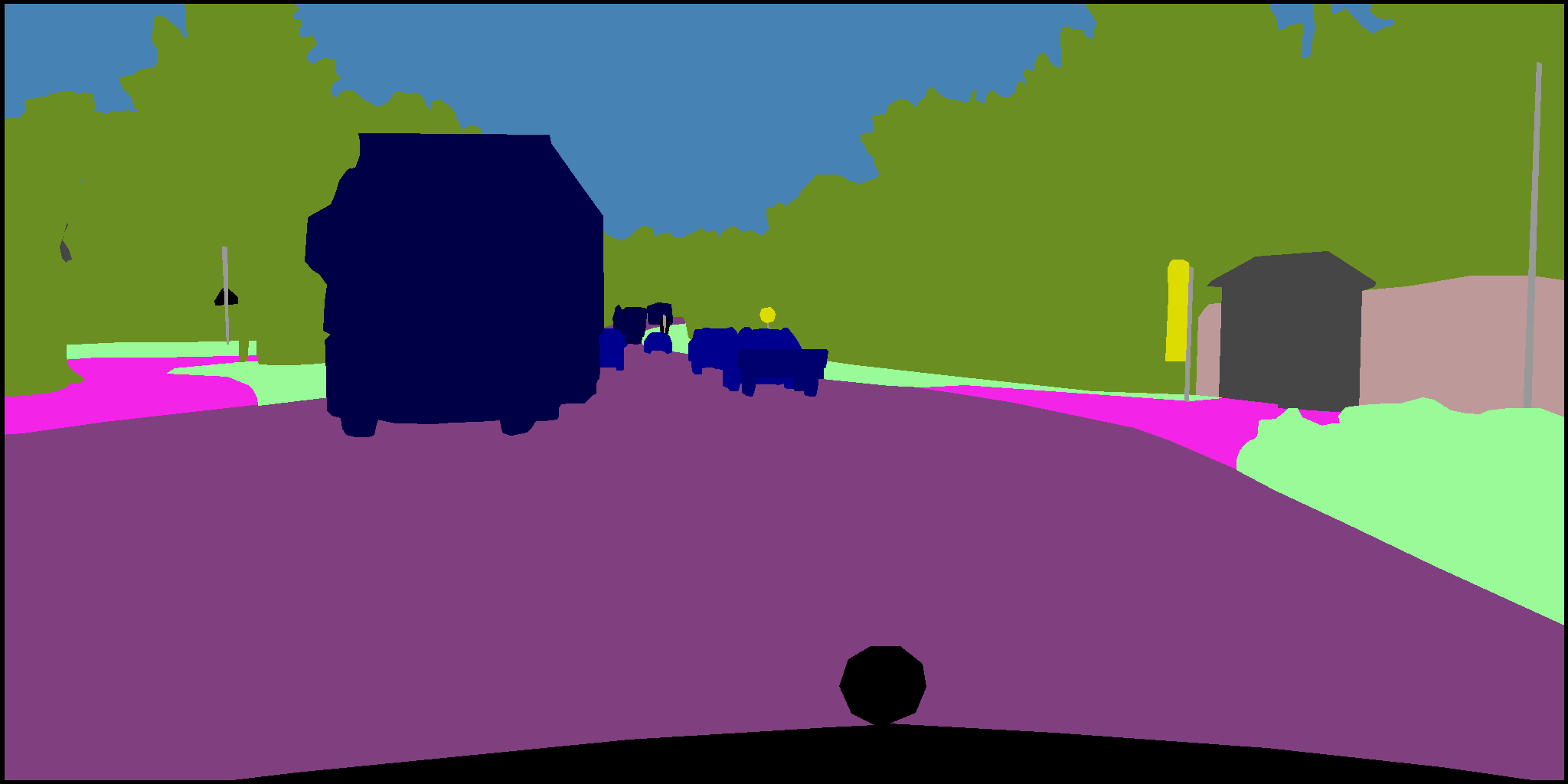}} 
   \hfil
  \subfloat{\includegraphics[width=0.19\textwidth,height=0.1425\textwidth]{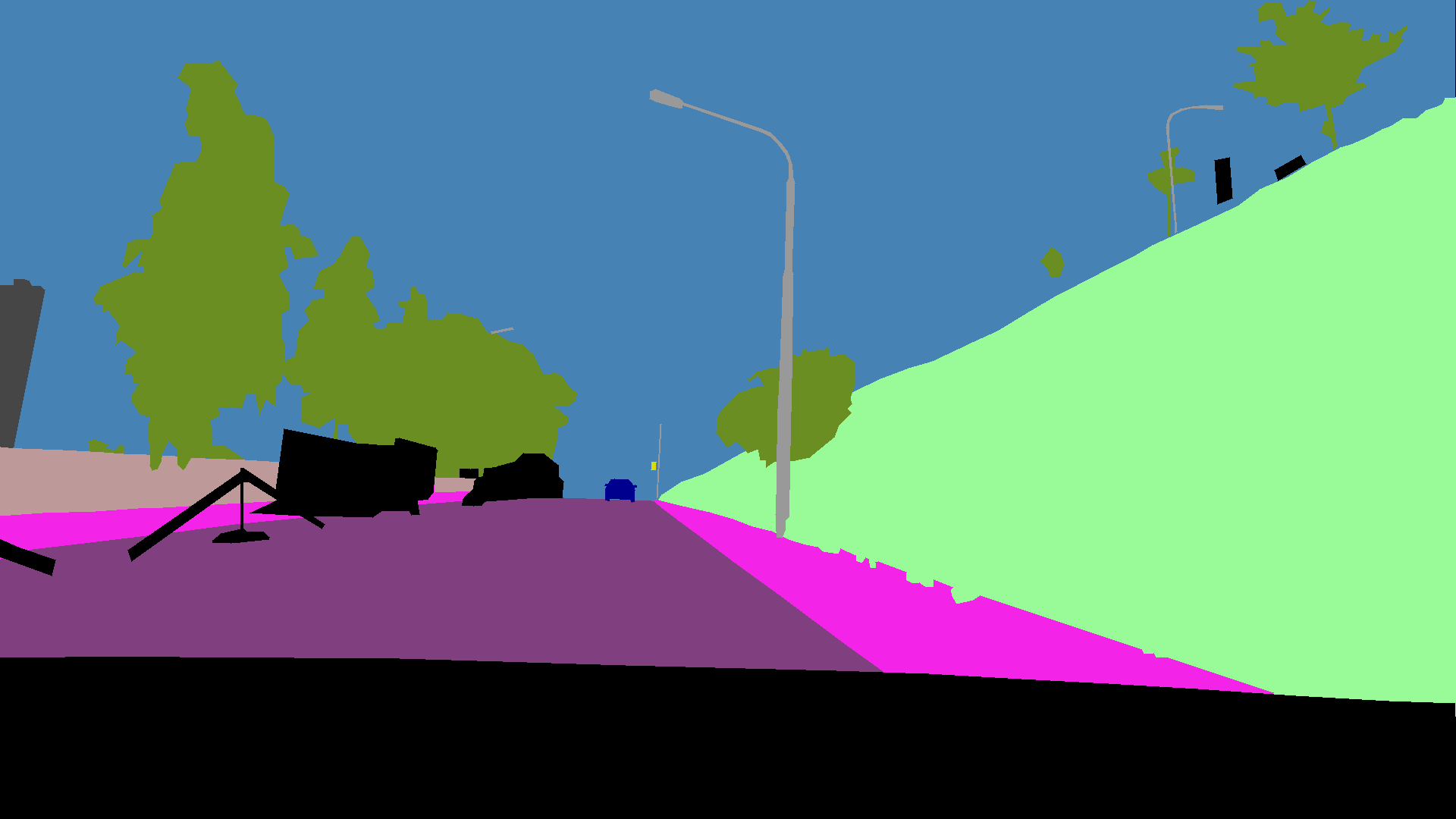}}\\ 
  \resizebox{\linewidth}{!}{
  \begin{tikzpicture}[tight background, scale=0.75, every node/.style={font=\large}]
      \draw[white, fill=void, draw=white] (0,0) rectangle (1 * 4, 1) node[pos=0.5] {Void};
      \draw[white, fill=road, draw=white] (1 * 4,0) rectangle (2 * 4, 1) node[pos=0.5] {Road};
      \draw[white, fill=sidewalk, draw=white] (2 * 4,0) rectangle (3 * 4, 1) node[pos=0.5] {Sidewalk};
      \draw[white, fill=building, draw=white] (3 * 4,0) rectangle (4 * 4, 1) node[pos=0.5] {Building};
      \draw[white, fill=wall, draw=white] (4 * 4,-0) rectangle (5 * 4, 1) node[pos=0.5] {Wall};
      \draw[black, fill=fence, draw=white] (5 * 4,-0) rectangle (6 * 4, 1) node[pos=0.5] {Fence};
      \draw[white, fill=pole, draw=white] (6 * 4,-0) rectangle (7 * 4, 1) node[pos=0.5] {Pole};
      \draw[white, fill=traffic light, draw=white] (7 * 4,-0) rectangle (8 * 4, 1) node[pos=0.5] {Traffic Light};
      \draw[black, fill=traffic sign, draw=white] (8 * 4,-0) rectangle (9 * 4, 1) node[pos=0.5] {Traffic Sign};
      \draw[white, fill=vegetation, draw=white] (9 * 4,-0) rectangle (10 * 4, 1) node[pos=0.5] {Vegetation};

      \draw[black, fill=terrain, draw=white] (0 * 4,-1) rectangle (1 * 4, 0) node[pos=0.5] {Terrain};
      \draw[white, fill=sky, draw=white] (1 * 4,-1) rectangle (4 * 2, 0) node[pos=0.5] {Sky};
      \draw[white, fill=person, draw=white] (2 * 4,-1) rectangle (3 * 4, 0) node[pos=0.5] {Person};
      \draw[white, fill=rider, draw=white] (3 * 4,-1) rectangle (4 * 4, 0) node[pos=0.5] {Rider};
      \draw[white, fill=car, draw=white] (4 * 4,-1) rectangle (5 * 4, 0) node[pos=0.5] {Car};
      \draw[white, fill=truck, draw=white] (5 * 4,-1) rectangle (6 * 4, 0) node[pos=0.5] {Truck};
      \draw[white, fill=bus, draw=white] (6 * 4,-1) rectangle (7 * 4, 0) node[pos=0.5] {Bus};
      \draw[white, fill=train, draw=white] (7 * 4,-1) rectangle (8 * 4, 0) node[pos=0.5] {Train};
      \draw[white, fill=motorcycle, draw=white] (8 * 4,-1) rectangle (9 * 4, 0) node[pos=0.5] {Motorcycle};
      \draw[white, fill=bicycle, draw=white] (9 * 4,-1) rectangle (10 * 4, 0) node[pos=0.5] {Bicycle};
  \end{tikzpicture}}
\caption{Qualitative semantic segmentation results under two weather conditions: clear weather (left) and foggy weather (right).}
\label{fig:sem:seg:multiple:weather}
\end{figure}

\subsection{Performance in Multiple Weather Conditions}
\label{sec:exp:multiple:weather}

We first note that the results which have been presented in Table~\ref{table:experiments:main} on the \emph{Foggy Driving} dataset~\cite{SFSU_synthetic}, which contains images of varying fog densities from very low to high, show that adaptation with CMAda to dense fog also brings a significant benefit for \emph{lower} fog densities.

In the following, we turn to evaluation of our Model Selection method presented in Section~\ref{sec:model:selection} for the task of \emph{semantic scene understanding in multiple weather conditions}. We consider two conditions: foggy weather and clear weather. This means that the test set comprises a mixture of images captured either in clear weather or under fog. In particular, we report the performance of three domain-specific methods and two variants of our Model Selection on three datasets. The three domain-specific methods are: 1) RefineNet, which is trained on Cityscapes dataset~\cite{refinenet} for clear weather, 2) CMAda2, which is trained for foggy weather, and 3) CMAda3+, which is also trained for foggy weather. The first variant of Model Selection uses RefineNet and CMAda2 as its two expert models and the second one uses RefineNet and CMAda3+ respectively. The three test datasets are \emph{Cityscapes-lindau-40}, \emph{Foggy Zurich-test}, and \emph{Clear-Foggy-80}, which is the union of the two previous sets. \emph{Cityscapes-lindau-40} contains the first 40 images (in alphabetical order) from the city of Lindau in the validation set of Cityscapes.

The performance of all five methods on the three datasets is reported in Table~\ref{table:model:selection}. We share a few observations. First, as discussed in previous sections, our adapted models significantly improve the recognition performance on foggy scenes. Second, it seems that some knowledge initially learned for recognition in clear-weather scenes is forgotten by our models during the adaptation process.  This is also evidenced by the visual comparison in Figure~\ref{fig:sem:seg:multiple:weather}, where the sky in the first image is misclassified after the adaptation.
This is because during the adaptation stages, we aim for the best expert model for (dense) foggy scenes and have not included any clear weather images. Adding some clear-weather images into the training data will alleviate this problem, but at a cost of lower performance on foggy scenes. Last but not least, both variants of our Model Selection method demonstrate higher performance than their constituent expert models. The second variant of Model Selection with RefineNet and CMAda3+ yields the best performance. It works especially well on the \emph{Clear-Foggy-80} dataset which contains $40$ foggy images and $40$ clear weather images, due to the good performance of the two expert models in their own domains. The improved performance with Model Selection implies that training multiple expert models---each for a different condition---and adaptively selecting the best one at testing time based on the input is a promising direction for semantic scene understanding in adverse conditions. We also demonstrate the improvement with Model Selection in Figure~\ref{fig:sem:seg:multiple:weather} when both clear weather and fog are considered.

\begin{table}[!tb]
  \centering
  \caption{Performance comparison of RefineNet (trained for clear weather), CMAda2 (trained for foggy weather), CMAda3+ (trained for foggy weather), and our Model Selection method on three datasets: \emph{Cityscapes-lindau-40} (clear weather), \emph{Foggy Zurich-test} (foggy weather) and the union of the two \emph{Clear-Foggy-80} (clear + foggy weather). ``MS\_R2'' stands for Model Selection with RefineNet and CMAda2 as the two expert models and ``MS\_R3+'' for Model Selection with RefineNet and CMAda3+ as the two expert models}
  \label{table:model:selection}
  \setlength\tabcolsep{3pt}
  \begin{tabular*}{\linewidth}{l @{\extracolsep{\fill}} ccccc}
  \multicolumn{6}{c}{Mean IoU over \emph{all} classes (\%)}\\
  \toprule
  Weather & RefineNet & CMAda2 & CMAda3+ & MS\_R2 & MS\_R3+\\
  \midrule
  Clear  & \textbf{67.2} & 65.1 & 59.6 & \textbf{67.2} & \textbf{67.2} \\
  Foggy & 34.6 & 42.9 & \textbf{46.8} & 42.9 & \textbf{46.8}  \\ 
  Clear + Foggy & 54.3 & 59.1 & 58.1 & 59.3 & \textbf{62.2} \\
  \bottomrule
  \end{tabular*}
\end{table}
  
\subsection{Investigating the Utility of Dehazing Preprocessing}
\label{sec:experiments:dehazing}



\begin{figure*}
  \centering
  \subfloat{\includegraphics[width=0.24\textwidth]{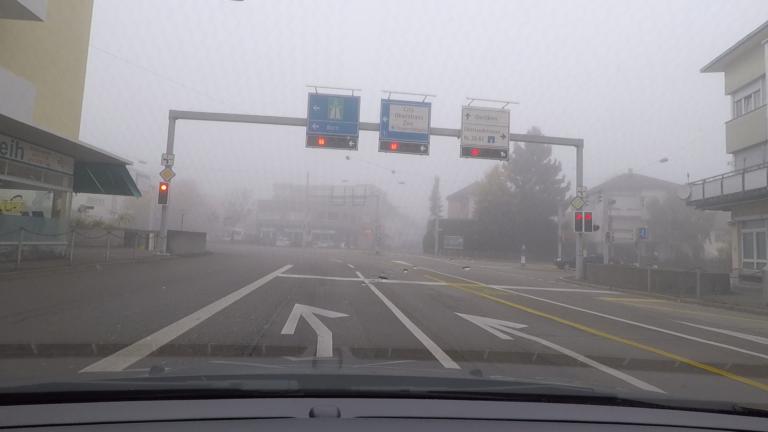}}
  \hfil
  \subfloat{\includegraphics[width=0.24\textwidth]{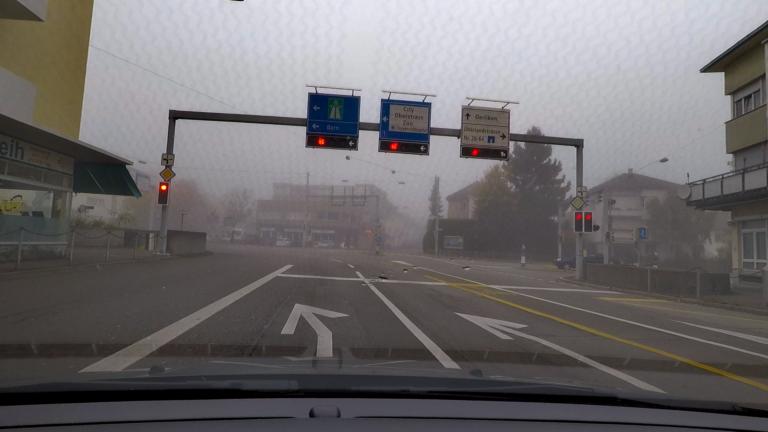}}
  \hfil
  \subfloat{\includegraphics[width=0.24\textwidth]{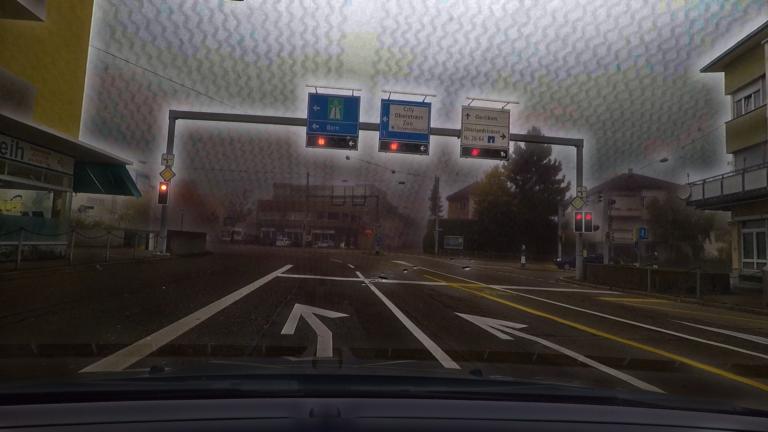}}
  \hfil
  \subfloat{\includegraphics[width=0.24\textwidth]{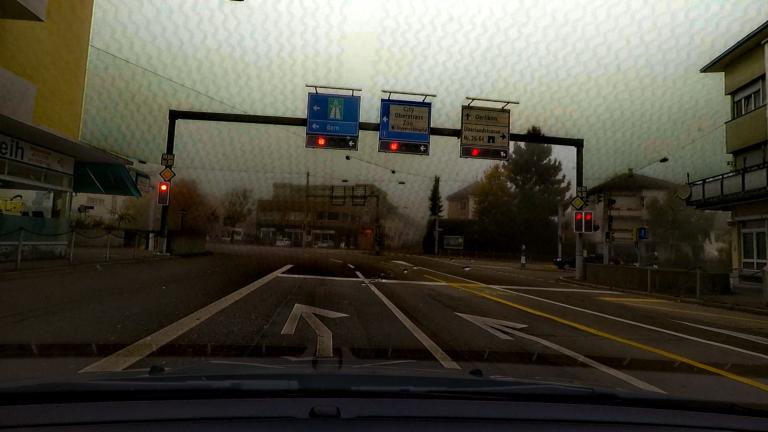}}\\
  \subfloat{\includegraphics[width=0.24\textwidth]{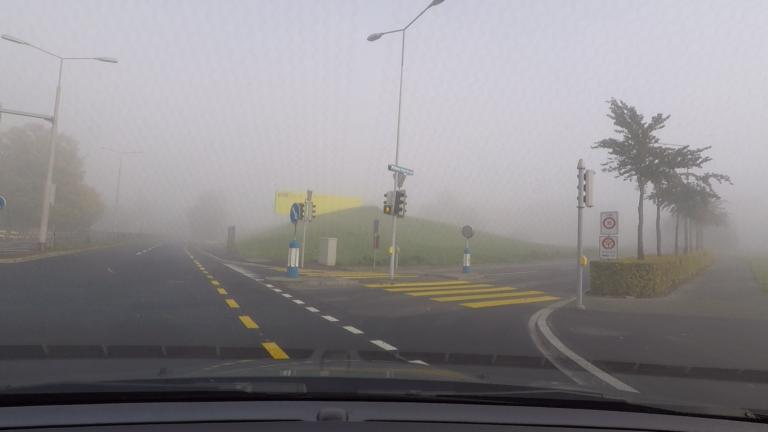}}
  \hfil
  \subfloat{\includegraphics[width=0.24\textwidth]{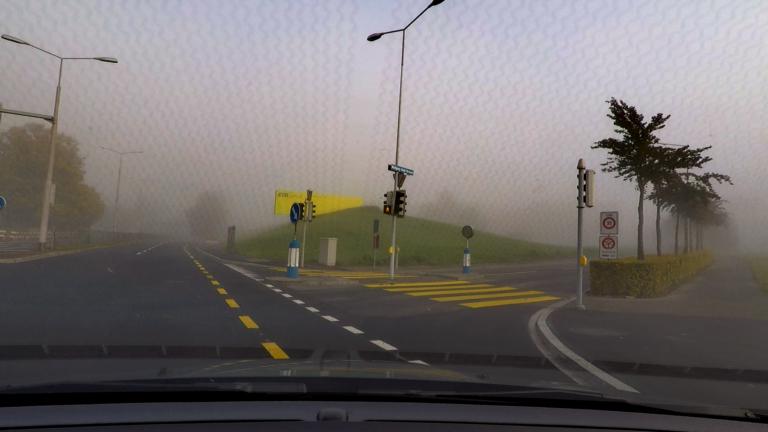}}
  \hfil
  \subfloat{\includegraphics[width=0.24\textwidth]{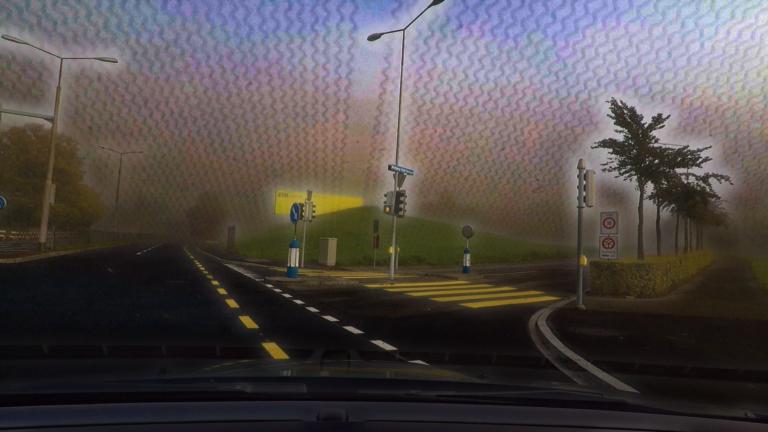}}
  \hfil
  \subfloat{\includegraphics[width=0.24\textwidth]{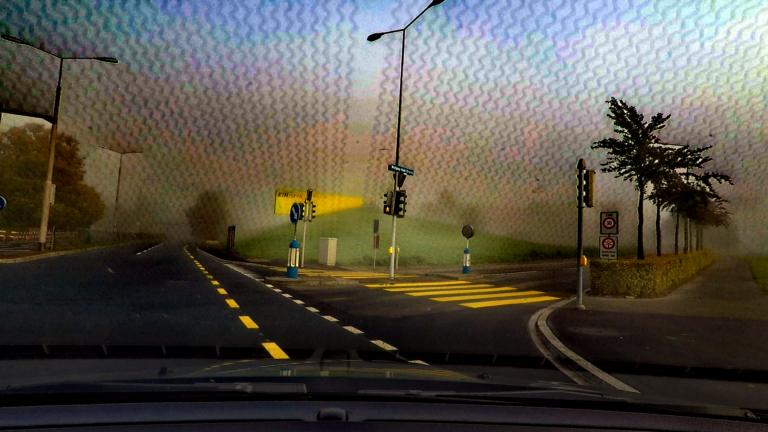}}\\
  \subfloat{\includegraphics[width=0.24\textwidth]{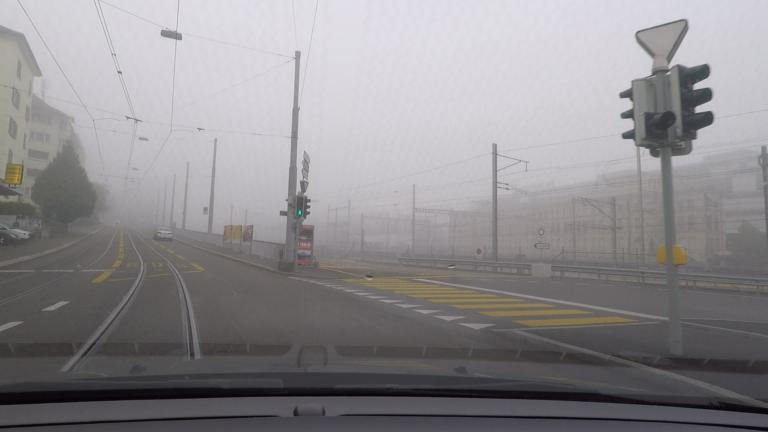}}
  \hfil
  \subfloat{\includegraphics[width=0.24\textwidth]{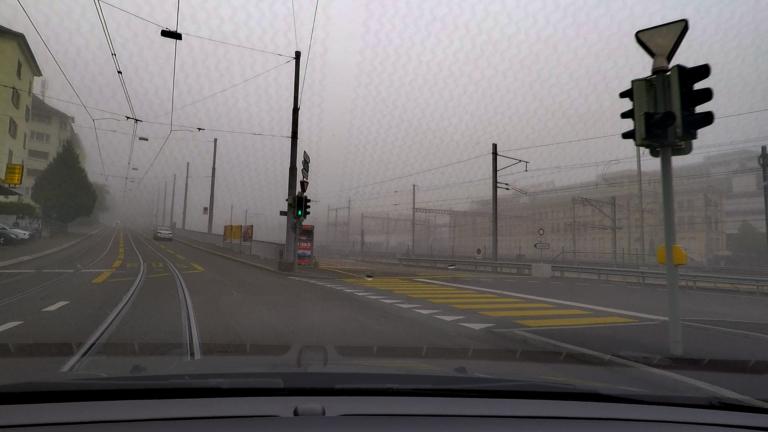}}
  \hfil
  \subfloat{\includegraphics[width=0.24\textwidth]{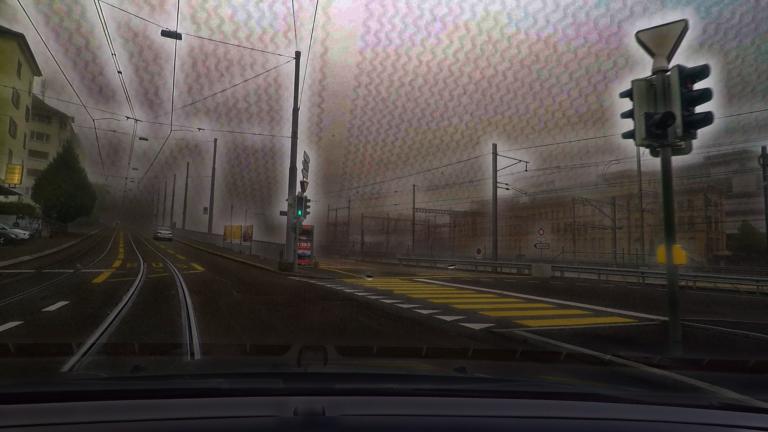}}
  \hfil
  \subfloat{\includegraphics[width=0.24\textwidth]{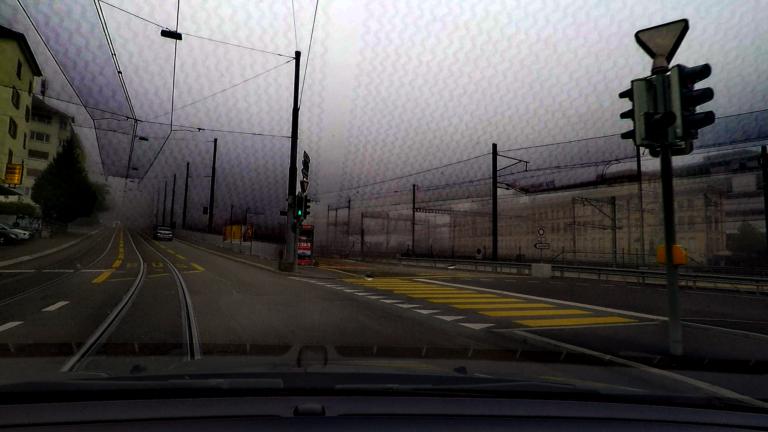}}\\
  \addtocounter{subfigure}{-12}
  \subfloat[]{\includegraphics[width=0.24\textwidth]{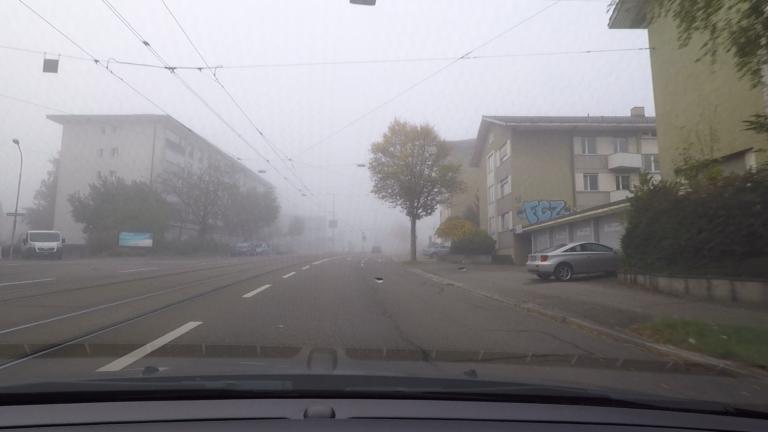}\label{fig:dehazing:foggy}}
  \hfil
  \subfloat[]{\includegraphics[width=0.24\textwidth]{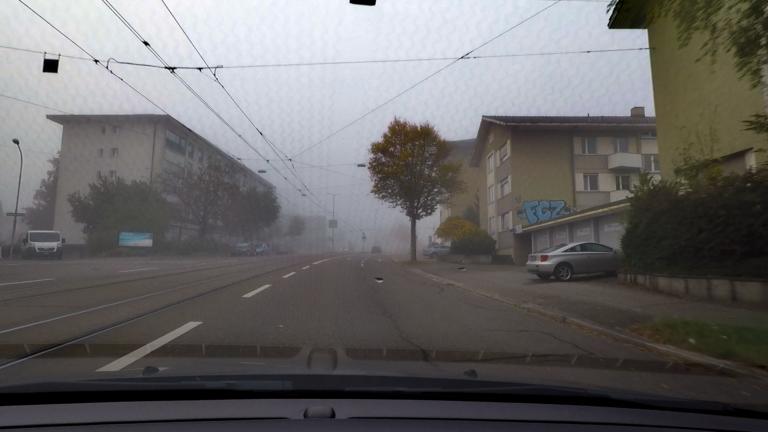}\label{fig:dehazing:mscnn}}
  \hfil
  \subfloat[]{\includegraphics[width=0.24\textwidth]{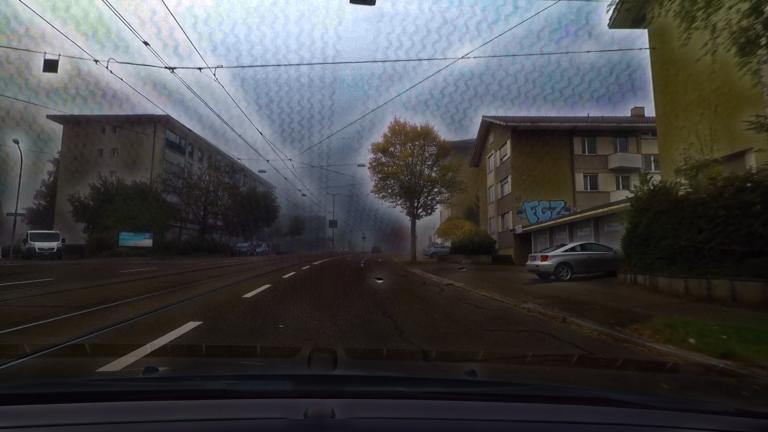}\label{fig:dehazing:dcp}}
  \hfil
  \subfloat[]{\includegraphics[width=0.24\textwidth]{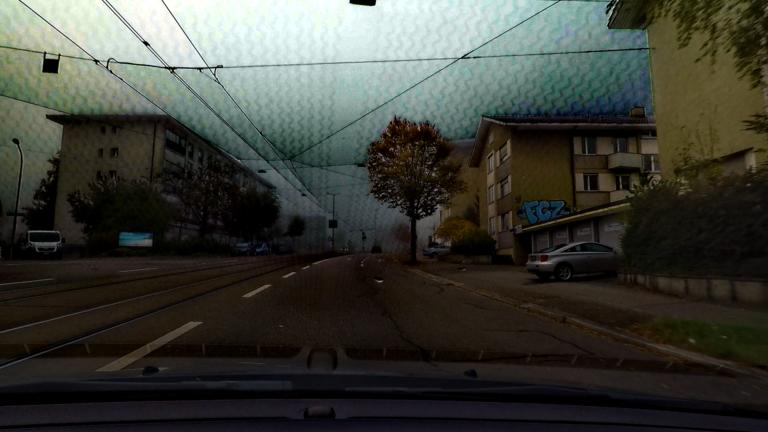}\label{fig:dehazing:nonlocal}}
  \caption{Representative images from \emph{Foggy Zurich-test} and dehazed versions of them obtained with the three dehazing methods that we consider in our experiments on utility of dehazing preprocessing. \protect\subref{fig:dehazing:foggy} \emph{Foggy Zurich-test} image. \protect\subref{fig:dehazing:mscnn} MSCNN~\cite{RLZ+16}. \protect\subref{fig:dehazing:dcp} DCP~\cite{dark:channel}. \protect\subref{fig:dehazing:nonlocal} Non-local~\cite{nonlocal:image:dehazing}. This figure is better seen on an screen and zoomed in}
  \label{fig:dehazing}
\end{figure*}

\sloppy{For completeness, we conduct an experimental comparison of the baseline RefineNet model of Table~\ref{table:experiments:main} and our single-stage CMAda pipeline using only synthetic training data against a dehazing preprocessing baseline, and report the results on \emph{Foggy Zurich-test} and \emph{Foggy Driving-dense} in Tables~\ref{table:dehazing:foggy_zurich} and \ref{table:dehazing:foggy_driving_dense} respectively. In particular, we consider dehazing as an optional preprocessing step before feeding the input foggy images to the segmentation model, and experiment with four options with respect to this dehazing preprocessing: no dehazing at all (already examined in Section~\ref{sec:experiments:synthetic}), multi-scale convolutional neural networks (MSCNN)~\cite{RLZ+16}, dark channel prior (DCP)~\cite{dark:channel}, and non-local dehazing~\cite{nonlocal:image:dehazing}. Apart from directly applying the original clear-weather RefineNet model on the dehazed test images, the results of which are included in the ``w/o FT'' rows of Tables~\ref{table:dehazing:foggy_zurich} and \ref{table:dehazing:foggy_driving_dense}, we also fine-tune this model on the dehazed versions of our synthetic \emph{Foggy Cityscapes-DBF} dataset, and compare against fine-tuning directly on the synthetic foggy images (already examined in Section~\ref{sec:experiments:synthetic}). Our experimental protocol is consistent: the same dehazing option is used both before fine-tuning and at testing time. The attenuation coefficient for \emph{Foggy Cityscapes-DBF} is $\beta=0.005$. The rest details are the same as in Section~\ref{sec:experiments:synthetic}. \emph{Not} applying dehazing generally leads to the best results irrespective of using the original model or fine-tuned versions of it. Fine-tuning without dehazing performs best in all cases but one (\emph{Foggy Driving-dense} and evaluation on all classes), which confirms the merit of our approach. This lack of significant improvement with dehazing preprocessing is in congruence with the findings of~\cite{SFSU_synthetic}, which has dissuaded us from including dehazing preprocessing in our default CMAda pipeline.}

Figure~\ref{fig:dehazing} illustrates the results of the examined dehazing methods on sample images from \emph{Foggy Zurich-test} and reveals the issues these methods face on real-world outdoor images with dense fog. Only MSCNN are able to slightly enhance the image contrast while introducing only minor artifacts. This correlates with the superior performance of the segmentation model that uses MSCNN for dehazing preprocessing compared to the models that use the other two methods, as reported in Table~\ref{table:dehazing:foggy_zurich}. Still, directly using the original foggy images generally outperforms all dehazing preprocessing alternatives.

\begin{table}[!tb]
  \centering
  \caption{Performance comparison on \emph{Foggy Zurich-test} of RefineNet (``w/o FT'') versus fine-tuned versions of it (``FT'') trained on \emph{Foggy Cityscapes-DBF} with attenuation coefficient $\beta=0.005$, for four options regarding dehazing: no dehazing, MSCNN~\cite{RLZ+16}, DCP~\cite{dark:channel}, and Non-local~\cite{nonlocal:image:dehazing}}
  \label{table:dehazing:foggy_zurich}
  \setlength\tabcolsep{3pt}
  \begin{tabular*}{\linewidth}{l @{\extracolsep{\fill}} cccc}
  \multicolumn{5}{c}{Mean IoU over \emph{all} classes (\%)}\\
  \toprule
   & No dehazing & MSCNN & DCP & Non-local\\
  \midrule
  w/o FT & \best{34.6} & 34.4 & 31.2 & 27.6\\
  FT & \best{36.7} & 36.1 & 34.2 & 29.1\\
  \bottomrule
  \end{tabular*}
  \\[\baselineskip]
  \setlength\tabcolsep{3pt}
  \begin{tabular*}{\linewidth}{l @{\extracolsep{\fill}} cccc}
  \multicolumn{5}{c}{Mean IoU over \emph{frequent} classes (\%)}\\
  \toprule
   & No dehazing & MSCNN & DCP & Non-local\\
  \midrule
  w/o FT & \best{51.8} & 48.6 & 42.9 & 41.1\\
  FT & \best{51.7} & 49.8 & 46.6 & 44.2\\
  \bottomrule
  \end{tabular*}
\end{table}

\begin{table}[!tb]
  \centering
  \caption{Performance comparison on {Foggy Driving-dense} of RefineNet (``w/o FT'') versus fine-tuned versions of it (``FT'') trained on \emph{Foggy Cityscapes-DBF} with attenuation coefficient $\beta=0.005$, for four options regarding dehazing: no dehazing, MSCNN~\cite{RLZ+16}, DCP~\cite{dark:channel}, and Non-local~\cite{nonlocal:image:dehazing}}
  \label{table:dehazing:foggy_driving_dense}
  \setlength\tabcolsep{3pt}
  \begin{tabular*}{\linewidth}{l @{\extracolsep{\fill}} cccc}
  \multicolumn{5}{c}{Mean IoU over \emph{all} classes (\%)}\\
  \toprule
   & No dehazing & MSCNN & DCP & Non-local\\
  \midrule
  w/o FT & 35.8 & \best{38.3} & 33.2 & 32.8\\
  FT & 36.6 & \best{40.0} & 35.8 & 37.5\\
  \bottomrule
 \end{tabular*}
  \\[\baselineskip]
  \setlength\tabcolsep{3pt}
  \begin{tabular*}{\linewidth}{l @{\extracolsep{\fill}} cccc}
  \multicolumn{5}{c}{Mean IoU over \emph{frequent} classes (\%)}\\
  \toprule
   & No dehazing & MSCNN & DCP & Non-local\\
  \midrule
  w/o FT & \best{57.6} & 55.5 & 47.4 & 50.7\\
  FT & \best{60.8} & 60.6 & 54.6 & 58.9\\
  \bottomrule
  \end{tabular*}
\end{table}

\section{Conclusion}
\label{sec:conclusion} 

In this article, we have shown the benefit of using partially synthetic as well as unlabeled real foggy data in a \emph{curriculum} adaptation framework to progressively improve performance of state-of-the-art semantic segmentation models in dense real fog. To this end, we have proposed a novel fog simulation approach on real scenes, which leverages the semantic annotation of the scene as additional input to a novel dual-reference cross-bilateral filter, and applied it to the Cityscapes dataset~\cite{Cityscapes} to obtain \emph{Foggy Cityscapes-DBF}. In addition, we have introduced a simple CNN-based fog density estimator which can benefit from large synthetic datasets such as \emph{Foggy Cityscapes-DBF} that provide straightforward ground truth for this task. On the real data side, we have presented \emph{Foggy Zurich}, a large-scale real-world dataset of foggy scenes, including pixel-level semantic annotations for 40 scenes with dense fog. Through extensive evaluation, we have showcased that: 1) our curriculum model adaptation exploits both our synthetic and our real data in a synergistic manner and significantly boosts performance on real fog without using any labeled real foggy image, and 2) our fog simulation and fog density estimation methods outperform their state-of-the-art counterparts.

\begin{acknowledgements}
This work is funded by Toyota Motor Europe via the research project TRACE-Z\"urich.
\end{acknowledgements}

\bibliographystyle{spmpsci}
\bibliography{references.bib}

\end{document}